\documentclass[journal]{IEEEtran}

\usepackage{cite}
\usepackage{amssymb}
\usepackage{amsmath}
\usepackage{epsfig}
\usepackage{graphicx}
\usepackage{amssymb}
\usepackage{amsfonts}
\usepackage{colortbl}
\usepackage{multirow}
\usepackage{longtable}
\usepackage{bm}
\usepackage{multicol}
\usepackage{algorithmic}
\usepackage{amsmath}
\usepackage{amsthm}
\usepackage[linesnumbered,ruled,vlined]{algorithm2e}
\usepackage{mathtools}
\usepackage[dvipsnames,table]{xcolor}
\usepackage{subfigure}
\usepackage{paralist}
\usepackage{color}
\usepackage{tabu}
\usepackage{rotating}
\usepackage{pifont}
\usepackage{threeparttable}
\usepackage{moresize}
\usepackage{cancel}
\usepackage{pgf-pie}
\usepackage{adjustbox}
\usepackage{url}
\usepackage{subcaption}
\usepackage{booktabs}
\usepackage{array}
\usepackage{makecell}
\usepackage{float}
\usepackage{setspace}

\newcolumntype{L}{>{\raggedleft}p{0.14\textwidth}}
\newcolumntype{R}{p{0.8\textwidth}}

\newtheorem*{example*}{Example}

    \usepackage{xr}
\makeatletter
\newcommand*{\addFileDependency}[1]{
  \typeout{(#1)}
  \@addtofilelist{#1}
  \IfFileExists{#1}{}{\typeout{No file #1.}}
}
\makeatother
\newcommand*{\myexternaldocument}[1]{
    \externaldocument{#1}
    \addFileDependency{#1.tex}
    \addFileDependency{#1.aux}
}
\myexternaldocument{supplement}

\renewcommand{\vec}{\mathbf}

\title{A Generalized and Configurable Benchmark Generator for Continuous Unconstrained Numerical Optimization}

\author{%
Amir H. Gandomi\textsuperscript{1,2},
Mohammad Nabi Omidvar\textsuperscript{3},
Rohit Salgotra\textsuperscript{1,4},
Kalyanmoy Deb\textsuperscript{5}\\[2mm]
\textsuperscript{1}Faculty of Engineering and Information Technology, University of Technology Sydney, Australia\\
\textsuperscript{2}Research and Innovation Center (EKIK), Obuda University, Budapest, Hungary\\
\textsuperscript{3}School of Computing, University of Leeds, and Leeds University Business School, UK\\
\textsuperscript{4}AGH University of Krakow, Poland\\
\textsuperscript{5}BEACON Center, Michigan State University, USA%
\thanks{The Australian Government supported this work through the Australian Research Council, Australia, under Project DE210101808.}%
}

\begin{document}
\maketitle

\begin{abstract}
As optimization challenges continue to evolve, so too must our tools and understanding. 
To effectively assess, validate, and compare optimization algorithms, it is crucial to use a benchmark test suite that encompasses a diverse range of problem instances with various characteristics.
Traditional benchmark suites often consist of numerous fixed test functions, making it challenging to align these with specific research objectives, such as the systematic evaluation of algorithms under controllable conditions. 
This paper introduces the Generalized Numerical Benchmark Generator (GNBG) for single-objective, box-constrained, continuous numerical optimization. 
Unlike the commonly used test suites that rely on multiple baseline functions and transformations, GNBG utilizes a single, parametric, and configurable baseline function. 
This design allows for control over various problem characteristics.
Researchers using GNBG can generate instances that cover a broad range of morphological features, from unimodal to highly multimodal functions, various local optima patterns, and symmetric to highly asymmetric structures. 
The generated problems can also vary in separability, variable interaction structures, dimensionality, conditioning, and basin shapes. 
These customizable features enable the systematic evaluation and comparison of optimization methods, allowing researchers to examine the strengths and weaknesses of algorithms under diverse and controllable conditions.
\end{abstract}

\IEEEpeerreviewmaketitle



 \section{Introduction}
 \label{sec:Introduction}
 
\IEEEPARstart{O}PTIMIZATION algorithms have been the subject of intense research and development over the past decades, with applications spanning a variety of domains, such as data science~\cite{cousineau2023estimating}, engineering~\cite{rao2019engineering}, and transportation~\cite{archetti2022optimization}. 
Reliable and comprehensive benchmarking of these algorithms is a crucial task.
A fundamental research question in this context is to determine how effectively an algorithm performs on problems that present specific characteristics, challenges, and levels of difficulty. 
While theoretical analyses offer insights, they can be prohibitively difficult to conduct for complex algorithms and problem instances. 
Consequently, empirical evaluation becomes the method of choice, typically executed by solving a predefined set of benchmark problem instances~\cite{beiranvand2017best}.

To ensure the robust design and effectiveness of optimization algorithms, the use of standardized benchmark test suites is essential~\cite{bartz2020benchmarking}.
These suites consist of mathematical functions with known characteristics, which enable researchers to investigate the strengths and weaknesses of optimization methods under different conditions~\cite{hansen2009real}. 
By providing a standardized basis for comparison, benchmark test suites facilitate the development of more effective optimization algorithms and advance the field of optimization.

A proper benchmark test suite should be designed to be easy to understand and facilitate a clear understanding of the behavior and performance of optimization algorithms within the search space. 
This aids researchers in visualizing the intended search behavior and identifying the weaknesses and strengths of the optimization algorithms. 
By analyzing the performance of algorithms in this manner, researchers can systematically modify the algorithms, ultimately leading to improved performance. 
The following are several key characteristics that are generally considered important for a comprehensive benchmark suite~\cite{whitley2002testing,shir2018compiling}.
\paragraph*{Diversity}
An ideal benchmark test suite should encompass a diverse collection of problem instances that exhibit a range of problem characteristics encountered in practical applications~\cite{malan2013survey,munoz2015algorithm,munoz2020generating}. 
This diversity enables a comprehensive evaluation and comparison of optimization algorithm performance under various conditions. 
\paragraph*{Complexity Variety}
A proper benchmark test suite should encompass problem instances with a range of complexity levels~\cite{uchoa2017new}, determined by various factors such as modality (unimodal to highly multimodal), dimensionality, separability, conditioning, and deceptiveness~\cite{kerschke2019comprehensive}.
\paragraph*{Algorithmic Neutrality}
To ensure a fair evaluation of optimization algorithms, a benchmark test suite should mitigate certain problem characteristics that inherently favor specific algorithms/operators. 
For example, symmetric problem instances, which are in favor of algorithms that rely on Gaussian distributions for generating new solutions~\cite{hansen2009real}, and problem instances with the optimum positioned on the boundary, which can advantage methods utilizing absorption boundary handling~\cite{gandomi2012evolutionary,helwig2012experimental}, should be avoided~\cite{beiranvand2017best}.
\paragraph*{Practicality}
The ultimate goal of any benchmarking exercise is to draw robust conclusions about the performance of algorithms. 
To make these conclusions as accurate and generalizable as possible, the benchmark suite should closely mirror the characteristics, complexities, and challenges commonly encountered in real-world problems~\cite{skaalnes2023new,uchoa2017new}. 

\paragraph*{Configurability}
Configurability is a critical aspect of a benchmark suite, which provides researchers with the ability to make fine-grained adjustments to a wide range of problem characteristics. 
This includes, but is not limited to, dimensionality, conditioning, complexity of variable interaction structures, and other morphological characteristics.
It is worth noting that the ability to configure the number of dimensions is often referred to as scalability~\cite{hansen2009real,omidvar2015designing}. 
\paragraph*{Known characteristics and optimal solution(s)}
The benchmark test suite should provide information on the morphological characteristics, the major challenges, and the position and value of the optimal solution(s) for each problem instance~\cite{olson2017pmlb}. 
This information plays a vital role in analyzing the convergence behavior, performance, strengths, and weaknesses of optimization algorithms.
\paragraph*{Accessibility}
A benchmark test suite should include publicly available source code and documentation, ensuring accessibility to the research community.

Numerous benchmark suites exist in the literature to evaluate and compare the performance of optimization algorithms across different sub-fields, such as large-scale optimization~\cite{omidvar2015designing}, multi-objective optimization~\cite{deb2002scalable}, dynamic optimization~\cite{yazdani2020benchmarking}, and constrained optimization~\cite{hellwig2019benchmarking}. 
The focus of this paper is specifically on box-constrained continuous single-objective global optimization, a sub-field that seeks to identify the global optimum of a given optimization problem within a specified search range.

Benchmarking in this context involves comparing the best found solutions by different algorithms using performance indicators~\cite{hansen2021coco}. 
Such global optimization problems are pervasive in various fields, particularly in mathematics and engineering disciplines~\cite{mei2021structural}. 
Employing appropriate benchmark test suites in this domain is not just an academic exercise; it lays the foundation for advancements in more complex optimization problems~\cite{weise2009global}, including dynamic~\cite{yazdani2021DOPsurveyPartA,yazdani2021DOPsurveyPartB}, constrained~\cite{kusakci2012constrained}, large-scale~\cite{omidvar2021reviewA,omidvar2021reviewB}, niching~\cite{li2016seeking}, and multi-objective optimization~\cite{saini2021multi}. 
For the sake of brevity, the term `optimization' used throughout the rest of this paper should be understood to refer specifically to `box-constrained continuous single-objective global optimization.'

Currently, the commonly used benchmarking approaches rely on a collection of well-established mathematical functions, such as the Sphere, Ellipsoid, Rosenbrock, Rastrigin, Schwefel, Griewangk, and Ackley functions, as well as compositions of these functions~\cite{suganthan2005problem,gouvea2016global,jamil2013literature,more1981testing,molga2005test,chelouah2003genetic}.
Often, these functions are subjected to standard transformations such as translation (shift) and rotation to simulate a wider range of problem characteristics~\cite{hansen2009real,awad2016problem,yue2019problem}. 
However, this approach has two major limitations.
\begin{itemize}
\item The inherent characteristics of these mathematical functions are generally predefined and fixed, which limits flexibility for fine-grained analysis. 
While these suites aim for comprehensive coverage by incorporating a wide range of mathematical functions, this abundance can actually complicate the task of understanding the benchmark suite.
As a result, analyses may not adequately reveal the strengths and weaknesses of algorithms across diverse problem characteristics.
\item Existing benchmark suites often lack the ability to configure and control specific problem characteristics, such as the width and depth of local optima, conditioning, or variable interaction structures, thereby hampering targeted evaluations.
This limitation can be a significant obstacle for researchers aiming to explore how optimization algorithms handle specific problem characteristics under various configurations, such as different degrees of conditioning and complexity of variable interaction structure.
\end{itemize}

While the commonly used benchmarks usually focus on a collection of mathematical functions, there have been efforts to develop generalized benchmark formulations~\cite{michalewicz2000test}. 
However, these generalized benchmarks are still limited in their ability to generate diverse controllable characteristics and configurations necessary for comprehensive evaluations.

In light of these limitations, this paper introduces the Generalized Numerical Benchmark Generator (GNBG), a configurable, flexible, and user-friendly tool explicitly designed to embody the desirable properties of an effective benchmark suite.
GNBG employs a single, parametric baseline function capable of generating a diverse range of problem instances with controllable characteristics and levels of difficulty. 
By manipulating various parameters within GNBG, users gain the ability to tailor the properties of the generated problem instances, including:
\begin{itemize}
\item Modality: GNBG can generate diverse problem instances, from smooth, unimodal search spaces to highly multimodal and rugged landscapes, with control over the width and depth of local optima. 
This adaptability allows researchers to comprehensively evaluate how well optimization algorithms navigate different types of terrain.

\item Local Optima Characteristics: GNBG constructs its search space through the integration of multiple independent components, each having its own `basin of attraction'--a zone where solutions tend to converge. 
Users can configure various aspects of these components, such as their locations, optimum values, and morphological features. 
This high level of control extends to the characteristics of any local optima within these basins, allowing for customization of their number, size, width, depth, and shape.

\item Gradient Characteristics: GNBG allows users to control not just the steepness of the components but also the specific rate of change or curvature of their basins. 
Users have the flexibility to define these characteristics on a per-component basis, with options ranging from highly sub-linear to super-linear rates of change.

\item Variable Interaction Structures: GNBG allows detailed control over variable interactions within generated problem instances. 
Users can customize rotation matrices to configure interaction structures, from fully separable to fully-connected non-separable, and set the strength of these interactions. 
Different regions of the search space can have distinct variable interaction patterns.

\item Conditioning: GNBG provides users with the ability to generate components with a wide range of condition numbers, spanning from well-conditioned to severely ill-conditioned components. 

\item Symmetry: GNBG affords the flexibility to generate both symmetric and highly asymmetric problem instances. 
This is achieved by allowing the strategic distribution of components with varied morphological characteristics across the search space. 
Furthermore, GNBG provides the capability to generate components with asymmetric basins of attraction. 

\item GNBG allows users to introduce varying degrees of deception into problem instances by manipulating the size, location, and depth of components. 
This enables the creation of scenarios where the global optimum is hidden within a wider local optimum or separated from high-quality local optima. 
Researchers can thus assess how well algorithms navigate misleading landscapes.

\item Scalability: All problem instances generated by GNBG are scalable with respect to dimensionality. 
\end{itemize}

While the user has insights into the characteristics of problem instances generated by GNBG, it is crucial to note that these instances are treated as black boxes by the optimization algorithms. 
That is, the algorithms operate without access to the internal structure or specific properties of these instances, interacting solely through the evaluation of candidate solutions and the function values they receive. 


The main contributions of GNBG can be summarized as follows:
\begin{itemize}
\item GNBG operates on a foundational, generalized framework using a singular parametric baseline function.
\item GNBG offers the flexibility to generate a multitude of problem instances, each presenting controllable degrees of challenges and various characteristics, allowing researchers to tailor them to specific research objectives.
\item One of GNBG's key features is its ability for isolated challenge evaluation. It can uniquely craft problem instances that emphasize specific challenging characteristics at varying intensities.
\item GNBG fulfills all the requirements of a high-quality benchmark, including attributes such as diversity, varied complexity, algorithmic neutrality, practicality, configurability, scalability, known characteristics and optimal solutions, and accessibility.
\end{itemize}

The rest of this paper is organized as follows.
Section~\ref{sec:GNBG} provides details about GNBG, explaining its architecture and how different parameter settings impact problem characteristics.
Section~\ref{sec:ProblemInstances} outlines how GNBG can be used to generate problem instances with specific characteristics.
Section~\ref{sec:Conclusion} concludes the paper, summarizing key findings and implications. 
Additionally, this manuscript is accompanied by a supplementary document that provides complementary context.
Sections~\ref{sec:sup:proof} and~\ref{sec:sup:GNBGparameters} of the supplementary document include mathematical proofs and a summary of the parameters of GNBG, respectively.
Section~\ref{sec:sup:ImpactONalgorithms} presents a preliminary empirical study exploring the influence of various problem characteristics on the performance of selected optimization algorithms. 
Finally, Section~\ref{sec:sup:Suite} introduces a test suite comprising 24 different problem instances generated by GNBG.

\section{Generalized Numerical Benchmark Generator}
\label{sec:GNBG}

In this section, we provide an overview of the Generalized Numerical Benchmark Generator (GNBG). 
We begin by presenting the baseline mathematical function central to GNBG, followed by a discussion of GNBG's parameters and their roles in shaping the generated optimization challenges\footnote{The MATLAB and Python source codes for the GNBG problem instance generator are available at~\cite{yazdani2024GNBGgeneratorMatlab,yazdani2024GNBGgeneratorPython}. Users can employ these source codes to generate custom problem instances as per their requirements.}.

\subsection{Baseline Mathematical Formulation}
\label{sec:sec:baseline}

The search space in GNBG is a composite landscape formed by aggregating multiple distinct components, each characterized by its unique basin of attraction. These components, which must all have the same dimensionality, contribute individual challenges and complexities to the overall search space.

To explain how these components form a comprehensive optimization problem, we introduce the baseline function of GNBG, expressed as:
\begin{align}
\label{eq:irGMPB}
\resizebox{\columnwidth}{!}{$
f(\vec{x})=   \min_{k\in\{1,\dots,o\}}\left\{ \sigma_{k} + \bigg(\mathbb{T}_k\Big((\mathbf{R}_{k}(\vec{x}-\vec{m}_k))^\top\Big)  \mathbf{H}_{k} \mathbb{T}_k\Big(\mathbf{R}_{k}(\vec{x}-\vec{m}_k)\Big)\bigg)^{\lambda_{k}} \right\},$}
\end{align}
\vspace{-15pt}
\begin{align}
\resizebox{\columnwidth}{!}{$
\mathrm{~~~~~~~~~Subject~to:}  \;\;\vec{x}\in \mathbb{X}:\;\mathbb{X}=\{\vec{x}\;|\;l_i \leq x_i \leq u_i\}, \; i \in\{1,2,\ldots,d\},$}\nonumber
\end{align}
where $f(\cdot)$ represents the GNBG function to be minimized, $d$ is the number of dimensions, and $\mathbb{X}$ denotes the $d$-dimensional search space, and $\vec{x}$ is a candidate solution. 
The search space is bounded by $l_i$ and $u_i$ for each dimension $i$. 
The term $o$ represents the number of components, each with its own parameters: $\sigma_k$, $\vec{m}_k$, $\mathbf{H}_{k}$, $\mathbf{R}_k$, and $\lambda_k$. 
For the $k$-th component, $\vec{m}_k$ defines its center, $\sigma_k$ specifies its minimum value ($f(\vec{m}_k)$), $\mathbf{H}_{k}$ is a diagonal matrix affecting basin heights, $\mathbf{R}_k$ is the rotation matrix, and $\lambda_k$ quantifies linearity. 
The $\min(\cdot)$ function defines the basin of attraction for each component and $\mathbb{T}_{k}(\vec{a})\mapsto \vec{b}$ is a non-linear transformation function introducing additional complexities.
This transformation function maps each element $a_j\in \vec{a}$ to:
\begin{align}
\label{eq:ir}
\resizebox{\columnwidth}{!}{$
    a_j \mapsto
    \begin{dcases}
    \exp{\bigg(\log(a_j)+\mu_{k,1}\Big(\sin{\big(\omega_{k,1}\log(a_j)\big)}+\sin{\big(\omega_{k,2}\log(a_j)\big)} \Big)\bigg)} & \text{if   } a_j>0 \\
    0 & \text{if   }a_j=0\\
    -\exp{\bigg(\log(|a_j|)+\mu_{k,2}\Big(\sin{\big(\omega_{k,3}\log(|a_j|)\big)}+\sin{\big(\omega_{k,4}\log(|a_j|)\big)} \Big)\bigg)} & \text{if   } a_j<0  
\end{dcases},$}
\end{align}
where for the $k$-th component, this transformation is guided by parameters $\bm\mu_k$ and $\bm\omega_k$, which define the symmetry and morphology of local optima on the basin of the $k$-th component.
The transformation function $\mathbb{T}_{k}(\cdot)$ operates based on the value of each element $a_j \in \vec{a}$: for $a_j > 0$, it applies an exponential modulation controlled by $\mu_{k,1}$ and frequency parameters $\omega_{k,1}$ and $\omega_{k,2}$; for $a_j = 0$, the value is mapped directly to zero, independent of $\bm\mu$ or $\bm\omega$ parameters, ensuring the minimum position $\vec{m}_k$ of the $k$-th component remains unchanged; and for $a_j < 0$, it mirrors the process for $a_j > 0$, acting on $|a_j|$ before negating it, controlled by $\mu_{k,2}$, $\omega_{k,3}$, and $\omega_{k,4}$.

\subsection{Parameter Sensitivity and Influence Analysis}
\label{sec:sec:ParameterAnalysis}

In this section, we conduct an in-depth analysis of the parameters in GNBG, supported by illustrative examples. 
Understanding how these parameters influence the morphology, complexity, and behavior of the landscape is crucial for effectively configuring GNBG to create customized problem instances that align with specific research objectives.

To better understand the influence of GNBG's various parameters, we start by simplifying the model to a more basic, unimodal form. 
This involves neutralizing certain parameters and transformations. 
Specifically, we focus on the parameters $o$, $\bm\mu$, and $\bm\omega$, which dictate the modality of the landscape.

\begin{figure*}[!t]
\centering
\begin{tabular}{ccc}
    \subfigure[{\scriptsize $\lambda=0.25$, sub-linear basin}]{\includegraphics[width=0.30\linewidth]{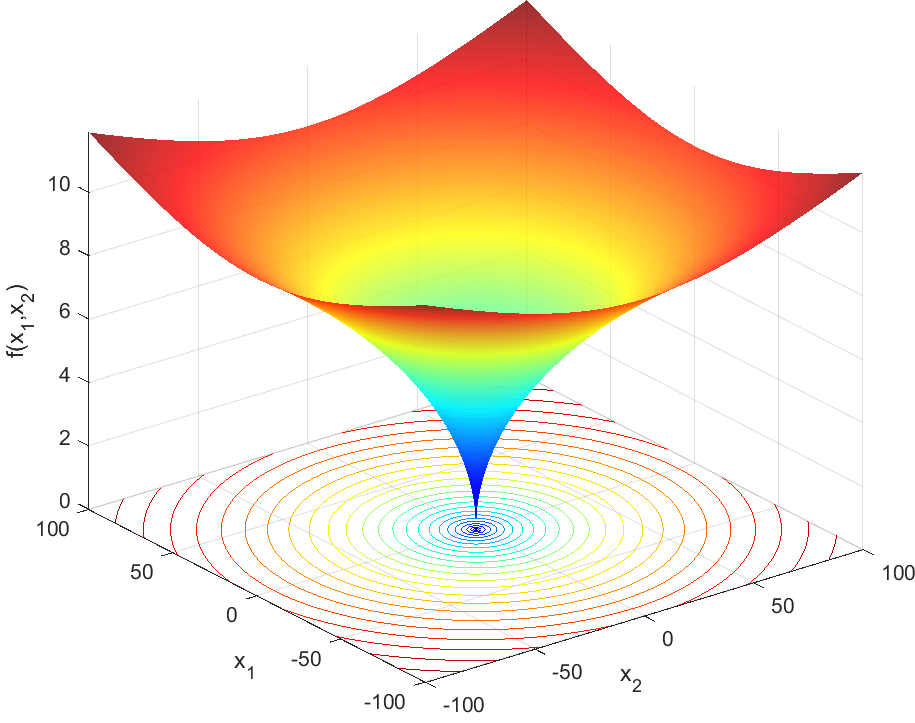}\label{fig:lambdaLog}}
    &
    \subfigure[{\scriptsize $\lambda=0.5$, linear basin}]{\includegraphics[width=0.30\linewidth]{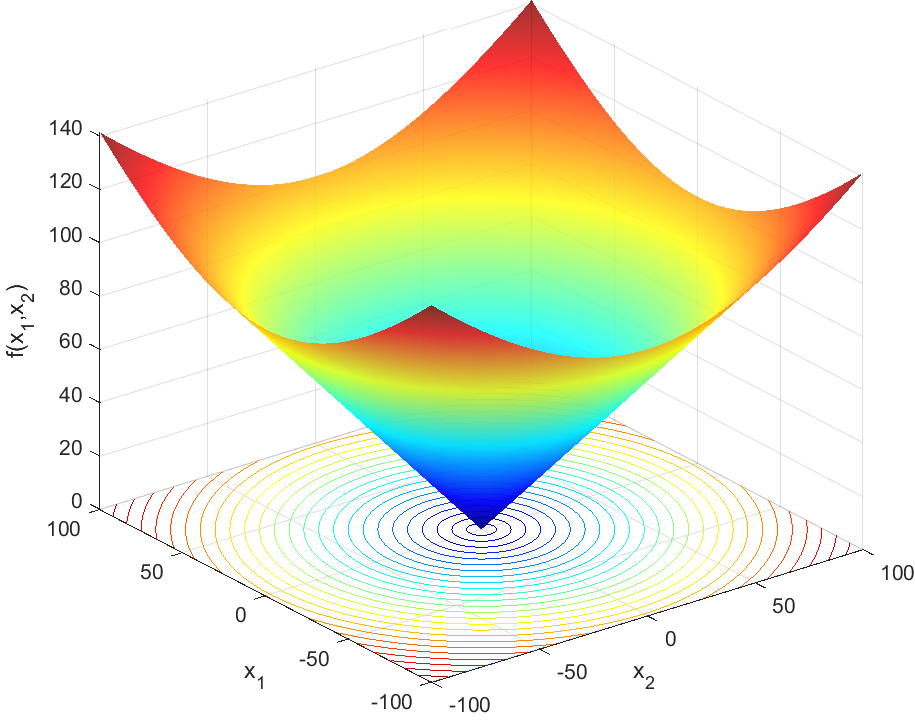}\label{fig:lambdaLinear}}
     &
     \subfigure[{\scriptsize $\lambda=1$, super-linear basin}]{\includegraphics[width=0.30\linewidth]{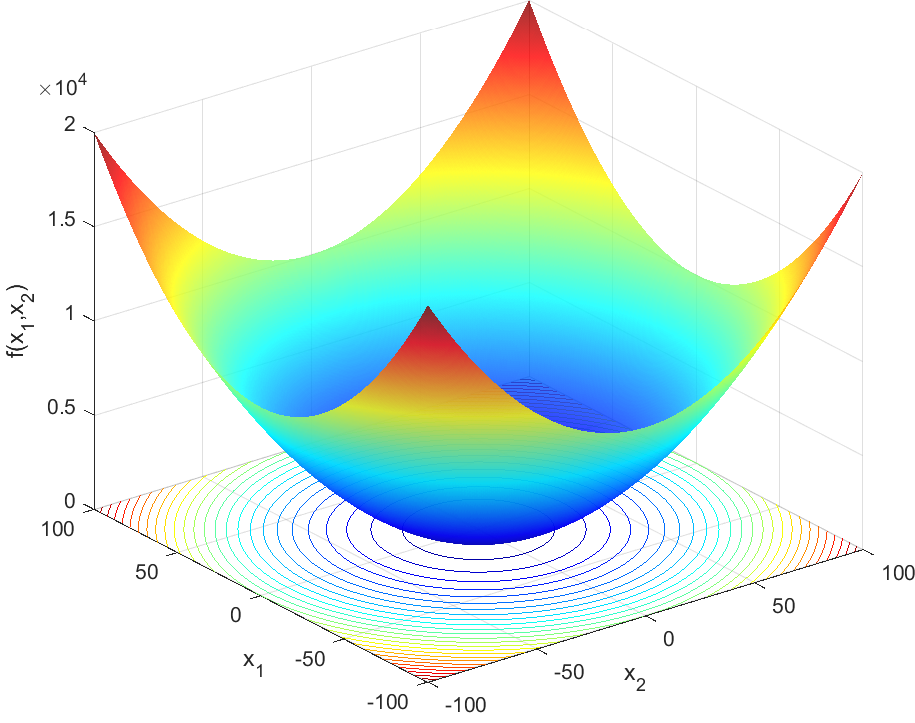}\label{fig:lambdaExp}}
\end{tabular}
\caption{Impact of $\lambda$ values on the morphology of a component generated by GNBG. 
For these illustrative examples, we set $d=2$, $o=1$, $\bm\mu=(0,0)$, $\bm\omega=(0,0,0,0)$, $\sigma=0$, $\vec{m}=(0,0)$, $\mathbf{H}=\mathbf{I}_{2 \times 2}$, and $\mathbf{R}=\mathbf{I}_{2 \times 2}$. 
Additionally, the 2-dimensional problem space is bounded to [-100,100] in each dimension. 
For a component generated by GNBG, $\lambda < 0.5$ yields a sub-linear basin, $\lambda = 0.5$ yields a linear basin, and $\lambda > 0.5$ yields a super-linear basin.}

\label{fig:Lambda}
\end{figure*}

To generate a unimodal landscape, we set $o=1$, indicating that the landscape is constructed from a single component.
In this configuration, the $\min(\cdot)$ function in GNBG's original equation becomes redundant and can be omitted.
Next, to make the component unimodal, we set elements in vectors $\bm\mu$ and $\bm\omega$ to zero.
This effectively neutralizes the transformation $\mathbb{T}$ in equation \eqref{eq:ir} and changes it to an identity mapping $a_j \mapsto a_j$.
Therefore, we can omit the transformation $\mathbb{T}$ in the GNBG's function in this case. With these adjustments, GNBG's baseline function can be rewritten as:
\begin{align}
\label{eq:Unimodal}
f(\vec{x})=  \sigma + \left((\mathbf{R}(\vec{x}-\vec{m}))^\top  \mathbf{H} \mathbf{R}(\vec{x}-\vec{m})\right)^{\lambda}.
\end{align}
In this simplified form, GNBG can only generate unimodal problem instances with $\vec{m}$ and $\sigma$ representing the global optimum position and value, respectively.
To further simplify GNBG, we set $\mathbf{H}$ and $\mathbf{R}$ to $\mathbf{I}_{d \times d}$.
Setting matrices $\mathbf{H}$ and $\mathbf{R}$ to the identity matrix neutralizes their impact on the problem landscape.
In this case, we can omit these matrices from GNBG's formulation.
In addition, we set $\sigma=0$ and $\vec{m}= \{ m_i = 0 \,|\, i = 1,2,\ldots, d \}$ to neutralize their impact and removing them from GNBG's formulation.
Therefore, using this configuration, GNBG's baseline can be rewritten as its simplest form\footnote{Equation \eqref{eq:Unimodal1} can be rewritten as $\left(\sum_{i=1}^{d}x_i^2\right)^{\lambda}$. Hence, if $\lambda$ is set to one (see Figure~\ref{fig:lambdaExp}), GNBG resembles the Sphere function.}:
\begin{align}
\label{eq:Unimodal1}
f(\vec{x})=  \left(\vec{x}^\top \vec{x}\right)^{\lambda}.
\end{align}

Now, we analyze the impact of $\lambda$ on the linearity of the component's basin. A linear basin is achieved with $\lambda=0.5$, sub-linear with $0<\lambda<0.5$, and super-linear with $\lambda>0.5$, as illustrated in Figure~\ref{fig:Lambda}.
 The value of $\lambda$ affects the rate at which the basin increases away from the center, decreasing or increasing the maximum function value when $\lambda$ is decreased or increased, respectively.

Next, we investigate the impact of elements of the principal diagonal of $\mathbf{H}$.
In equation \eqref{eq:Unimodal1}, the impact of $\mathbf{H}$ was eliminated by setting it to $\mathbf{I}_{d \times d}$.
By reintroducing $\mathbf{H}$ into equation \eqref{eq:Unimodal1}, GNBG becomes:
\begin{align}
\label{eq:Unimodal2}
f(\vec{x})=  \left(\vec{x}^\top \mathbf{H} \vec{x}\right)^{\lambda}.
\end{align}
$\mathbf{H}$ is a $d \times d$ diagonal matrix, i.e., $\mathbf{H} = \mathrm{diag}(h_1,h_2, \ldots,h_d)\in \mathbb{R}^{d \times d}$, where $h_i = \mathbf{H}(i,i)$.
The principal diagonal elements of $\mathbf{H}$ serve to scale the heights of the component's basin across different dimensions.
Equation \eqref{eq:Unimodal2} can be rewritten as $\left(\sum_{i=1}^{d}h_ix_i^2\right)^{\lambda}$, which indicates that the basin of the component along the $i$-th dimension is scaled by a factor of $h_i^\lambda$.

Furthermore, $\mathbf{H}$ influences the condition number of the component.
The condition number of $\mathbf{H}$ is defined as the ratio of its largest value to its smallest value among its principal diagonal elements, i.e., $\frac{\max_i |h_i|}{\min_i |h_i| }$.
The condition number of $\mathbf{H}$ directly affects the condition number of the component; however, its effect can be either amplified or dampened by the value of $\lambda$.
In Equation \eqref{eq:Unimodal2}, if $\lambda$ is set to one and each element $\mathbf{H}(i,i)$ is set to $10^{6\frac{i-1}{d-1}}$, GNBG resembles the Ellipsoidal function.

\begin{figure*}[!t]
\centering
\begin{tabular}{ccc}
    \subfigure[{\scriptsize $\mathbf{H}=\mathrm{diag}(1,1)$, condition number of $\mathbf{H}$ is 1.}]{\includegraphics[width=0.30\linewidth]{Figures/lambdaLog.eps}\label{fig:WellCon}}
      &
     \subfigure[{\scriptsize $\mathbf{H}=\mathrm{diag}(10,1)$, condition number of $\mathbf{H}$ is 10.}]{\includegraphics[width=0.30\linewidth]{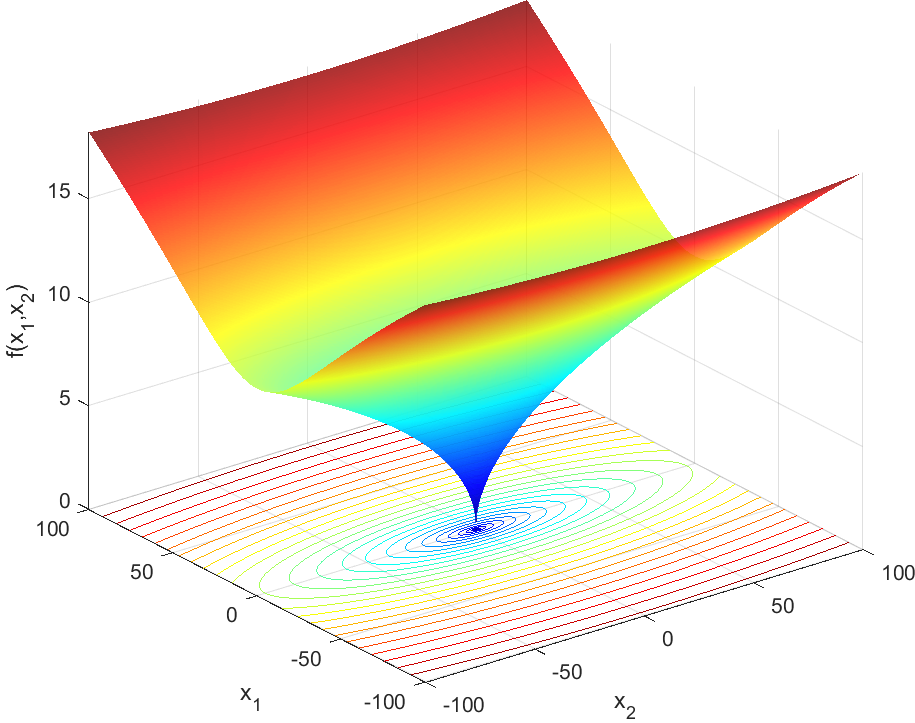}\label{fig:illCon1}}
        &
    \subfigure[{\scriptsize $\mathbf{H}=\mathrm{diag}(0.01,1)$, condition number of $\mathbf{H}$ is 100.}]{\includegraphics[width=0.30\linewidth]{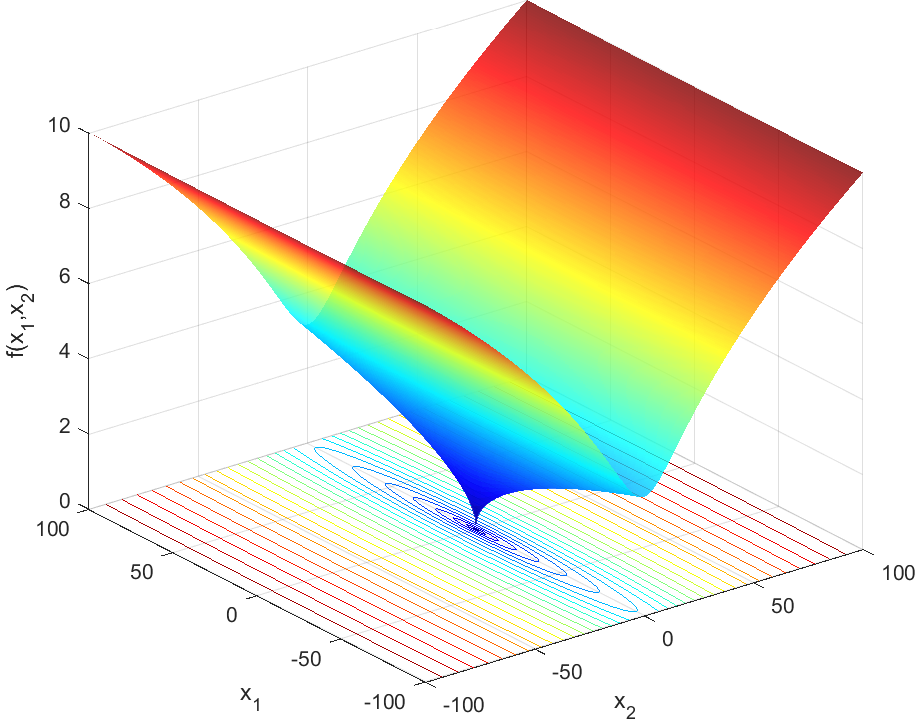}\label{fig:illCon2}}
\end{tabular}
\caption{Impact of $\mathbf{H}$ values on the morphology of a component generated by GNBG. 
For these illustrative examples, we set $d=2$, $o=1$, $\bm\mu=(0,0)$, $\bm\omega=(0,0,0,0)$, $\sigma=0$, $\vec{m}=(0,0)$, $\lambda=0.25$, and $\mathbf{R}=\mathbf{I}_{2 \times 2}$. 
Additionally, the 2-dimensional problem space is bounded to [-100,100] in each dimension.}
\label{fig:Himpact}
\end{figure*}

Figures~\ref{fig:WellCon}, \ref{fig:illCon1}, and \ref{fig:illCon2} illustrate how varying the values of $\mathbf{H}$ affects the characteristics of the component's basin. 
In Figure~\ref{fig:WellCon}, the principal diagonal elements of $\mathbf{H}$ have identical values, resulting in a well-conditioned basin. 
In contrast, the $\mathbf{H}$ matrices used to generate Figures~\ref{fig:illCon1} and \ref{fig:illCon2} are ill-conditioned, resulting in basins that are likewise ill-conditioned.

In Equation \eqref{eq:Unimodal2}, we assumed $\mathbf{R}=\mathbf{I}_{d\times d}$, so we could remove it from the baseline of GNBG.
By setting $\mathbf{R}$ to a non-identity orthogonal matrix, the equation changes to the following:
\begin{align}
\label{eq:Unimodal3}
f(\vec{x})=  \left((\mathbf{R}\vec{x})^\top  \mathbf{H} \mathbf{R}\vec{x}\right)^{\lambda}.
\end{align}
In GNBG, the rotation matrix $\mathbf{R}$ introduces complex variable interactions while preserving the component's original traits such as scale, height, and shape. 
$\mathbf{R}$ rotates the basin of the component around its center without altering its minimum position, affecting the variable interactions within the basin.

In GNBG, Givens rotation matrices ($\mathbf{G}$) are employed for their ability to allow targeted, customizable variable interactions. 
Unlike commonly used randomly generated orthogonal matrices, which usually affect the entire interaction structure, Givens matrices enable systematic adjustments between selected pairs of variables. 
This feature allows fine-tuning specific interactions without altering the entire structure, creating complex yet comprehensible optimization landscapes. 
Further details on constructing the rotation matrix $\mathbf{R}$ using Givens rotation matrices will be discussed later.

In cases where a component is not rotationally invariant--that is, it depends on the orientation of variables-- we can modify the interaction between variable pairs using Givens rotation matrices. A Givens rotation matrix in a two-dimensional space is expressed as:
\begin{align}
\label{eq:2DimGivens}
\mathbf{G} = \begin{pmatrix}
\cos(\theta) & -\sin(\theta) \\
\sin(\theta) & \cos(\theta)
\end{pmatrix}.
\end{align}
This matrix performs a rotation of angle $\theta$ in a plane defined by two coordinate axes. 
Extending this to higher dimensions, a Givens rotation matrix can selectively rotate variables within any two-dimensional subspace spanned by a pair of axes, while keeping all other dimensions unchanged. 
Therefore, to alter the variable interaction between each pair of variables $p$ and $q$, we can construct the matrix as follows:
\begin{align}
\label{eq:DdimensionalGivens}
\mathbf{G}[i,j]=    \begin{dcases}
    1 & \text{if   } i=j \wedge i,j\neq p \wedge i,j\neq q \\
    \cos(\theta) & \text{if   } i=j=p \vee  i=j=q\\
    -\sin(\theta) & \text{if   } i=p \wedge j=q\\
    \sin(\theta) & \text{if   } i=q \wedge j=p\\
    0 & \text{otherwise}  
\end{dcases}.
\end{align}
For example, consider a Givens rotation matrix designed to modify the interaction between the third and seventh variables in an 8-dimensional space. The matrix takes the form:
\begin{align}
\label{eq:GivensExample}
\mathbf{G} = \begin{pmatrix}
1 & 0 & 0 & 0 & 0 & 0 & 0 & 0 \\
0 & 1 & 0 & 0 & 0 & 0 & 0 & 0 \\
0 & 0 & \cos(\theta_{3,7}) & 0 & 0 & 0 & -\sin(\theta_{3,7}) & 0 \\
0 & 0 & 0 & 1 & 0 & 0 & 0 & 0 \\
0 & 0 & 0 & 0 & 1 & 0 & 0 & 0 \\
0 & 0 & 0 & 0 & 0 & 1 & 0 & 0 \\
0 & 0 & \sin(\theta_{3,7}) & 0 & 0 & 0 & \cos(\theta_{3,7}) & 0 \\
0 & 0 & 0 & 0 & 0 & 0 & 0 & 1
\end{pmatrix}
\end{align}
Setting $\theta_{p,q}=0$ keeps the Givens rotation matrix as the identity matrix $\mathbf{I}_{d \times d}$, which means the variable interaction between the  $p$-th and $q$-th variables remains unaltered. 
The choice of $\theta$ can be used to modulate the `strength' of interaction between each pair of variables.
Specifically, setting $\theta$ to values away from the main axes (i.e., $k\frac{\pi}{2}, k \in \mathbb{Z}$) leads to stronger interactions.
For instance, $\theta = \frac{\pi}{4}$ results in a stronger variable interaction compared to $\theta = \frac{\pi}{20}$.
Therefore, by setting $\theta_{p,q}$ to values that are not multiples of $\frac{\pi}{2}$ (i.e., non-axis-overlapped values), we establish interactions between variables $p$ and $q$.

To construct the complete rotation matrix $\mathbf{R}$, we first define an interaction matrix $\bm\Theta$ for each component. 
This $d \times d$ matrix has all elements on and below the principal diagonal set to zero. 
Each element above the diagonal in the $p$-th row and $q$-th column contains an angle, denoted as $\bm\Theta_k(p,q)$ (where $p < q$). 
This angle sets the extent of rotation applied to the projection of $\vec{x}$ in the basin of attraction of the $k$-th component onto the plane $x_p-x_q$. 
After defining $\bm\Theta_k$ for each component $k$, we employ Algorithm~\ref{alg:RotationControlled} to compute the rotation matrix $\mathbf{R}_k$.

\begin{algorithm}[!tp]
\footnotesize
$\mathbf{R}_{k}=\mathbf{I}_{d \times d}$\;
\For{$p=1$ to $d-1$}{\label{algline:Loop1}
\For{$q=p+1$ to $d$}{\label{algline:Loop2}
\If{$\bm\Theta_{k}(p,q)\neq 0$}{
$\mathbf{G}=\mathbf{I}_{d \times d}$\;
$\mathbf{G}(p,p)= \cos\left(\bm\Theta_{k}(p,q)\right)$;~\tcp*[h]{$\bm\Theta_{k}(a,b)$ and $\mathbf{G}(a,b)$ are the elements at $a$-th row and $b$-th column of matrices $\bm\Theta_{k}$ and $\mathbf{G}$, respectively.}\\
$\mathbf{G}(q,q) = \cos\left(\bm\Theta_{k}(p,q)\right)$\;
$\mathbf{G}(p,q) =  -\sin\left(\bm\Theta_{k}(p,q)\right)$\;
$\mathbf{G}(q,p) =  \sin\left(\bm\Theta_{k}(p,q)\right)$\;
$\mathbf{R}_{k} = \mathbf{R}_{k} \times \mathbf{G}$;~\tcp*[h]{$\mathbf{G}$ is the Givens rotation matrix for $x_{p}-x_{q}$ plane based on $\bm\Theta_{k}(p,q)$.}\\
}}}
Return $\mathbf{R}_{k}$\;
\caption{Pseudo code for calculating the rotation matrix $\mathbf{R}_{k}$ based on $\bm\Theta_k$.\\
{\footnotesize\textbf{Input}: $d$ and $\bm\Theta_k$\\
\textbf{Output}: {$\mathbf{R}_{k}$}}
}
\label{alg:RotationControlled}
\end{algorithm}

\begin{figure*}[!t]
\centering
\begin{adjustbox}{max width=0.93\textwidth}
\begin{tabular}{cc}
\centering
\begin{minipage}{0.4\textwidth}
\centering
\subfigure[{\scriptsize All angles are set to zero. This configuration does not alter any variable interaction.}]{\scriptsize 
$\bm\Theta = \begin{pmatrix}
0 & 0 & 0 & 0 & 0 & 0 & 0 & 0\\
0 & 0 & 0 & 0 & 0 & 0 & 0 & 0\\
0 & 0 & 0 & 0 & 0 & 0 & 0 & 0\\
0 & 0 & 0 & 0 & 0 & 0 & 0 & 0\\
0 & 0 & 0 & 0 & 0 & 0 & 0 & 0\\
0 & 0 & 0 & 0 & 0 & 0 & 0 & 0\\
0 & 0 & 0 & 0 & 0 & 0 & 0 & 0\\
0 & 0 & 0 & 0 & 0 & 0 & 0 & 0
\end{pmatrix}$     
 \label{fig:ThetaMatrix0}}
\end{minipage}
&
\begin{minipage}{0.45\textwidth}
\centering
\subfigure[{\scriptsize Fully disconnected variable interaction graph.}]{ \includegraphics[width=0.4\textwidth]{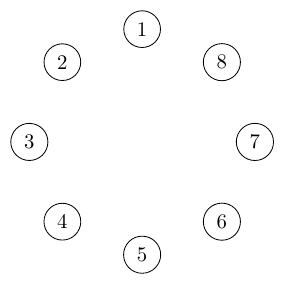}\label{fig:InteractionGraph0}}       
\end{minipage}
\\
\begin{minipage}{0.45\textwidth}
\centering
\subfigure[{\scriptsize All angles are set to values that are not a multiple of $\frac{\pi}{2}$, resulting in a fully non-separable fully-connected variable interaction structure.}]{\scriptsize
$\bm\Theta = \begin{pmatrix}
0 & \theta_{1,2} & \theta_{1,3} & \theta_{1,4} & \theta_{1,5} & \theta_{1,6} & \theta_{1,7} & \theta_{1,8} \\
0 & 0 & \theta_{2,3} & \theta_{2,4} & \theta_{2,5} & \theta_{2,6} & \theta_{2,7} & \theta_{2,8} \\
0 & 0 & 0 & \theta_{3,4} & \theta_{3,5} & \theta_{3,6} & \theta_{3,7} & \theta_{3,8} \\
0 & 0 & 0 & 0 & \theta_{4,5} & \theta_{4,6} & \theta_{4,7} & \theta_{4,8} \\
0 & 0 & 0 & 0 & 0 & \theta_{5,6} & \theta_{5,7} & \theta_{5,8} \\
0 & 0 & 0 & 0 & 0 & 0 & \theta_{6,7} & \theta_{6,8} \\
0 & 0 & 0 & 0 & 0 & 0 & 0 & \theta_{7,8} \\
0 & 0 & 0 & 0 & 0 & 0 & 0 & 0
\end{pmatrix}$     
 \label{fig:ThetaMatrix1}}
\end{minipage}
&
\begin{minipage}{0.45\textwidth}
\centering
\subfigure[{\scriptsize Fully connected variable interaction graph.}]{ \includegraphics[width=0.4\textwidth]{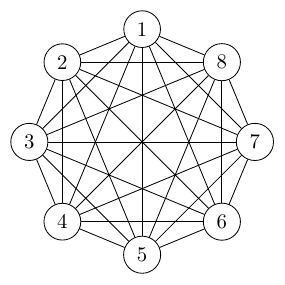}\label{fig:InteractionGraph1}}       
\end{minipage}
\\
\begin{minipage}{0.45\textwidth}
\centering
\subfigure[{\scriptsize A non-separable structure with the minimum number of variable interactions.
There is an interaction only between $i$-th and $(i+1)$-th variables.}]{\scriptsize
$\bm\Theta = \begin{pmatrix}
0 & \theta_{1,2} & 0 & 0 & 0 & 0 & 0 & 0 \\
0 & 0 & \theta_{2,3} & 0 &0& 0 &0 & 0 \\
0 & 0 & 0 & \theta_{3,4} & 0 & 0 & 0 & 0 \\
0 & 0 & 0 & 0 & \theta_{4,5} & 0&0 & 0 \\
0 & 0 & 0 & 0 & 0 & \theta_{5,6} & 0 & 0 \\
0 & 0 & 0 & 0 & 0 & 0 & \theta_{6,7} & 0 \\
0 & 0 & 0 & 0 & 0 & 0 & 0 & \theta_{7,8} \\
0 & 0 & 0 & 0 & 0 & 0 & 0 & 0
\end{pmatrix}$     
 \label{fig:ThetaMatrix2}}
\end{minipage}
&
\begin{minipage}{0.45\textwidth}
\centering
\subfigure[{\scriptsize A chain-like connected variable interaction graph.}]{ \includegraphics[width=0.4\textwidth]{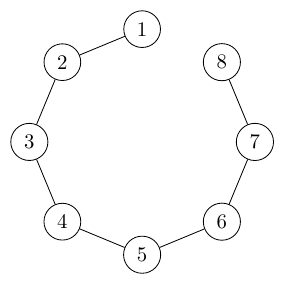}\label{fig:InteractionGraph2}}       
\end{minipage}
\\
\begin{minipage}{0.45\textwidth}
\centering
\subfigure[{\scriptsize $\bm\Theta$ is set to create a variable interaction structure containing two separable variables $x_3$ and $x_8$, a pair of connected variables $\{x_1,x_2\}$, and a group of connected (not fully-connected) variables $\{x_4,x_5,x_6,x_7\}$. }]{\scriptsize
$\bm\Theta = \begin{pmatrix}
0 & \theta_{1,2} &0 & 0 & 0 & 0 & 0 & 0 \\
0 & 0 & 0 & 0 & 0& 0 & 0 & 0 \\
0 & 0 & 0 &0 &0 & 0 & 0 & 0 \\
0 & 0 & 0 & 0 & \theta_{4,5} & \theta_{4,6} &\theta_{4,7} & 0 \\
0 & 0 & 0 & 0 & 0 & \theta_{5,6} & 0 & 0 \\
0 & 0 & 0 & 0 & 0 & 0  & \theta_{6,7} & 0\\
0 & 0 & 0 & 0 & 0 & 0 & 0 & 0 \\
0 & 0 & 0 & 0 & 0 & 0 & 0 & 0
\end{pmatrix}$     
 \label{fig:ThetaMatrix4}}
\end{minipage}
&
\begin{minipage}{0.45\textwidth}
\centering
\subfigure[{\scriptsize A disconnected variable interaction graph with two groups of variables and two separable variables.}]{ \includegraphics[width=0.4\textwidth]{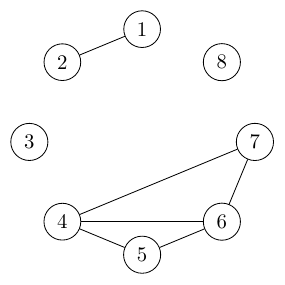}\label{fig:InteractionGraph4}}       
\end{minipage}
\end{tabular}
\end{adjustbox}
\caption{
Examples of how, by configuring $\bm\Theta$, GNBG can generate desired variable interaction structures in an 8-dimensional component. 
We assume that the component is initially fully-separable and rotation-dependent. 
The variable interaction graphs associated with each matrix $\bm\Theta$ are illustrated in the right column.
}
\label{fig:interactionStructure}
\end{figure*}

Using Algorithm~\ref{alg:RotationControlled}, we can generate various types of variable interaction structures, ranging from different degrees of separability, depending on the angles included in $\bm\Theta_k$.
It should be noted that $\mathbf{R}_k$ can be used to alter variable interaction when the $k$-th component is rotation-dependent\footnote{GNBG is capable of generating rotation-invariant components.}.
In Figure~\ref{fig:interactionStructure}, we demonstrate several examples illustrating how different configurations of $\bm\Theta$ can be used to generate various variable interaction structures within an 8-dimensional component.

Besides manually configuring $\bm\Theta$ for specific desired variable interaction structures, GNBG also allows for the generation of components with random variable interaction structures and strengths. 
To achieve this, we introduce a parameter $\mathfrak{p}$ to represent the probability of each element above the principal diagonal in $\bm\Theta$ being either zero or a randomly generated angle. 
For each $\bm\Theta_{k}(p,q)$, a random number is generated from a uniform distribution. 
If this number is less than or equal to $\mathfrak{p}$, then $\bm\Theta_{k}(p,q)$ is set to zero ($q > p$). 
Otherwise, it is set to a random angle in the range $(0, 2\pi)$. 
Smaller values of $\mathfrak{p}$ result in variable interaction structures with fewer connections between variables, while larger values generate more complex structures with increased connectivity. 
Setting $\mathfrak{p}$ to zero or one yields fully separable and fully connected variable interaction structures, respectively. 
Users may also choose to randomly assign angles from a predefined set of values based on $\mathfrak{p}$ rather than generating them in a continuous range.

\begin{figure*}[!t]
\centering
\begin{adjustbox}{max width=\textwidth, max height=\textheight}
\begin{tabular}{ccc}
    \subfigure[{\scriptsize $\bm\Theta(1,2)=0$}]{\includegraphics[width=0.30\linewidth]{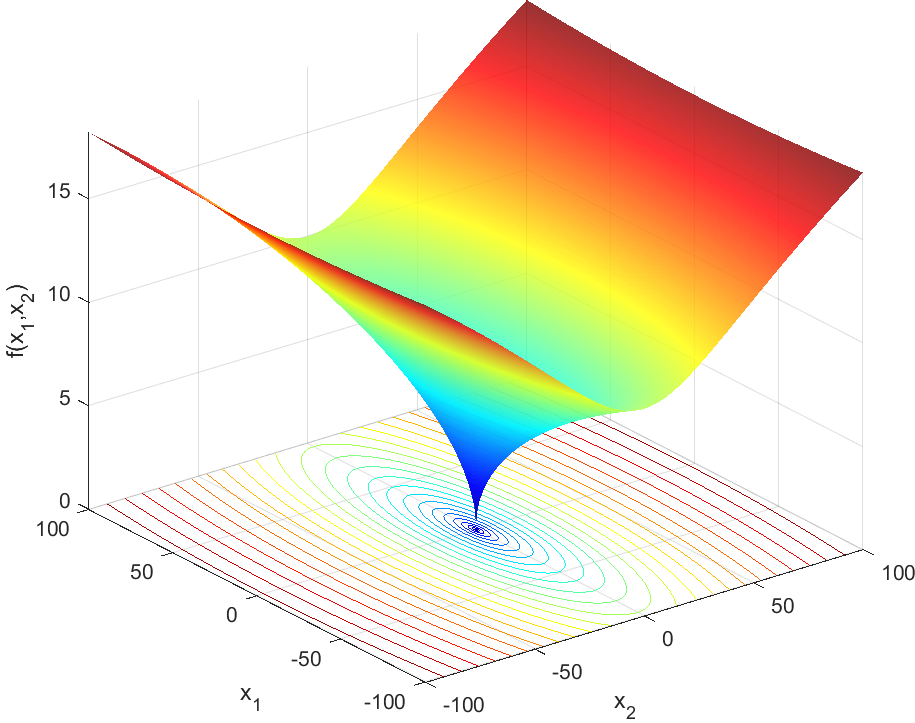}\label{fig:angle0}}
    &
    \subfigure[{\scriptsize $\bm\Theta(1,2)=\frac{\pi}{4}$}]{\includegraphics[width=0.30\linewidth]{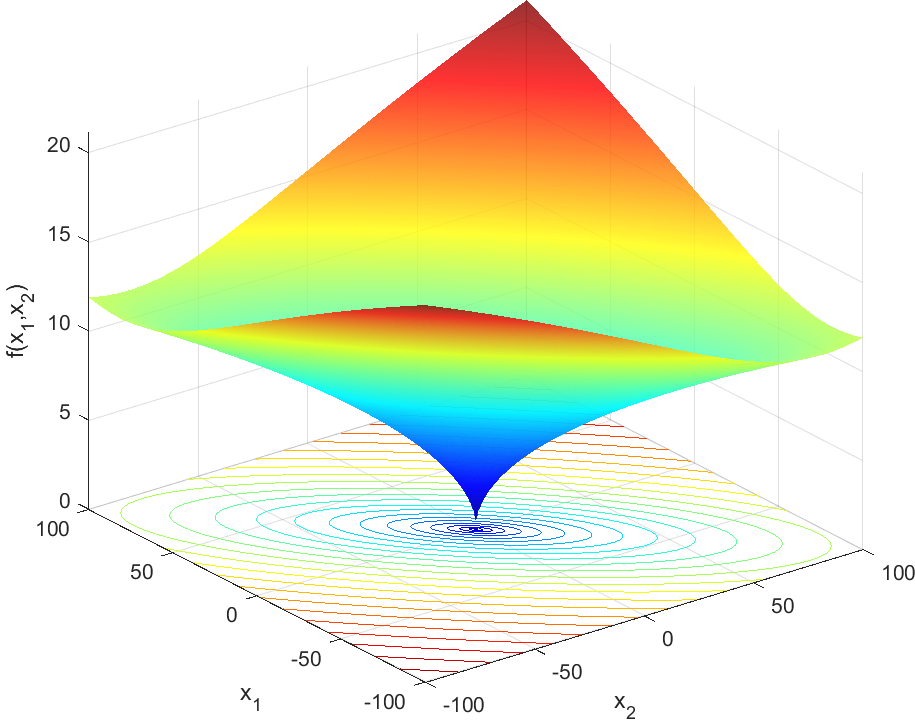}\label{fig:angle45}}
     &
     \subfigure[{\scriptsize $\bm\Theta(1,2)=\frac{5\pi}{12}$}]{\includegraphics[width=0.30\linewidth]{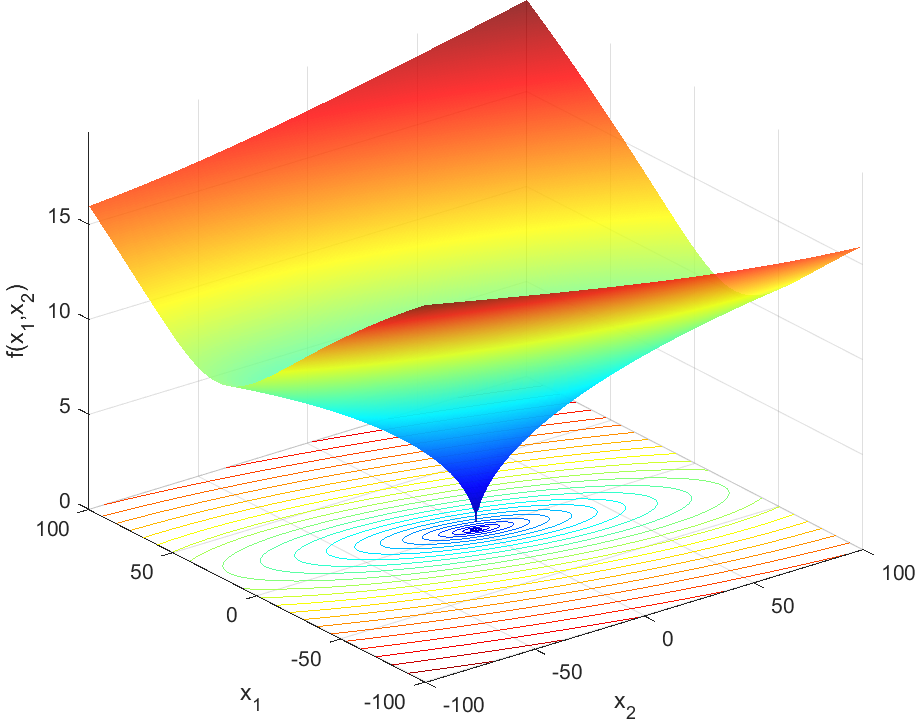}\label{fig:angle90}}
\end{tabular}
\end{adjustbox}
\caption{
Impact of rotating the projection of $\vec{x}$ in the basin of a component onto the $x_{1}–x_{2}$ plane with different angles $\bm\Theta(1,2)$.
For generating these illustrative examples, we set $d=2$, $o=1$, $\bm\mu=(0,0)$, $\bm\omega=(0,0,0,0)$, $\sigma=0$, $\vec{m}=(0,0)$, $\lambda=0.25$, $\mathbf{H}=\mathrm{diag}(1,10)$, and $\mathbf{R}$ is obtained by Algorithm~\ref{alg:RotationControlled} based on the given angle $\bm\Theta(1,2)$. 
Additionally, the 2-dimensional problem space is bounded to [-100,100] in each dimension.
}
\label{fig:rotation}
\end{figure*}

Figure~\ref{fig:rotation} illustrates how rotation affects the basin of attraction for a given component when $\mathbf{R}$ is generated using Algorithm~\ref{alg:RotationControlled} with a user-defined angle. 
To facilitate visualization, we set $d = 2$. 
In this case, there is only a single plane $x_1-x_2$, and the output of Algorithm~\ref{alg:RotationControlled} corresponds to equation \eqref{eq:2DimGivens}.

Note that based on our investigation using differential grouping~\cite{omidvar2017dg2} and experimentally optimizing each dimension separately, a component generated by GNBG is fully separable if $\lambda$ equals one, regardless of other parameters, when the rotation matrix $\mathbf{R}$ is set to the identity matrix. 
In this situation, the component can be optimized independently in each dimension. 
Setting $\lambda$ to values other than one makes the variable interaction structure additively non-separable, as shown by differential grouping in our investigations. 
However, we have provided a mathematical proof in Section~\ref{sec:sup:proof} of the supplementary documents that it can still be optimized dimension-wise. 
Therefore, when $\mathbf{R}$ is the identity matrix, the component generated by GNBG can be optimized in each dimension separately, regardless of other parameters. 
Adjusting the rotation matrix $\mathbf{R}$ introduces interactions that make targeted dimensions non-separable, requiring joint optimization of those variables.

\begin{figure*}[!t]
\centering
\begin{tabular}{ccc}
    \subfigure[{\scriptsize $\bm\mu=(0.2,0.2)$, $\bm\omega=(10,10,10,10)$.}]{\includegraphics[width=0.30\linewidth]{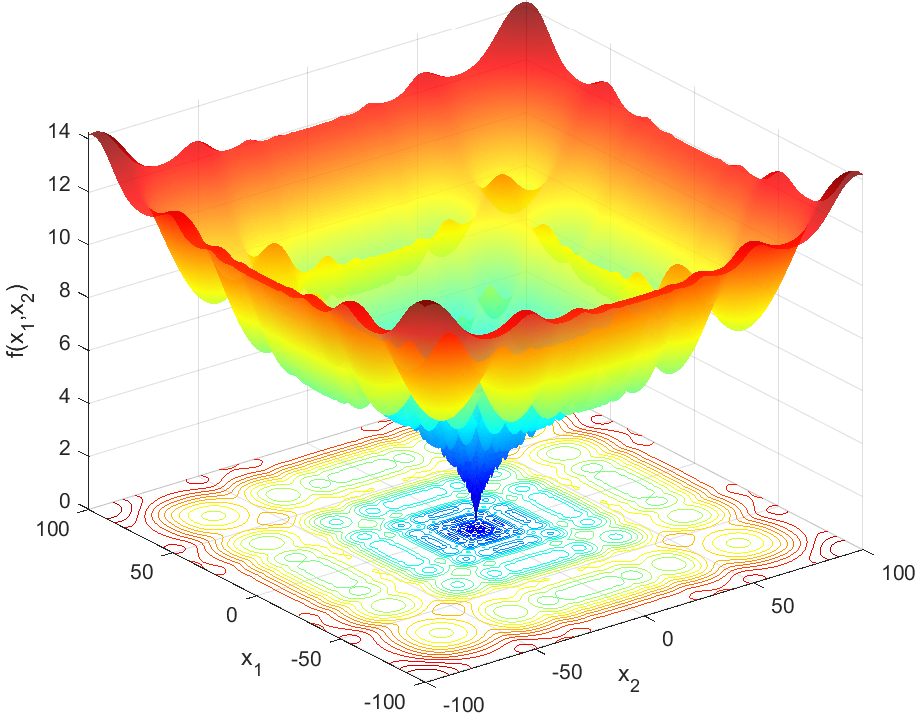}\label{fig:Beta10}}
     &
     \subfigure[{\scriptsize $\bm\mu=(0.2,0.2)$, $\bm\omega=(20,20,20,20)$.}]{\includegraphics[width=0.30\linewidth]{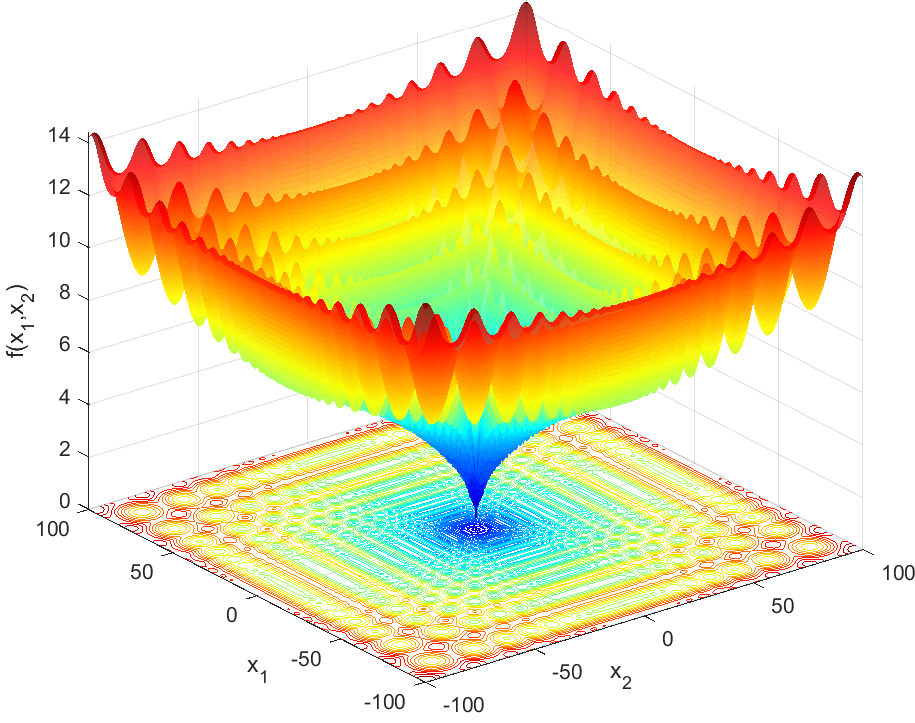}\label{fig:Beta20}}
&
\subfigure[{\scriptsize $\bm\mu=(0.2,0.2)$, $\bm\omega=(50,50,50,50)$.}]{\includegraphics[width=0.30\linewidth]{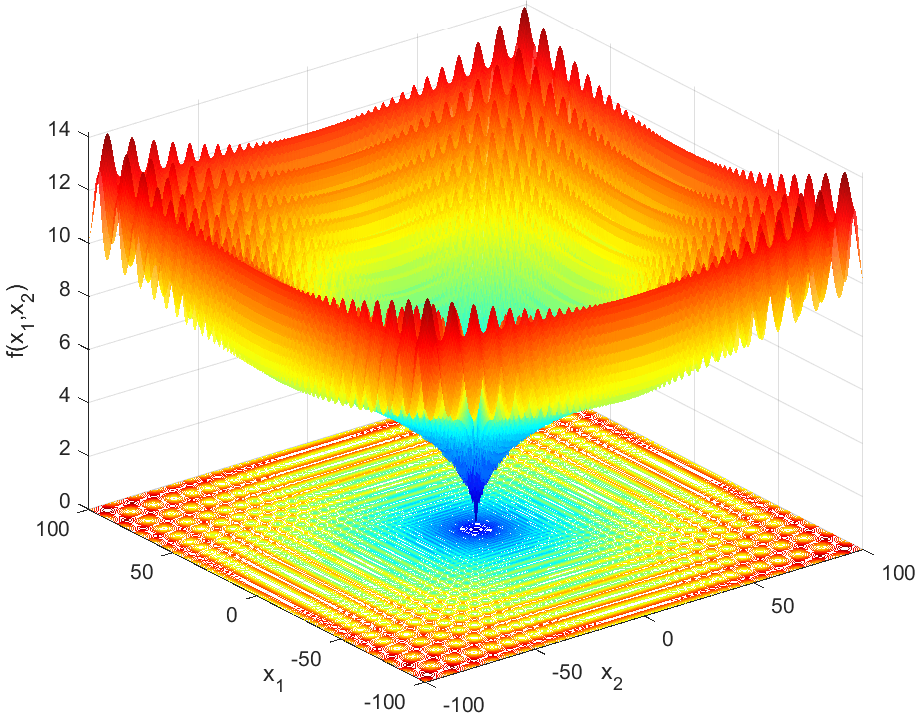}\label{fig:Beta50}}
\\
 \subfigure[{\scriptsize $\bm\mu=(0.1,0.1)$, $\bm\omega=(10,10,10,10)$.}]{\includegraphics[width=0.30\linewidth]{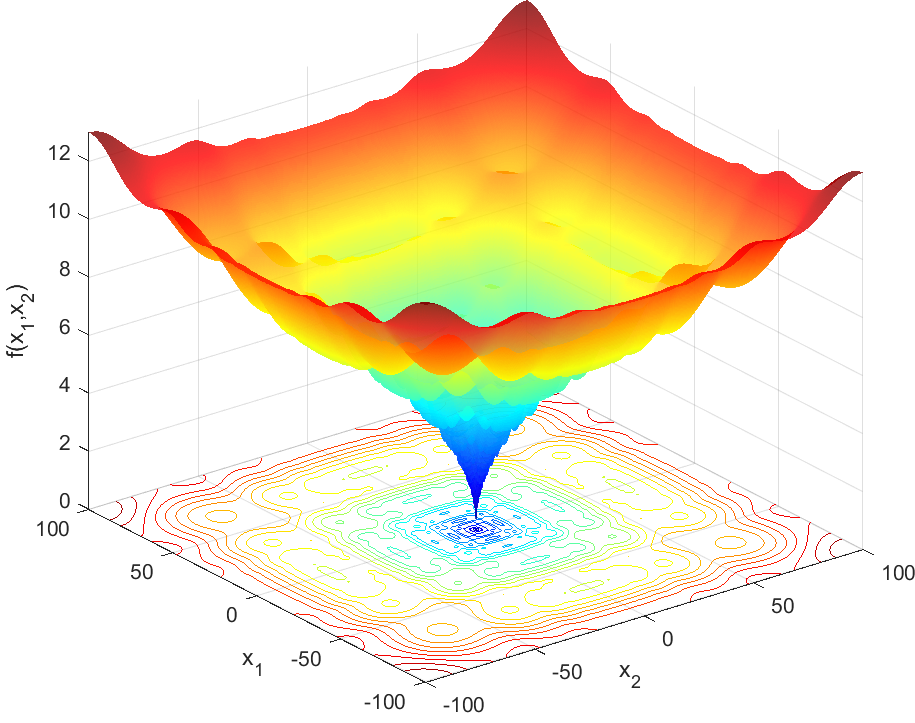}\label{fig:Alpha0point2}}
  &
\subfigure[{\scriptsize $\bm\mu=(0.5,0.5)$, $\bm\omega=(10,10,10,10)$.}]{\includegraphics[width=0.30\linewidth]{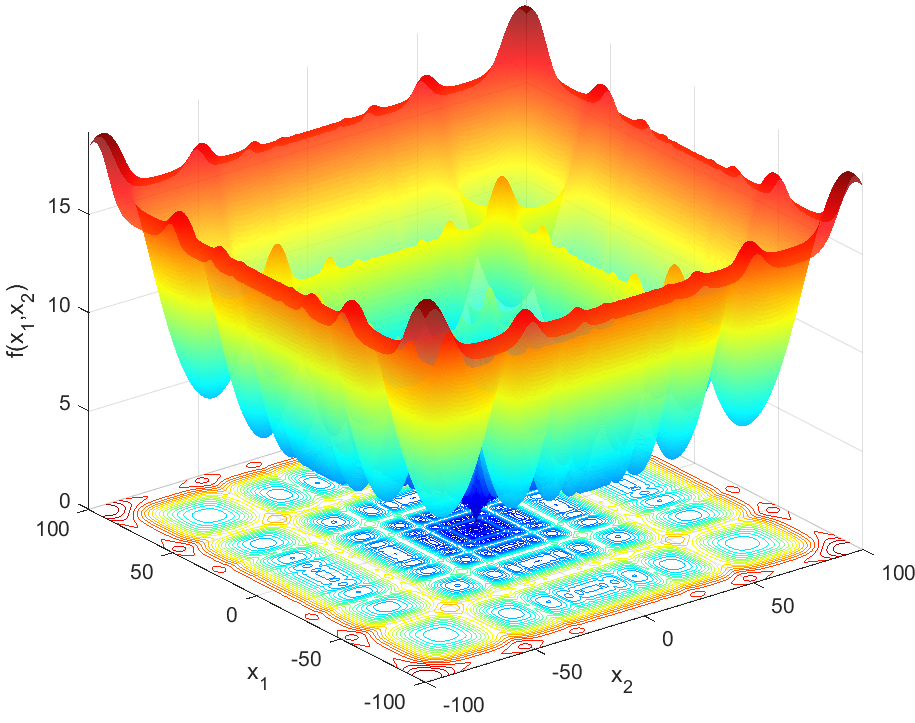}\label{fig:Alpha1}}
  &
\subfigure[{\scriptsize $\bm\mu=(1,1)$, $\bm\omega=(10,10,10,10)$.}]{\includegraphics[width=0.30\linewidth]{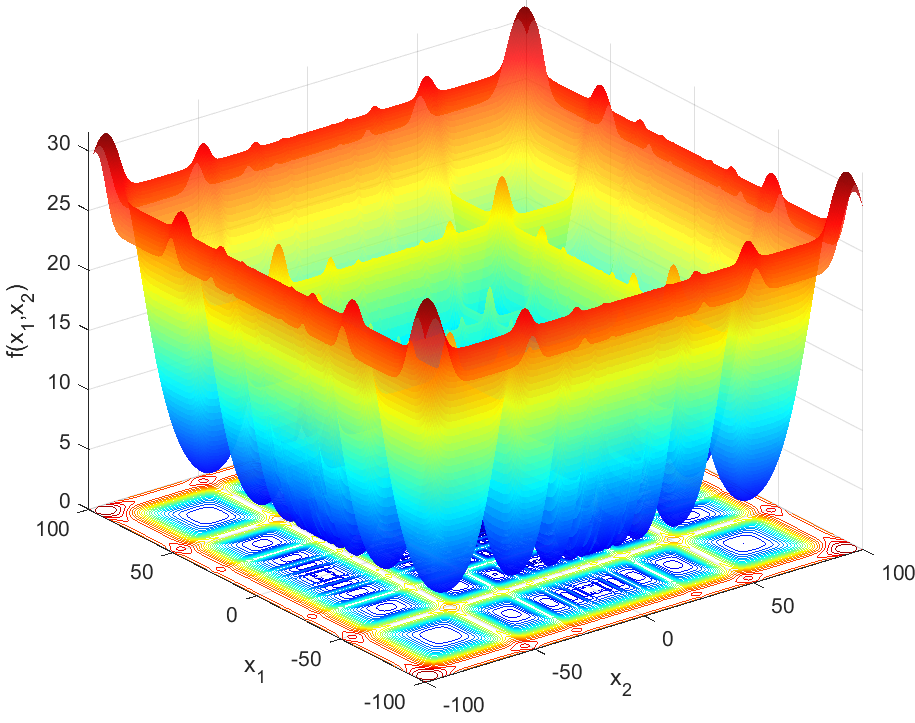}\label{fig:Alpha1point5}}
\\
 \subfigure[{\scriptsize $\bm\mu=(0.5,0.2)$, $\bm\omega=(10,10,10,10)$.}]{\includegraphics[width=0.30\linewidth]{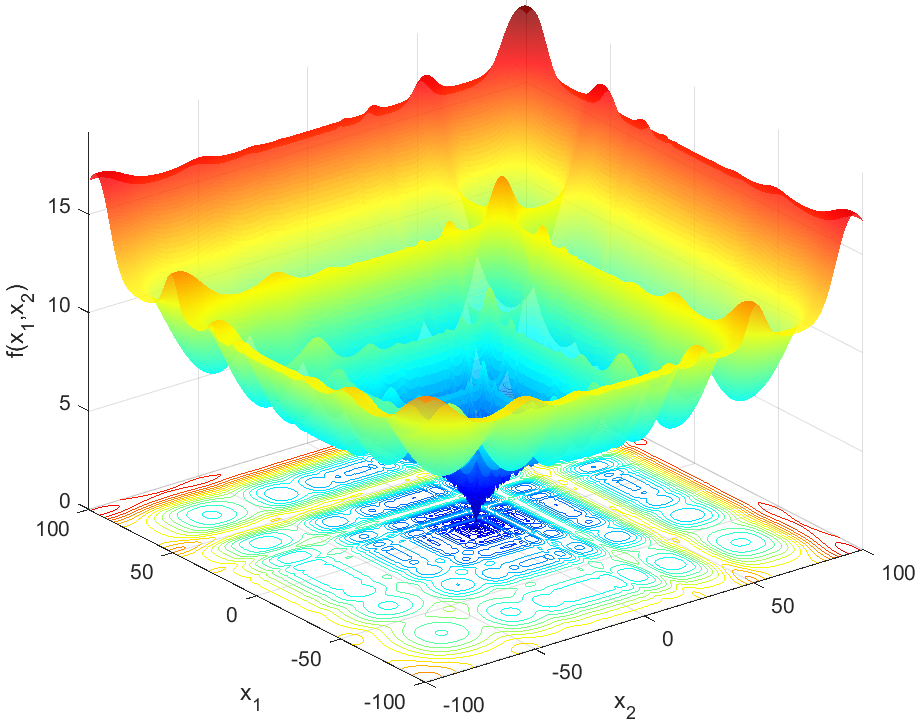}\label{fig:VariousAlpha}}
 &
 \subfigure[{\scriptsize $\bm\mu=(0.2,0.2)$, $\bm\omega=(50,10,20,1)$.}]{\includegraphics[width=0.30\linewidth]{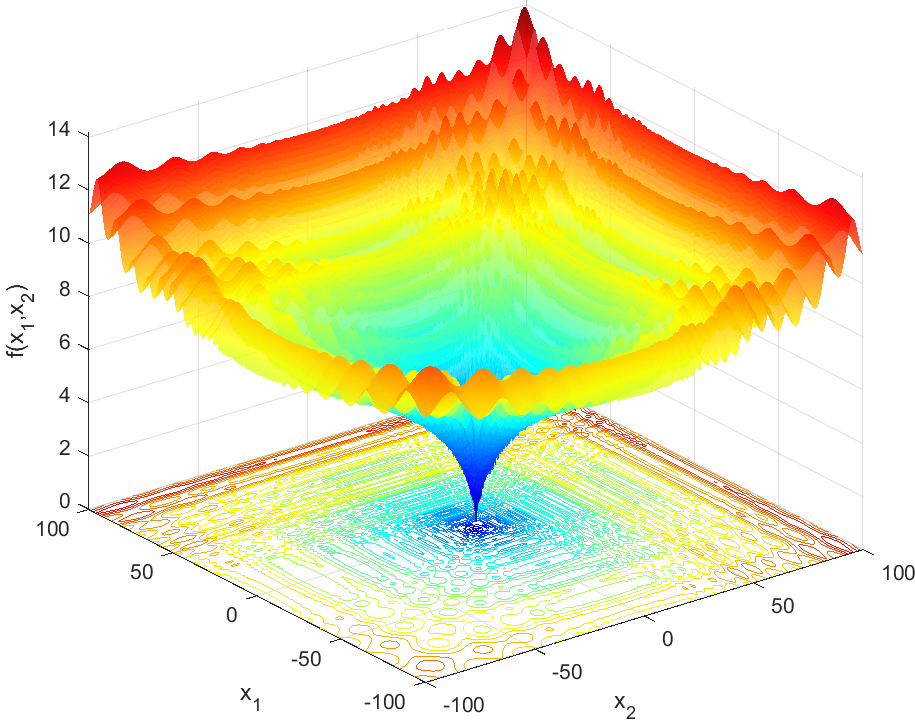}\label{fig:VariousBeta}}
 &
  \subfigure[{\scriptsize  $\bm\mu=(0.5,0.2)$, $\bm\omega=(50,10,20,1)$.}]{\includegraphics[width=0.30\linewidth]{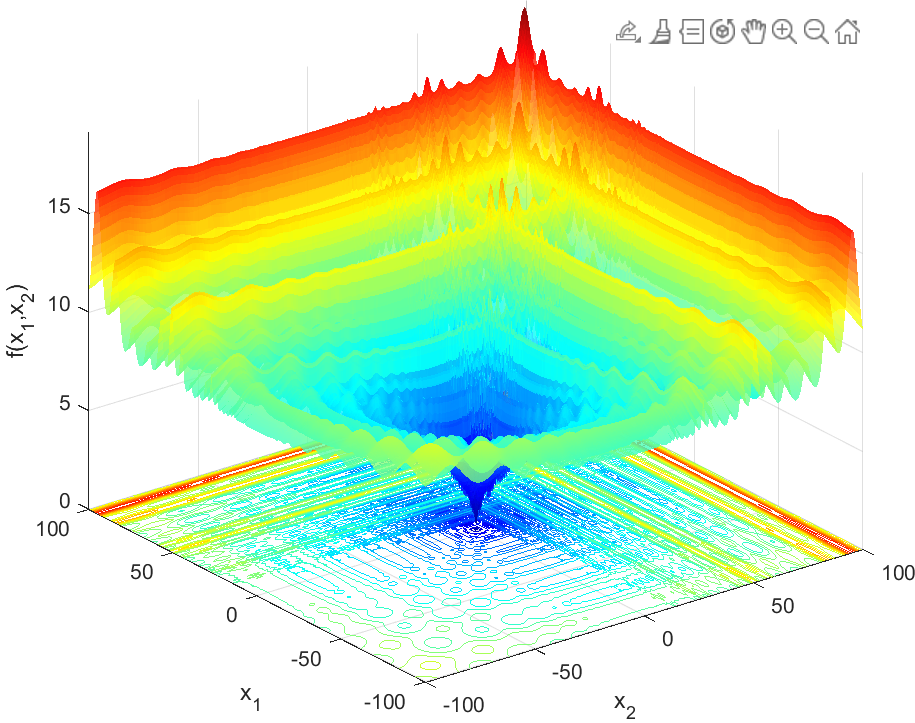}\label{fig:VariousAlphaBeta}}
\end{tabular}
\caption{Impact of $\bm\mu$ and $\bm\omega$ values on a 2-dimensional component generated by GNBG.
 For these illustrative examples, the 2-dimensional problem space is bounded to [-100,100] in each dimension, $o=1$, $\sigma=0$, $\vec{m}=(0,0)$, $\lambda=0.25$, $\mathbf{H}=\mathrm{diag}(1,1)$, and $\mathbf{R}=\mathbf{I}_{2 \times 2}$.
This configuration focuses on the transformation impact and simplifies GNBG to $f(\vec{x}) = \left( \mathbb{T}(\vec{x}^\top) \mathbb{T} (\vec{x}) \right)^{\lambda}$. 
 }
\label{fig:Multimodal}
\end{figure*}

Now we analyze the influence of the $\bm\mu$ and $\bm\omega$ parameters in the transformation $\mathbb{T}$.
By setting $\bm\mu$ and $\bm\omega$ to non-zero values, the transformation $\mathbb{T}$ in equation~\eqref{eq:ir} is enabled and reintroduced to Equation~\eqref{eq:Unimodal3}, which changes the formulation to:
\begin{align}
\label{eq:Multimodal1}
f(\vec{x})=  \left(  \mathbb{T}\left((\mathbf{R}\vec{x})^\top\right)  \mathbf{H} \mathbb{T} \left(\mathbf{R}\vec{x}\right)  \right)^{\lambda}.
\end{align}
By setting the elements of $\bm\mu$ and $\bm\omega$ to non-zero values, we induce the formation of local optima within the component's basin. 
In this transformation, sinusoidal functions are utilized to create irregularities and local optima, and the values of $\bm\mu$ and $\bm\omega$ determine the size, morphology, and symmetry of these local optima. 
Specifically:
\begin{itemize}
    \item The $\bm\mu$ parameters control the amplitude of the sinusoidal components, which affect the depth of local optima. 
    Different values of $\mu_{k,1}$ and $\mu_{k,2}$ introduce asymmetric non-linearity.
    \item The $\bm\omega$ parameters dictate the frequency of the sinusoidal functions, which affect the number and vastness of the local optima. 
Asymmetric patterns in the basins can be introduced through different settings for  $\omega_1, \omega_2, \omega_3,$ and $\omega_4$.
\end{itemize}
Consequently, by adjusting $\bm\mu$ and $\bm\omega$, the characteristics of the transformed basin can be manipulated.
Figure~\ref{fig:Multimodal} consists of nine plots that illustrate the impact of $\bm\mu$ and $\bm\omega$ on the local optima in the basin. 
From these plots, we observe that increasing $\bm\omega$ values leads to an increase in the number of local optima within the basin, while also reducing their width.
In addition, we observe that while $\bm\mu$ impacts the amplitude and thus the depth of the local optima---without affecting their number or width---it is worth noting that $\bm\omega$ exclusively influences the frequency and spatial distribution of these optima. 
Finally, it can be seen that different values for $\bm\mu$ elements and/or $\bm\omega$ elements  result in asymmetry in the basin.

Up to this point, we have set $\sigma = 0$ and $\vec{m} = \{ m_i = 0 \,|\, i = 1, 2, \ldots, d \}$ to neutralize their impact.
This choice resulted in positioning the minimum (or base) of the component at $[0, 0, \ldots, 0] \in \mathbb{R}^{1 \times d}$ and making its minimum function value equal to zero.
In GNBG, the $\sigma_k$ and $\vec{m}_k$ parameters can be utilized to specify the minimum function value and the position of the $k$-th component, respectively.
Incorporating $\sigma$ and $\vec{m}$ into equation \eqref{eq:Multimodal1}, we get:
\begin{align}
\label{eq:Multimodal2}
f(\vec{x})=  \sigma + \left(\mathbb{T}\left(\mathbf{R}(\vec{x}-\vec{m}))^\top\right)  \mathbf{H} \mathbb{T}\left(\mathbf{R}(\vec{x}-\vec{m})\right)\right)^{\lambda},
\end{align}
which represents the complete form of GNBG for generating a single component. 
Here, $\sigma_k$ and $\vec{m}_k$ serve to translate (or shift) the minimum of the $k$-th component in the objective space and the solution space, respectively.
By manipulating $\vec{m}_{k}$ values, users can precisely define the minimum positions of the components.
In GNBG, $f(\vec{m}_k) = \sigma_k$ determines the minimum value of the $k$-th component.

The next parameter of GNBG is $o$, which defines the number of components in the search space.
By setting $o = 1$, we removed the impact of this parameter.
By setting the value of $o$ to be larger than one, equation~\eqref{eq:Multimodal2} extends to \eqref{eq:irGMPB}, resulting in a problem space that includes multiple components, each with its own basin defined by the $\mathrm{min}(\cdot)$ function in \eqref{eq:irGMPB}. 
Each component $k$ in GNBG has its own parameter settings, and the only parameter that all components share is the dimension.
The minimum position of the component with the smallest $\sigma$ value corresponds to the global optimum position.
Furthermore, the objective value of the global optimum is equal to the smallest $\sigma$ value among all components.

Note that increasing the number of components directly impacts the computational complexity of function evaluations. 
For instance, evaluating the objective value of a candidate solution in a 30-dimensional problem instance with 10 components takes approximately twice as long as an objective function evaluation in a 30-dimensional problem instance with five components.

\begin{figure*}[!t]
\centering
\begin{tabular}{ccc}
    \subfigure[{\scriptsize $o=2$}]{\includegraphics[width=0.30\linewidth]{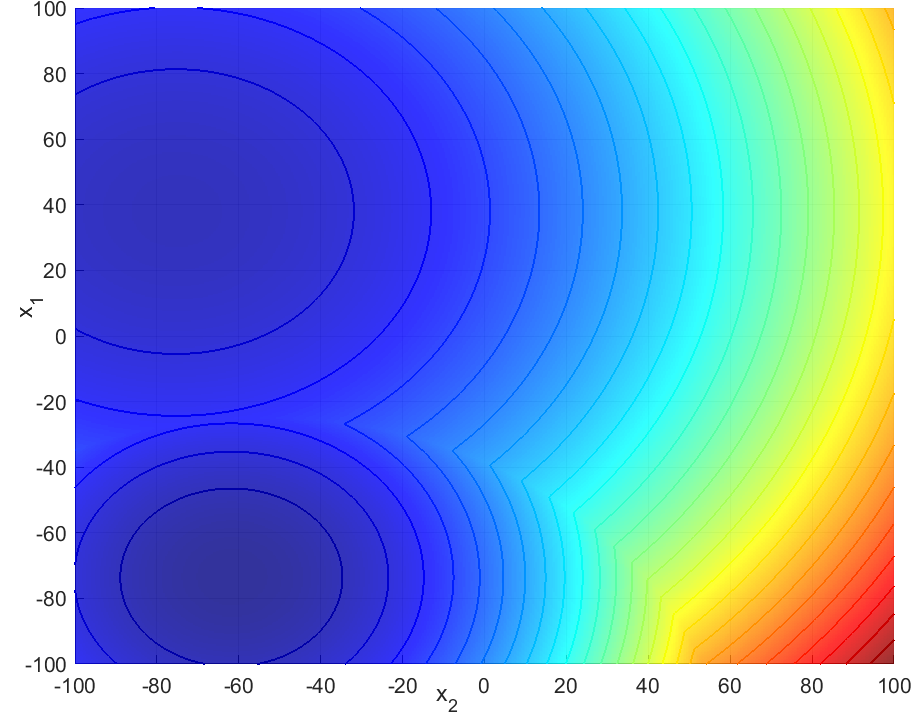}\label{fig:o2}}
&  \subfigure[{\scriptsize $o=10$}]{\includegraphics[width=0.30\linewidth]{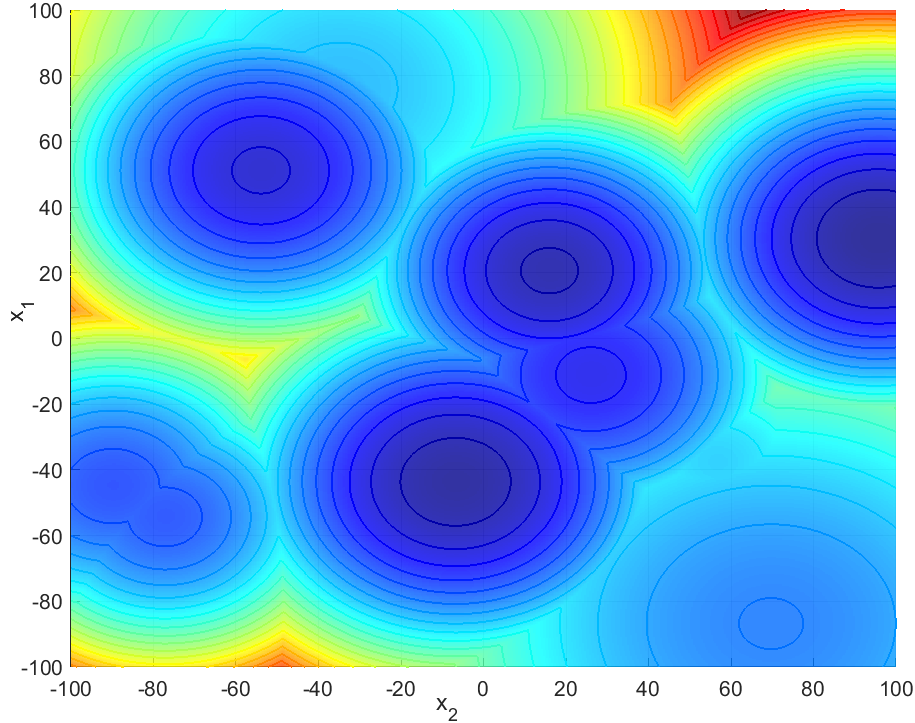}\label{fig:o10}}
&  \subfigure[{\scriptsize $o=25$}]{\includegraphics[width=0.30\linewidth]{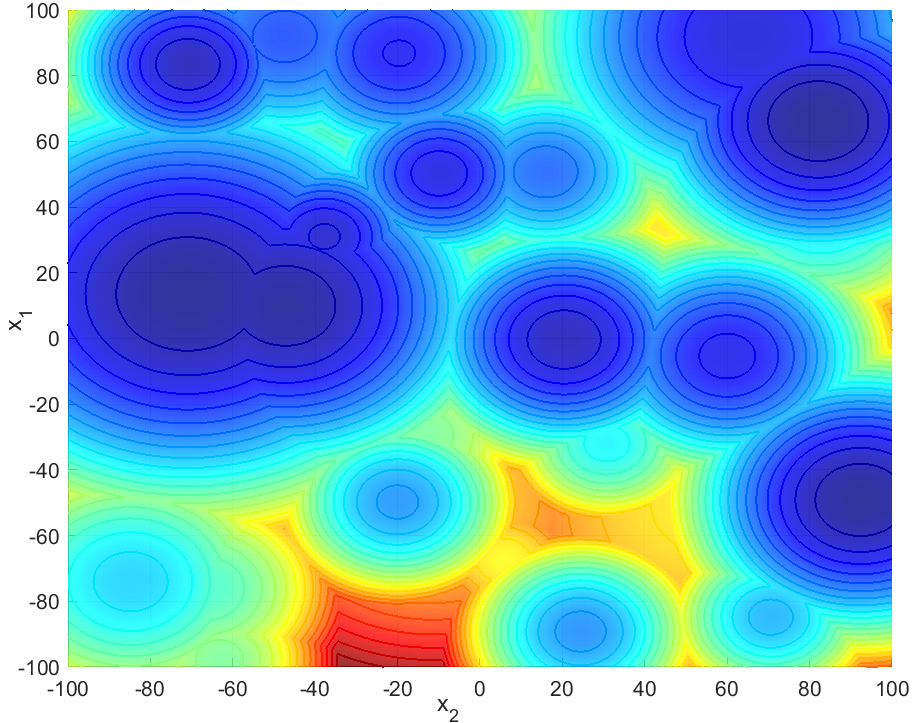}\label{fig:o25}}
\end{tabular}
\caption{
Solution spaces produced by GNBG with multiple components. 
To highlight the effects of using multiple components without the interference of other morphological traits, we employed simplified component forms: we set $\lambda_k$ to one, $\mathbf{R}$ as the identity matrix, ensured well-conditioned components by setting $\mathbf{H}_k(1,1) = \mathbf{H}_k(2,2)$, and neutralized $\mathbb{T}$ by setting all elements of $\bm\mu_k$ and $\bm\omega_k$ to zero. 
Moreover, we randomly selected the $\sigma_k$ values from the interval [0,10], $\mathbf{H}_k$ values from [0.001,0.01], and the minimum position for each component, $\vec{m}_k$, from [-100,100].
}
\label{fig:MultiComponents}
\end{figure*}

Figure~\ref{fig:MultiComponents} illustrates three problem instances each possessing multiple components. 
An important observation is that the number of discernible components in Figures~\ref{fig:o10} and~\ref{fig:o25} is less than the specified values of $o$ for each landscape. 
This discrepancy is a significant factor to consider when designing problem instances with multiple components, especially when some of parameters are generated randomly. 
There exists the potential for some components to be overshadowed or `dominated' by the basins of larger components. 
By `dominated', we imply that these components might not significantly influence the search space but might add to the computational complexity. 
A strategy to pinpoint such dominated components is by evaluating the function value at each component's minimum position. 
If the condition $f(\vec{m}_k) < \sigma_k$ holds for the $k$-th component, it signals that the component is ensnared and dominated by other components' basins.

Finally, we discuss the role of the dimensionality parameter $d$ in GNBG.
As $d$ increases, the complexity of the search space grows, potentially intensifying the ``curse of dimensionality'' and leading to the ineffectiveness of conventional optimization algorithms~\cite{omidvar2021reviewA,omidvar2021reviewB}.

Section~\ref{sec:sup:GNBGparameters} in the supplementary document summarizes the parameters of GNBG.

\section{Problem Instance Generation by GMPB}
\label{sec:ProblemInstances}

In this section, we explain how to utilize GNBG's configurability and flexibility to generate problem instances for specific research objectives. 
By adjusting GNBG's controllable parameters, researchers can create instances for detailed analysis of optimization algorithm performance. 
This approach reveals an algorithm's strengths and weaknesses against different problem characteristics, each of which can be adjusted to varying degrees. 
We will discuss how to fine-tune GNBG's parameters to produce instances with diverse basin linearity, conditioning, variable interaction structures, multi-modality, deceptiveness through multiple competitive components, and various combinations of these characteristics.

\subsection{Exploring Basin Linearity}
\label{sub:sub:linearity}

In GNBG, the parameter $\lambda_k$ determines the linearity of the basin for the $k$-th component. 
To examine how variations in basin linearity affect optimization algorithm performance, it is important to systematically explore different $\lambda$ values. 
To focus solely on basin linearity, we must eliminate other influencing factors like multimodality, variable interaction, and ill-conditioning. 
To achieve this, we set the number of components $o$ to one, creating a single-component landscape. 
Additionally, we neutralize the transformation $\mathbb{T}$ by setting all elements of vectors $\bm{\mu}$ and $\bm{\omega}$ to zero. 
Both $\mathbf{H}$ and $\mathbf{R}$ are set as identity matrices to produce well-conditioned and unrotated components.

With these settings in place, $\lambda$ can be adjusted to create problem instances that highlight the impact of basin linearity on algorithm performance. 
Values of $\lambda$ less than 0.5 yield sub-linear basins, where the curvature around the optimal solution becomes increasingly narrow as $\lambda$ decreases. 
A $\lambda$ value of 0.5 results in a linear basin, while values greater than 0.5 create super-linear basins.
To thoroughly investigate the influence of basin linearity, one can set $\lambda$ from a discrete set such as ${0.1, 0.25, 0.5, 0.75, 1}$. 
This approach facilitates a comparative analysis, providing insights into the strengths and weaknesses of optimization algorithms in different basin conditions.

\subsection{Investigating various conditioning}
\label{sub:sub:conditioning}

Ill-conditioning is a characteristic frequently encountered in real-world optimization problems. 
Traditional approaches to studying algorithm robustness to ill-conditioning often rely on baseline functions with fixed condition numbers, such as the Ellipsoid function~\cite{hansen2009real}. 
GNBG offers a more flexible alternative: by manipulating the condition number of $\mathbf{H}_k$, users can control the condition number of the $k$-th component. 

To focus exclusively on the impact of ill-conditioning on the performance of algorithms, we neutralize other challenging characteristics. 
Specifically, we set the number of components $o$ to one and neutralize the transformation $\mathbb{T}$ to create a unimodal landscape.
Additionally, we set $\lambda=1$ and $\mathbf{R} = \mathbf{I}$.

For a comprehensive study, users can vary the condition number of $\mathbf{H}$ across a wide range, such as $\{10, 10^2,\ldots, 10^7\}$. 
To achieve specific condition number $c$, two randomly chosen elements of the diagonal of $\mathbf{H}$ are set to $a$ and $b$ such that $b > a$ and $\frac{b}{a} = c$. 
Remaining diagonal elements are randomly sampled from the range $[a, b]$ according to a Beta distribution\footnote{Here, we set $0 < \alpha = \beta \leq 1$ for Beta distribution, where smaller $\alpha$ and $\beta$ values increase the likelihood of generating numbers closer to $a$ or $b$.}.

\subsection{Exploring Variable Interaction Structures}
\label{sub:sub:interactions}

Real-world optimization problems present a variety of variable interaction structures, ranging from fully separable to fully connected configurations~\cite{omidvar2017dg2}. 
GNBG allows for flexibility in generating these structures for each component $k$ by adjusting the $\bm\Theta_k$ matrix. 
Users can either manually configure $\bm\Theta_k$ or use an alternative setting based on random generation to achieve different degrees of separability and connection strengths, controlled by the parameter $\mathfrak{p}_k$ in the range $[0,1]$. 
A $\mathfrak{p}_k$ value of 0 maintains the original variable interaction structure, while a value of 1 creates a fully connected structure. 
Intermediate values of $\mathfrak{p}_k$ produce components that are either partially separable or fully non-separable but not fully connected.

To isolate the influence of variable interaction structures on algorithmic performance, other factors must be controlled.
We set the number of components $o$ to one and vectors $\bm{\mu}$ and $\bm{\omega}$ to zero to generate a unimodal landscape.
The diagonal elements of $\mathbf{H}$ are randomly generated from the uniform distribution in the range $[1,100]$, resulting in a maximum condition number of 100---a value that has a negligible  impact on most algorithms (see Table~\ref{tab:ConditioningResults}).
Users can evaluate performance by setting $\mathfrak{p}_k$ values across a range such as ${0, 0.25, 0.5, 0.75, 1}$.
Angles between variable pairs are randomly selected from the range $[- \pi, \pi]$.
Angles closer to the axes imply weaker connections, while those deviating significantly from the axes lead to stronger interactions.
For more focused studies, these angles can be preset to specific values---such as $\frac{\pi}{4}$ and $\frac{5\pi}{180}$---to explore the impact of strong and weak connections within a fully connected structure, respectively.

\subsection{Exploring the Impact of Multimodality with various characteristics}
\label{sub:sub:multimodality}

Optimization landscapes often have numerous local optima, challenging optimization algorithms prone to premature convergence.
GNBG generates these complex landscapes using the nonlinear transformation $\mathbb{T}$.
The geometry---width, depth, and multiplicity---of the local optima within each component's basin can be controlled through the vectors $\bm\mu_k$ and $\bm\omega_k$.

To focus on how optimization algorithms handle landscapes with varying local optima, we create test instances without other influencing factors like ill-conditioning and rotation.
We achieve this by setting $\mathbf{H}=\mathbf{R}=\mathbf{I}_{d \times d}$.
Setting $\lambda$ to one avoids complexities from sub-linear basins.
Although setting $o$ greater than one usually creates multimodal instances, we set it to one to focus on a single component, avoiding challenges from multiple promising regions, including deceptiveness.
Finally, to achieve a symmetric distribution of local optima within the basin, we set $\mu_1=\mu_2$ and $\omega_1=\omega_2=\omega_3=\omega_4$.

For a thorough evaluation, users should vary $\bm\mu$ and $\bm\omega$ across a wide range, such as $\{0.1, 0.25, 0.5, 0.75, 1\}$ and $\{5, 25, 50, 100\}$, respectively. 
Lower values of $\bm\mu$ yield shallower local optima, while higher values deepen them. The $\bm\omega$ values modulate both the width and density of the local optima---higher values narrow the optima while increasing their multiplicity.

\subsection{Exploring the Impact of Multiple Competitive Components}
\label{sub:sub:MultipleComponents}

Many real-world optimization landscapes contain multiple promising regions, each with vast basins of attraction surrounding high-quality solutions~\cite{yazdani2023robust,de1999evolving}.
These regions can mislead optimization algorithms into suboptimal solutions.
GNBG allows the creation of such landscapes by specifying multiple components, each with distinct attributes like position and size.
Note that incorporating multiple components can introduce significant asymmetry.
Furthermore, even with $\mathbf{R}_k$ set as the identity matrix, multiple components make the problem instance fully non-separable~\cite{yazdani2018thesis}.

To isolate the impact of multiple promising regions, we standardize certain component attributes. 
Specifically, we set $\lambda_k = 1$ for all components to maintain a uniformly scaled landscape. 
Additionally, we neutralize other influencing factors by setting all elements of $\bm{\mu}_k$ and $\bm{\omega}_k$ to zero and configuring $\mathbf{R}_k$ as the identity matrix. 
All elements of the principal diagonal of $\mathbf{H}_k$ are set to a uniform value to ensure well-conditioned components.

The users can manually configure other component parameters, such as location $\vec{m}_k$, minimum function value $\sigma_k$, and the matrix $\mathbf{H}_k$. 
This flexibility allows for the design of problem instances with specific deceptiveness characteristics. 
However, manual configuration can be challenging with a large number of components.
To simplify this process, these parameters can be randomized within user-defined ranges, allowing users to focus primarily on varying the value of $o$. 
For a comprehensive analysis of the impact of multiple promising regions, we recommend varying the number of components $o$ across a broad spectrum, for example, $o \in \{1, 2, 5, 10, 25, 50\}$.

\subsection{Exploring Complex Combinations of Challenges and Features}
\label{sub:sub:combination}

Sections~\ref{sub:sub:linearity} through \ref{sub:sub:MultipleComponents} discussed generating problem instances to examine specific characteristics in isolation, ensuring external factors do not affect algorithm performance.
These specialized tests are a key feature of GNBG.
However, real-world scenarios often combine multiple challenges, which increase complexity.
Using GNBG's adaptability, researchers can create a wide range of problem instances that reflect real-world optimization problems.
The following outlines methods to create problem instances that combine various characteristics and challenges, tailored to specific research objectives.

\begin{figure*}[!t]
\centering
\begin{tabular}{ccc}
     \subfigure[{\scriptsize A landscape containing an ill-conditioned, sub-linear, rotated, multimodal, and asymmetric component.}]{\includegraphics[width=0.30\linewidth]{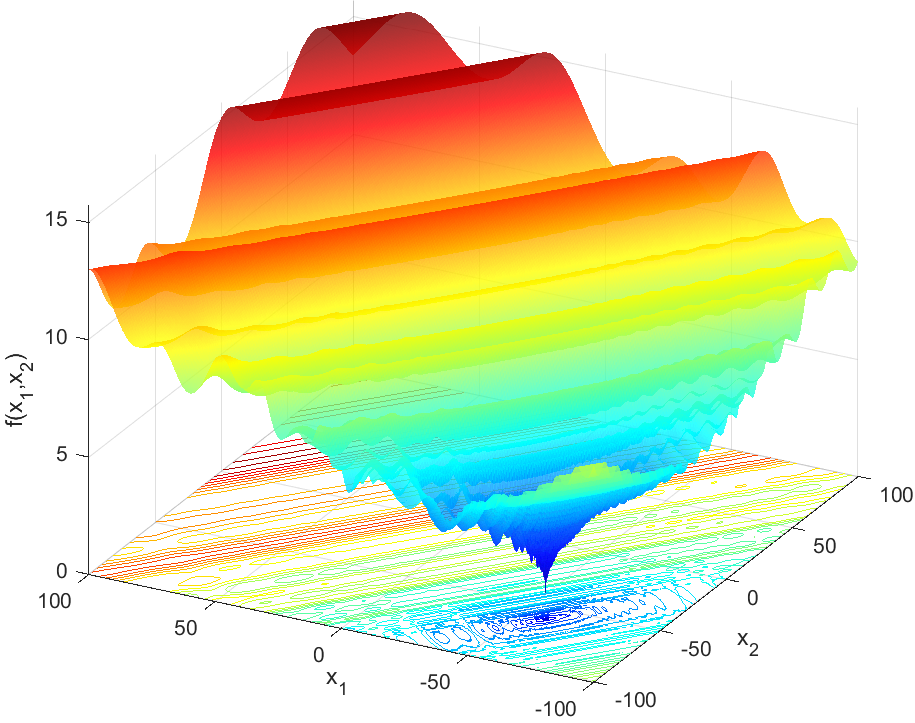}\label{fig:Mix0}}
     &
      \subfigure[{\scriptsize A landscape containing two components with different configurations.}]{\includegraphics[width=0.30\linewidth]{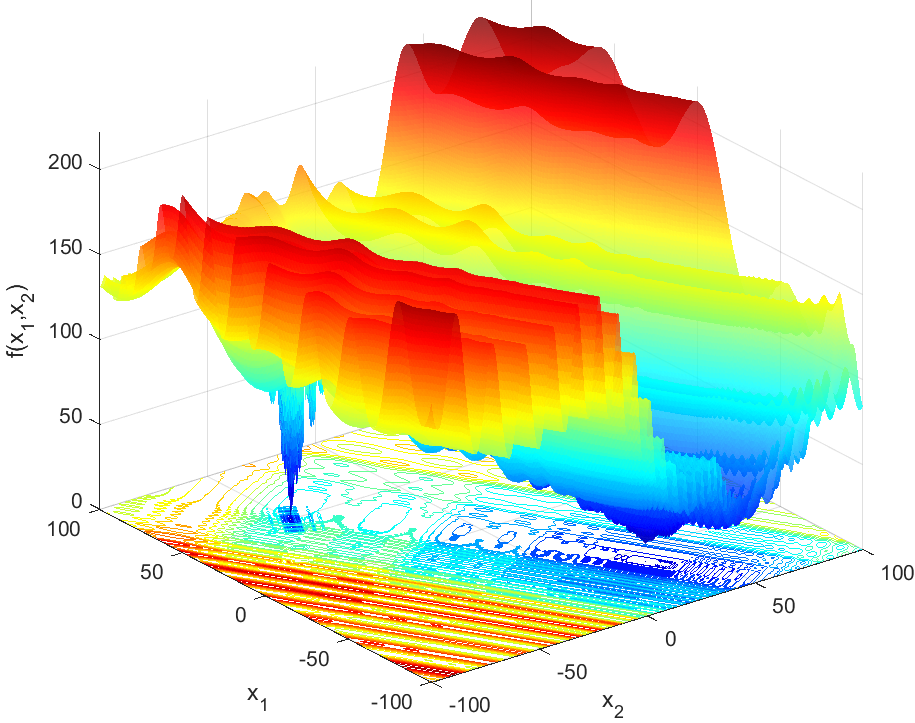}\label{fig:2Components}}
     &    \subfigure[{\scriptsize An unimodal valley constructed by an ill-conditioned and sub-linear component.}]{\includegraphics[width=0.30\linewidth]{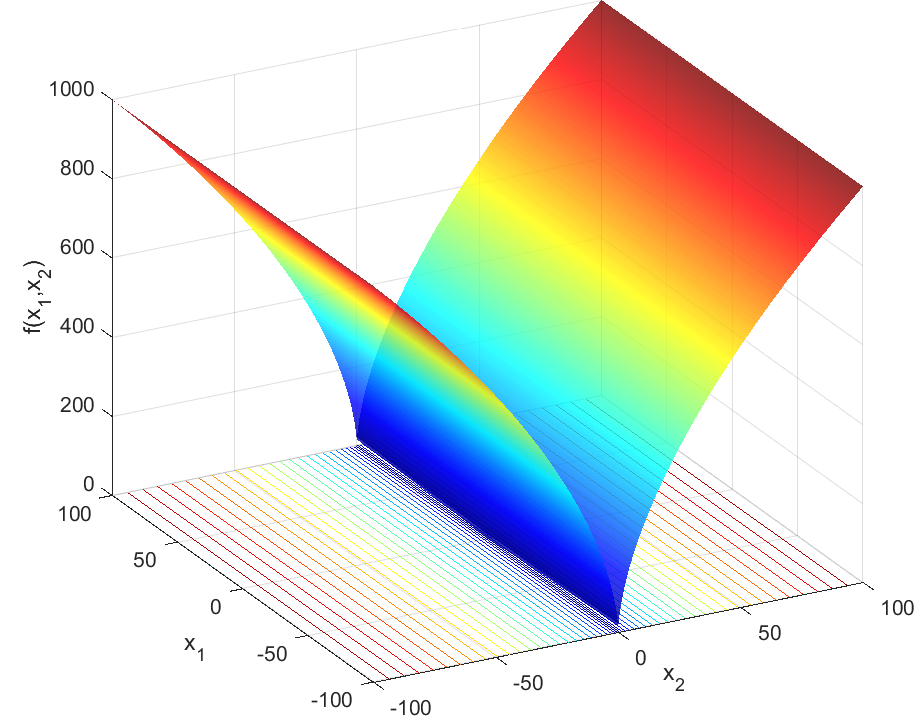}\label{fig:Mix1}}
     \\
     \subfigure[{\scriptsize A landscape containing multiple valleys constructed by an ill-conditioned, sub-linear, and multimodal component.}]{\includegraphics[width=0.30\linewidth]{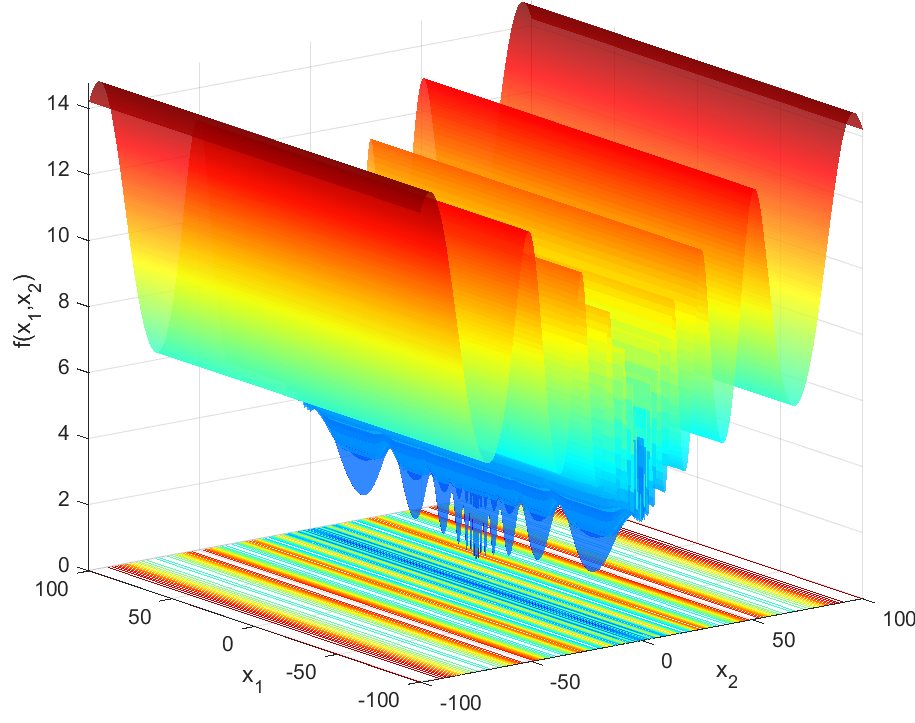}\label{fig:Mix2}}
    &
     \subfigure[{\scriptsize A landscape generated by two overlapped homogeneous components with different rotation angles.}]{\includegraphics[width=0.30\linewidth]{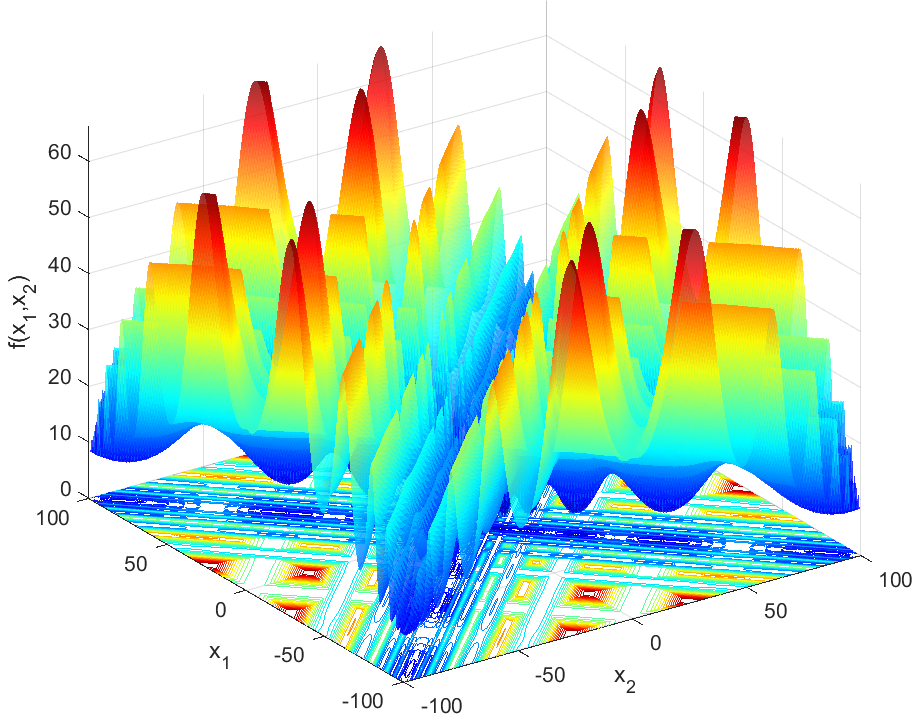}\label{fig:2OverlappedPeaks}}
    &
     \subfigure[{\scriptsize A landscape generated by three overlapped homogeneous components with different rotation angles.}]{\includegraphics[width=0.30\linewidth]{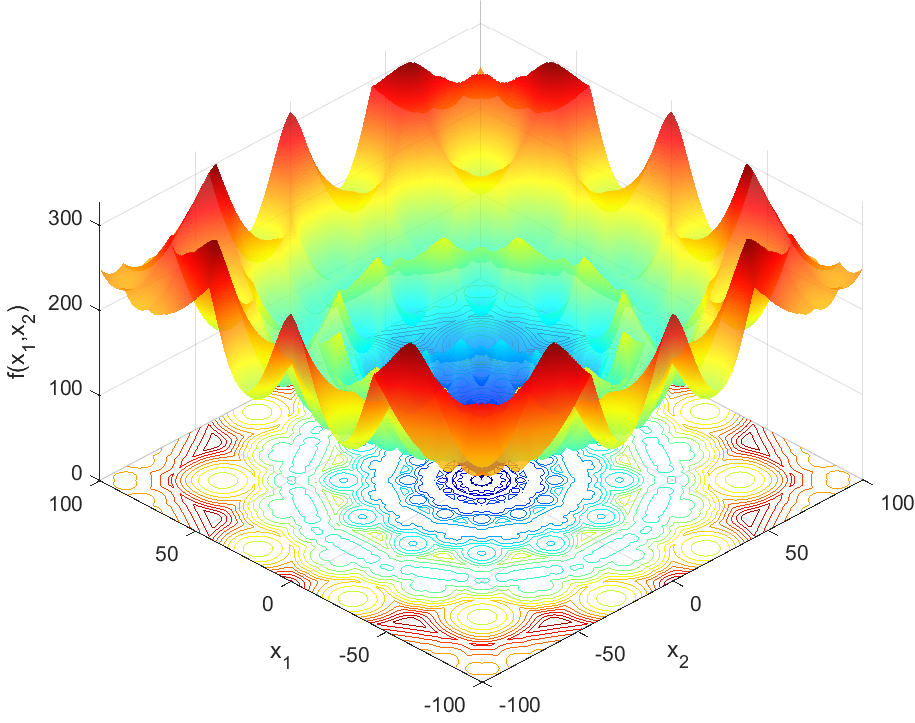}\label{fig:3OverlappedPeaks}}
\end{tabular}
\caption{Problem instances generated by GNBG with various combinations of characteristics and challenges.}
\label{fig:Combination}
\end{figure*}

Single-component problem instances with multiple challenges can be created by combining configurations from Sections~\ref{sub:sub:linearity} to \ref{sub:sub:multimodality}
For example, configuring $\mathbf{H}$ as in Section~\ref{sub:sub:conditioning} and adjusting $\bm\mu$ and $\bm\omega$ as in Section~\ref{sub:sub:multimodality} allows for the creation of ill-conditioned and multimodal problem landscapes.
This enables targeted studies of algorithm performance under specific complexities.
GNBG allows for the generation of various problem instances with different combinations of characteristics in a controlled manner.
Figure~~\ref{fig:Mix0} shows a landscape with an ill-conditioned, sub-linear, rotated, multimodal, and asymmetric component.
For more complex problem instances, GNBG can construct landscapes with multiple components, each with distinct characteristics.
Figure~\ref{fig:2Components} illustrates a problem instance with two components, each with distinct settings like size, shape, and rotation angle, where one component contains the global optimum, while the other spans a larger area, making the landscape inherently deceptive.

A specific class of challenging optimization problems features landscapes where the optimal solution lies within a valley, such as the Rosenbrock function\footnote{Also known as the banana valley.}. 
In these problems, after an algorithm converges to the valley's base, it must navigate a specific path to find the optimal solution---a task that proves difficult for many algorithms. 
By incorporating sub-linear and highly ill-conditioned basins, we can create such valleys using GNBG. 
A higher degree of ill-conditioning in the matrix $\mathbf{H}$ combined with lower $\lambda$ values results in narrower valleys, typically making the problem more challenging. 
Figure~\ref{fig:Mix1} shows an example of a unimodal problem instance with a distinct narrow valley.
Employing the transformation $\mathbb{T}$ can increase the problem's complexity significantly, leading to landscapes with multiple valleys, only one of which houses the global optimum. 
This dominant valley may contain local minima, further complicating the task of navigating its irregular and rugged terrain to find the global optimum solution. 
Figure~\ref{fig:Mix2} illustrates such a complex problem landscape.
The challenge intensifies when these problems have complex variable interaction structures, which further complicate the path-following process needed to locate the global optimum.

Figures~\ref{fig:3OverlappedPeaks} and ~\ref{fig:2OverlappedPeaks} highlight another capability of GNBG when employing multiple components: the generation of a ``hybrid component'' by overlapping several components with identical minimum positions and values. 
In these illustrated landscapes, two and three components are utilized, respectively. 
Though each component shares the same parameter settings in a given plot, they exhibit different rotation angles. 
This technique allows for the fusion of various components, resulting in a hybrid component with unique morphological characteristics not achievable with a single component.

\subsection{Exploring Scalability: Adjusting the Dimensionality of Problem Instances}
\label{sub:sub:scalability}

One of the key attributes of GNBG is its scalability.
The problem instances it generates can be easily adjusted with respect to dimension $d$.
Researchers have the flexibility to specify $d$ based on their investigative goals.
This flexibility is crucial when testing the robustness of optimization algorithms, especially as the dimensionality increases.
Such scalability becomes particularly important when examining algorithmic behaviors in the presence of challenges like ill-conditioning, multimodality, and complex variable interaction structures.
GNBG offers the ability to produce problem instances across a wide range of dimensions, facilitating experiments with, for instance, dimensions set as $ d \in \{10, 20, 30\} $ or even wider ranges.

\subsection{Preliminary Experimental Investigations}
\label{sub:sub:supplement:Experiment}

One unique ability of GNBG is to facilitate an in-depth examination of the capability of optimization algorithms in facing isolated challenges with controllable degrees, as outlined in Sections~\ref{sub:sub:linearity} to \ref{sub:sub:MultipleComponents}.
In Section~\ref{sec:sup:ImpactONalgorithms} of the supplementary document, we provide an initial exploration into how varying problem characteristics, generated by the GNBG, impact the performance of three well-known optimization algorithms: Pattern Search (PS)~\cite{torczon1997convergence}, Particle Swarm Optimization (PSO)~\cite{kennedy1995particle}, and Differential Evolution (DE)~\cite{storn1997differential}. 
The experiments focus on different attributes such as basin linearity, conditioning, variable interaction structures, multimodality, and the existence of multiple competitive components.
For each characteristic, specific parameter settings are altered to observe how these changes influence algorithm performance. 
The study provides insights into the strengths and weaknesses of each algorithm when tackling problems with varying degrees of complexity, which highlight areas for further research and improvement.

\subsection{GNBG-Generated Test Suite}
\label{sub:sub:supplement:Suite}

In Section~\ref{sec:sup:Suite} of the supplementary document, we introduce a comprehensive test suite composed of 24 problem instances generated by GNBG. These instances cover a wide range of characteristics, including modality, ruggedness, symmetry, conditioning, variable interactions, basin linearity, and deceptiveness. 
The test suite is categorized into three types: unimodal instances, multimodal instances with a single component, and multimodal instances with multiple components. 
This suite provides researchers with a platform to benchmark the effectiveness of their optimization algorithms against diverse and challenging scenarios, facilitating a deeper understanding of algorithmic performance under controlled and varied conditions. 
The MATLAB and Python source codes~\cite{yazdani2024GNBsuiteMatlab,yazdani2024GNBsuitePython} for these problem instances are made available for further experimentation and validation.

\section{Conclusion}
\label{sec:Conclusion}

This paper introduced the Generalized Numerical Benchmark Generator (GNBG), an innovative tool for generating problem instances with a diverse range of controllable characteristics.
Using a unique parametric baseline function, GNBG offers researchers control over attributes like dimensionality, variable interaction structures, conditioning, basin morphology, multimodality, ruggedness, symmetry, and deceptiveness.

Through an initial investigation reported in Section~\ref{sec:sup:ImpactONalgorithms} of the supplementary document, we examined the influence of these attributes on the performance of multiple optimization algorithms. 
This exploration revealed the strengths and weaknesses of these algorithms when dealing with complexities caused by different degrees of problem characteristics. 
Our results highlight the significant impact of GNBG’s adjustable characteristics on algorithmic performance. 
Elements such as complex variable interaction structures and specific attributes of local optima, including their depth and width, posed varying challenges to the algorithms under examination. 
While our insights are informative, they are, by nature, preliminary, indicating a clear need for broader and more in-depth studies. 

Future work should include a comprehensive empirical study focusing on a wide range of optimization algorithms. 
This study should examine problem instances generated by various parameter settings of GNBG to produce different degrees of isolated challenges described in Sections~\ref{sub:sub:linearity} through \ref{sub:sub:MultipleComponents} and the 24 predefined problem instances detailed in Section~\ref{sec:sup:Suite} of the supplementary document.


\setstretch{0.96}
\small

\bibliography{bib}

@article{mei2016competitive,
    title={A competitive divide-and-conquer algorithm for unconstrained large-scale black-box optimization},
    author={Mei, Yi and Omidvar, Mohammad Nabi and Li, Xiaodong and Yao, Xin},
    journal={ACM Transactions on Mathematical Software},
    volume={42},
    number={2},
    pages={13},
    year={2016},
    publisher={ACM},
}

@article{michalewicz2000test,
  title={Test-case generator for nonlinear continuous parameter optimization techniques},
  author={Michalewicz, Zbigniew and Deb, Kalyanmoy and Schmidt, Martin and Stidsen, Thomas},
  journal={IEEE Transactions on Evolutionary Computation},
  volume={4},
  number={3},
  pages={197--215},
  year={2000},
  publisher={IEEE}
}

@article{munoz2020generating,
  title={Generating new space-filling test instances for continuous black-box optimization},
  author={Mu{\~n}oz, Mario A and Smith-Miles, Kate},
  journal={Evolutionary computation},
  volume={28},
  number={3},
  pages={379--404},
  year={2020},
  publisher={MIT Press One Rogers Street, Cambridge, MA 02142-1209, USA journals-info~…}
}

@manual{yazdani2024GNBGgeneratorMatlab,
      title  = "Generalized Numerical Benchmark Generator (GNBG)-Instance Generator (MATLAB Source Code)",
      author = "Danial Yazdani",
      url    = "https://github.com/Danial-Yazdani/GNBG_Generator.MATLAB",
      year   = "2024"
    }

@manual{yazdani2024GNBGgeneratorPython,
      title  = "Generalized Numerical Benchmark Generator (GNBG)-Instance Generator (Python Source Code)",
      author = "Danial Yazdani",
      url    = "https://github.com/Danial-Yazdani/GNBG_Generator.Python",
      year   = "2024"
    }

@manual{yazdani2024GNBsuiteMatlab,
      title  = "24 Problem Instances Generated by Generalized Numerical Benchmark Generator (GNBG) (MATLAB Source Code)",
      author = "Danial Yazdani",
      url    = "https://github.com/Danial-Yazdani/GNBG_Instances.MATLAB",
      year   = "2024"
    }

@manual{yazdani2024GNBsuitePython,
      title  = "24 Problem Instances Generated by Generalized Numerical Benchmark Generator (GNBG) (Python Source Code)",
      author = "Danial Yazdani",
      url    = "https://github.com/Danial-Yazdani/GNBG_Instances.Python",
      year   = "2024"
    }

@article{audet2002analysis,
  title={Analysis of generalized pattern searches},
  author={Audet, Charles and Dennis Jr, John E},
  journal={SIAM Journal on optimization},
  volume={13},
  number={3},
  pages={889--903},
  year={2002},
  publisher={SIAM}
}

@article{abramson2009orthomads,
  title={OrthoMADS: A deterministic MADS instance with orthogonal directions},
  author={Abramson, Mark A and Audet, Charles and Dennis Jr, John E and Digabel, S{\'e}bastien Le},
  journal={SIAM Journal on Optimization},
  volume={20},
  number={2},
  pages={948--966},
  year={2009},
  publisher={SIAM}
}

@article{yazdani2023robust,
  title={Robust Optimization Over Time: A Critical Review},
  author={Yazdani, Danial and Omidvar, Mohammad Nabi and Yazdani, Donya and Branke, J{\"u}rgen and Nguyen, Trung Thanh and Gandomi, Amir H and Jin, Yaochu and Yao, Xin},
  journal={IEEE Transactions on Evolutionary Computation},
  year={2023},
  publisher={IEEE}
}

@article{bartz2020benchmarking,
  title={Benchmarking in optimization: Best practice and open issues},
  author={Bartz-Beielstein, Thomas and Doerr, Carola and Berg, Daan van den and Bossek, Jakob and Chandrasekaran, Sowmya and Eftimov, Tome and Fischbach, Andreas and Kerschke, Pascal and La Cava, William and Lopez-Ibanez, Manuel and others},
  journal={arXiv preprint arXiv:2007.03488},
  year={2020}
}

@article{beiranvand2017best,
  title={Best practices for comparing optimization algorithms},
  author={Beiranvand, Vahid and Hare, Warren and Lucet, Yves},
  journal={Optimization and Engineering},
  volume={18},
  pages={815--848},
  year={2017},
  publisher={Springer}
}

@article{olson2017pmlb,
  title={PMLB: a large benchmark suite for machine learning evaluation and comparison},
  author={Olson, Randal S and La Cava, William and Orzechowski, Patryk and Urbanowicz, Ryan J and Moore, Jason H},
  journal={BioData mining},
  volume={10},
  pages={1--13},
  year={2017},
  publisher={Springer}
}

@article{kerschke2019comprehensive,
  title={Comprehensive feature-based landscape analysis of continuous and constrained optimization problems using the R-package flacco},
  author={Kerschke, Pascal and Trautmann, Heike},
  journal={Applications in Statistical Computing: From Music Data Analysis to Industrial Quality Improvement},
  pages={93--123},
  year={2019},
  publisher={Springer}
}

@book{rao2019engineering,
  title={Engineering optimization: theory and practice},
  author={Rao, Singiresu S},
  year={2019},
  publisher={John Wiley \& Sons}
}

@article{archetti2022optimization,
  title={Optimization in multimodal freight transportation problems: A Survey},
  author={Archetti, Claudia and Peirano, Lorenzo and Speranza, M Grazia},
  journal={European Journal of Operational Research},
  volume={299},
  number={1},
  pages={1--20},
  year={2022},
  publisher={Elsevier}
}

@article{cousineau2023estimating,
  title={Estimating causal effects with optimization-based methods: A review and empirical comparison},
  author={Cousineau, Martin and Verter, Vedat and Murphy, Susan A and Pineau, Joelle},
  journal={European Journal of Operational Research},
  volume={304},
  number={2},
  pages={367--380},
  year={2023},
  publisher={Elsevier}
}

@article{malan2013survey,
  title={A survey of techniques for characterising fitness landscapes and some possible ways forward},
  author={Malan, Katherine M and Engelbrecht, Andries P},
  journal={Information Sciences},
  volume={241},
  pages={148--163},
  year={2013},
  publisher={Elsevier}
}

@article{munoz2015algorithm,
  title={Algorithm selection for black-box continuous optimization problems: A survey on methods and challenges},
  author={Mu{\~n}oz, Mario A and Sun, Yuan and Kirley, Michael and Halgamuge, Saman K},
  journal={Information Sciences},
  volume={317},
  pages={224--245},
  year={2015},
  publisher={Elsevier}
}

@article{jamil2013literature,
  title={A literature survey of benchmark functions for global optimisation problems},
  author={Jamil, Momin and Yang, Xin-She},
  journal={International Journal of Mathematical Modelling and Numerical Optimisation},
  volume={4},
  number={2},
  pages={150--194},
  year={2013},
  publisher={Inderscience Publishers Ltd}
}

@article{gouvea2016global,
  title={Global optimization using q-gradients},
  author={Gouv{\^e}a, {\'E}rica JC and Regis, Rommel G and Soterroni, Aline C and Scarabello, Marluce C and Ramos, Fernando M},
  journal={European Journal of Operational Research},
  volume={251},
  number={3},
  pages={727--738},
  year={2016},
  publisher={Elsevier}
}

@article{chelouah2003genetic,
  title={Genetic and Nelder--Mead algorithms hybridized for a more accurate global optimization of continuous multiminima functions},
  author={Chelouah, Rachid and Siarry, Patrick},
  journal={European Journal of operational research},
  volume={148},
  number={2},
  pages={335--348},
  year={2003},
  publisher={Elsevier}
}

@article{uchoa2017new,
  title={New benchmark instances for the capacitated vehicle routing problem},
  author={Uchoa, Eduardo and Pecin, Diego and Pessoa, Artur and Poggi, Marcus and Vidal, Thibaut and Subramanian, Anand},
  journal={European Journal of Operational Research},
  volume={257},
  number={3},
  pages={845--858},
  year={2017},
  publisher={Elsevier}
}

@article{skaalnes2023new,
  title={New Benchmark Instances for the Inventory Routing Problem},
  author={Sk{\aa}lnes, J{\o}rgen and Ahmed, Mohamed Ben and Hvattum, Lars Magnus and St{\aa}lhane, Magnus},
  journal={European Journal of Operational Research},
  year={in press, 2023},
  publisher={Elsevier}
}

@inproceedings{whitley2002testing,
  title={Testing, evaluation and performance of optimization and learning systems},
  author={Whitley, D and Watson, Jean-Paul and Howe, Adele and Barbulescu, Laura},
  booktitle={Adaptive Computing in Design and Manufacture V},
  pages={27--39},
  year={2002},
  organization={Springer}
}

@inproceedings{shir2018compiling,
  title={Compiling a benchmarking test-suite for combinatorial black-box optimization: a position paper},
  author={Shir, Ofer M and Doerr, Carola and B{\"a}ck, Thomas},
  booktitle={Proceedings of the Genetic and Evolutionary Computation Conference Companion},
  pages={1753--1760},
  year={2018}
}

@article{torczon1997convergence,
  title={On the convergence of pattern search algorithms},
  author={Torczon, Virginia},
  journal={SIAM Journal on optimization},
  volume={7},
  number={1},
  pages={1--25},
  year={1997},
  publisher={SIAM}
}

@article{saini2021multi,
  title={Multi-objective optimization techniques: a survey of the state-of-the-art and applications: Multi-objective optimization techniques},
  author={Saini, Naveen and Saha, Sriparna},
  journal={The European Physical Journal Special Topics},
  volume={230},
  number={10},
  pages={2319--2335},
  year={2021},
  publisher={Springer}
}

@article{hellwig2019benchmarking,
  title={Benchmarking evolutionary algorithms for single objective real-valued constrained optimization--a critical review},
  author={Hellwig, Michael and Beyer, Hans-Georg},
  journal={Swarm and evolutionary computation},
  volume={44},
  pages={927--944},
  year={2019},
  publisher={Elsevier}
}

@Techreport{hansen2009real,
  title={Real-parameter black-box optimization benchmarking 2009: Noiseless functions definitions},
  author={Hansen, Nikolaus and Finck, Steffen and Ros, Raymond and Auger, Anne},
  year={2009},
  school={INRIA}
}

@article{mei2021structural,
  title={Structural optimization in civil engineering: a literature review},
  author={Mei, Linfeng and Wang, Qian},
  journal={Buildings},
  volume={11},
  number={2},
  pages={66},
  year={2021},
  publisher={MDPI}
}

@article{weise2009global,
  title={Global optimization algorithms-theory and application},
  author={Weise, Thomas},
  journal={Self-Published Thomas Weise},
  volume={361},
  year={2009}
}

@Techreport{awad2016problem,
  title={Problem definitions and evaluation criteria for the {CEC} 2017 special session and competition on single objective bound constrained real-parameter numerical optimization},
  author={Awad, NH and Ali, MZ and Liang, JJ and Qu, BY and Suganthan, PN},
  booktitle={Technical report},
  year={2016},
  publisher={Nanyang Technological University Singapore}
}

@article{hansen2021coco,
  title={{COCO}: A platform for comparing continuous optimizers in a black-box setting},
  author={Hansen, Nikolaus and Auger, Anne and Ros, Raymond and Mersmann, Olaf and Tu{\v{s}}ar, Tea and Brockhoff, Dimo},
  journal={Optimization Methods and Software},
  volume={36},
  number={1},
  pages={114--144},
  year={2021},
  publisher={Taylor \& Francis}
}

@article{gandomi2012evolutionary,
  title={Evolutionary boundary constraint handling scheme},
  author={Gandomi, Amir Hossein and Yang, Xin-She},
  journal={Neural Computing and Applications},
  volume={21},
  pages={1449--1462},
  year={2012},
  publisher={Springer}
}

@Techreport{yue2019problem,
  title={Problem definitions and evaluation criteria for the {CEC} 2020 special session and competition on single objective bound constrained numerical optimization},
  author={Yue, CT and Price, Kenneth V and Suganthan, Ponnuthurai N and Liang, JJ and Ali, Mostafa Z and Qu, BY and Awad, Noor H and Biswas, Partha P},
  journal={Comput. Intell. Lab., Zhengzhou Univ., Zhengzhou, China, Tech. Rep},
  year={2019}
}

@article{molga2005test,
  title={Test functions for optimization needs},
  author={Molga, Marcin and Smutnicki, Czes{\l}aw},
  journal={Test functions for optimization needs},
  volume={101},
  pages={48},
  year={2005}
}

@article{more1981testing,
  title={Testing unconstrained optimization software},
  author={Mor{\'e}, Jorge J and Garbow, Burton S and Hillstrom, Kenneth E},
  journal={ACM Transactions on Mathematical Software (TOMS)},
  volume={7},
  number={1},
  pages={17--41},
  year={1981},
  publisher={ACM New York, NY, USA}
}

@article{helwig2012experimental,
  title={Experimental analysis of bound handling techniques in particle swarm optimization},
  author={Helwig, Sabine and Branke, Juergen and Mostaghim, Sanaz},
  journal={IEEE Transactions on Evolutionary Computation},
  volume={17},
  number={2},
  pages={259--271},
  year={2012},
  publisher={IEEE}
}

@Inproceedings{eberhart2001comparing,
    author        = "Russell Eberhart and Y. Shi",
    title         = "Comparing inertia weights and constriction factors in particle swarm optimization",
    booktitle       = "IEEE Congress on Evolutionary Computation",
    volume        = "1",
    year          = "2001",
    pages         = "84--88",
publisher = "IEEE",
}

@Techreport{suganthan2005problem,
    author =       "P. N. Suganthan and N. Hansen and J. J. Liang and K. Deb and Y.-P. Chen and A. Auger and S. Tiwari",
    year =         "2005",
    title =        "Problem definitions and evaluation criteria for the CEC 2005 special session on real-parameter
optimization",
    institution =  "Nanyang Technological University",
    type =         "",
    number =       "",
    address =      "",
    month =        "",
    note =         "",
}

@inproceedings{de1999evolving,
  title={Evolving in a changing world},
  author={De Jong, Kenneth},
  booktitle={International Symposium on Methodologies for Intelligent Systems},
  pages={512--519},
  year={1999},
  organization={Springer}
}

@article{storn1997differential,
  title={Differential evolution--a simple and efficient heuristic for global optimization over continuous spaces},
  author={Storn, Rainer and Price, Kenneth},
  journal={Journal of global optimization},
  volume={11},
  pages={341--359},
  year={1997},
  publisher={Springer}
}

@article{omidvar2017dg2,
    title={{DG2}: A faster and more accurate differential grouping for large-scale black-box optimization},
    author={Omidvar, Mohammad Nabi and Yang, Ming and Mei, Yi and Li, Xiaodong and Yao, Xin},
    journal={IEEE Transactions on Evolutionary Computation},
    volume={21},
    number={6},
    pages={929--942},
    year={2017},
    publisher={IEEE},
}

@Phdthesis{yazdani2018thesis,
    author =       "Danial Yazdani",
    year =         "2018",
    title =        "Particle swarm optimization for dynamically changing environments with particular focus on scalability and switching cost",
    school =       "Liverpool John Moores University",
    address =      "Liverpool, UK",
}

@article{yazdani2019scaling,
  author={Yazdani, Danial and Omidvar, Mohammad Nabi and Branke, Juergen and Nguyen, Trung Thanh and Yao, Xin},
  journal={IEEE Transactions on Evolutionary Computation}, 
  title={Scaling Up Dynamic Optimization Problems: A Divide-and-Conquer Approach}, 
  year={2020},
  volume={24},
  number={1},
  pages={1-15},
publisher={IEEE},
}

@article{das2010differential,
  title={Differential evolution: A survey of the state-of-the-art},
  author={Das, Swagatam and Suganthan, Ponnuthurai Nagaratnam},
  journal={IEEE transactions on Evolutionary Computation},
  volume={15},
  number={1},
  pages={4--31},
  year={2010},
  publisher={IEEE}
}

@inproceedings{kennedy1995particle,
  title={Particle swarm optimization},
  author={Kennedy, James and Eberhart, Russell},
  booktitle={Proceedings of ICNN'95-International Conference on Neural Networks},
  volume={4},
  pages={1942--1948},
  year={1995},
  organization={IEEE}
}

@misc{bonyadi2017particle,
  title={Particle swarm optimization for single objective continuous space problems: a review},
  author={Bonyadi, Mohammad Reza and Michalewicz, Zbigniew},
  journal={Evolutionary Computation},
  volume={25},
  number={1},
  pages={1-54},
    year={2017},
  publisher={MIT Press}
}

@article{omidvar2015designing,
    title={Designing benchmark problems for large-scale continuous optimization},
    author={Mohammad Nabi Omidvar and Xiaodong Li and Ke Tang},
    journal={Information Sciences},
    volume={316},
    number={},
    pages={419--436},
    year={2015},
    doi = "10.1016/j.ins.2014.12.062",
}

@article{yazdani2020benchmarking,
  author={Yazdani, Danial and Omidvar, Mohammad Nabi and Cheng, Ran and Branke, Jürgen and Nguyen, Trung Thanh and Yao, Xin},
  journal={IEEE Transactions on Cybernetics}, 
  title={Benchmarking Continuous Dynamic Optimization: Survey and Generalized Test Suite}, 
  year={2022},
  volume={52},
  number={5},
  pages={3380-3393},
  publisher={IEEE},
}

@article{yazdani2021DOPsurveyPartA,
  title={A Survey of Evolutionary Continuous Dynamic Optimization Over Two Decades -- Part {A}},
  author={Yazdani, Danial and Cheng, Ran and Yazdani, Donya and Branke, J{\"u}rgen and Jin, Yaochu and Yao, Xin},
  journal={IEEE Transactions on Evolutionary Computation},
  volume={25},
  number={4},
  pages={609--629},
  year={2021},
  publisher={IEEE}
}

@article{yazdani2021DOPsurveyPartB,
  title={A Survey of Evolutionary Continuous Dynamic Optimization Over Two Decades -- Part {B}},
  author={Yazdani, Danial and Cheng, Ran and Yazdani, Donya and Branke, J{\"u}rgen and Jin, Yaochu and Yao, Xin},
  journal={IEEE Transactions on Evolutionary Computation},
  volume={25},
  number={4},
  pages={630--650},
  year={2021},
  publisher={IEEE}
}

@article{kusakci2012constrained,
  title={Constrained optimization with evolutionary algorithms: a comprehensive review},
  author={Kusakci, Ali Osman and Can, Mehmet},
  journal={Southeast Europe journal of soft computing},
  volume={1},
  number={2},
  year={2012}
}

@inproceedings{deb2002scalable,
  title={Scalable multi-objective optimization test problems},
  author={Deb, Kalyanmoy and Thiele, Lothar and Laumanns, Marco and Zitzler, Eckart},
  booktitle={Proceedings of the 2002 Congress on Evolutionary Computation. CEC'02 (Cat. No. 02TH8600)},
  volume={1},
  pages={825--830},
  year={2002},
  organization={IEEE}
}

@article{li2016seeking,
  title={Seeking multiple solutions: An updated survey on niching methods and their applications},
  author={Li, Xiaodong and Epitropakis, Michael G and Deb, Kalyanmoy and Engelbrecht, Andries},
  journal={IEEE Transactions on Evolutionary Computation},
  volume={21},
  number={4},
  pages={518--538},
  year={2016},
  publisher={IEEE}
}

@article{omidvar2021reviewA,
  title={A review of population-based metaheuristics for large-scale black-box global optimization—{Part I}},
  author={Omidvar, Mohammad Nabi and Li, Xiaodong and Yao, Xin},
  journal={IEEE Transactions on Evolutionary Computation},
  volume={26},
  number={5},
  pages={802--822},
  year={2021},
  publisher={IEEE}
}

@article{omidvar2021reviewB,
  title={A review of population-based metaheuristics for large-scale black-box global optimization—{Part II}},
  author={Omidvar, Mohammad Nabi and Li, Xiaodong and Yao, Xin},
  journal={IEEE Transactions on Evolutionary Computation},
  volume={26},
  number={5},
  pages={823--843},
  year={2021},
  publisher={IEEE}
}
\bibliographystyle{IEEEtran}

\end{document}


\maketitle

\begin{abstract}
This supplementary document provides additional context for `A Generalized and Configurable Benchmark Generator for Continuous Unconstrained Numerical Optimization.' 
It is organized as follows: Section~\ref{sec:sup:proof} offers a mathematical proof demonstrating that components generated by the Generalized Numerical Benchmark Generator (GNBG) can be optimized dimension-wise if the rotation matrix is set to the identity matrix.
Section~\ref{sec:sup:GNBGparameters} summarizes  the parameters of GNBG.
Section~\ref{sec:sup:ImpactONalgorithms} presents a preliminary empirical study examining the influence of various problem characteristics on the performance of selected optimization algorithms. 
Finally, Section~\ref{sec:sup:Suite} introduces a test suite that includes 24 different problem instances generated by GNBG.
\end{abstract}

\begin{IEEEkeywords}
Global optimization, Benchmark generator, Test suite, Performance evaluation, Optimization algorithms. 
 \end{IEEEkeywords}
\IEEEpeerreviewmaketitle



 \section{Mathematical Proof of Dimensional-Wise Optimizability in GNBG Components with Identity Rotation Matrix}
 \label{sec:sup:proof}

For a component generated by GNBG, setting $\lambda$ to one and the rotation matrix $\mathbf{R}$ to $\mathbf{I}$ makes the component fully separable. 
However, while keeping $\mathbf{R}=\mathbf{I}$, when $\lambda \neq 1$, the component becomes additively non-separable, yet it can still be optimized in a dimension-wise manner, allowing the optimum to be found in each dimension separately. 
In this section, we mathematically prove that if the rotation matrix is set to the identity matrix, a component generated by GNBG can be optimized in a dimension-wise manner, regardless of the values of its other parameters.
Here, we assume that the shift parameter $\vec{m}$ is set to the zero vector, as this translation does not alter the function's interaction structure.

For a continuous function, variable interaction is defined as follows.
\begin{definition}[Mei et al.~\cite{mei2016competitive}]\label{def:interaction}
Let $f: \mathbb{R}^n \rightarrow \mathbb{R}$ be a twice differentiable function. Decision variables $x_i$ and $x_j$ interact if  a candidate solution $\mathbf{x}^\star$ exists, such that
\begin{equation}\label{Equation: direct interaction}
\frac{\partial^2 f(\mathbf{x}^\star)}{\partial x_i \partial x_j} \neq 0. \nonumber
\end{equation}
\end{definition}

In what follows, we derive the first order partial derivative of a single component generated by Equation \eqref{eq:irGMPB} with respect to the decision variables $\vec{z}$ and show that the second-order partial derivatives are non-zero suggesting that the function is (additively) non-separable. We then show that the extremum of the functions can be found in a dimension-wise manner by setting the first-order derivatives to zero.

\begin{align}
    \frac{\partial f(\vec{z})}{\partial\vec{z}} 
    &=\frac{\partial}{\partial	 \vec{z}} \big(( \mathbb{T}( \mathbf{R}\vec{z})^\top\mathbf{H}\mathbb{T}(\mathbf{R}\vec{z}) \big) ^\lambda \\
    &= \underbrace{
        \lambda\left(\mathbb{T}(\mathbf{R}\vec{z})^\top\mathbf{H}\mathbb{T}(\mathbf{R}\vec{z})\right)^{(\lambda-1)}}_{\xi
    }
    \frac{\partial}{\partial \vec{z}} 
    \mathbb{T}(\mathbf{R}\vec{z})^\top\mathbf{H}\mathbb{T}(\mathbf{R}\vec{z})
\end{align}
Letting $\xi = \lambda\left(\mathbb{T}(\mathbf{R}\vec{z})^\top\mathbf{H}\mathbb{T}(\mathbf{R}\vec{z})\right)^{(\lambda-1)}$, assuming $\mathbf{R} = \mathbf{I}$, and $\mathbf{H}$ is symmetric, we have:
\begin{align}
    \frac{\partial f(\vec{z})}{\partial\vec{z}} 
    &= \xi \frac{\partial}{\partial \vec{z}} 
    \left(\mathbb{T}(\mathbf{R}\vec{z})^\top\mathbf{H}\mathbb{T}(\mathbf{R}\vec{z}) \right)\\
    &= 2\xi\mathbb{T}(\mathbf{R}\vec{z})^\top\mathbf{H} \frac{\partial}{\partial\vec{z}}\mathbb{T}(\mathbf{R}\vec{z}) \\
    &= 2\xi\mathbb{T}(\vec{z})^\top\mathbf{H} \frac{\partial}{\partial\vec{z}}\mathbb{T}(\vec{z})
\end{align}
Now, to calculate $\frac{\partial\mathbb{T}(\vec{z})}{\partial\vec{z}}$, we rewrite $\mathbb{T}(\vec{z})$ as follows:
\begin{align}
    \mathbb{T}\left(\vec{z}\right) &= \exp\bigg( \ln(\vec{z})+\mu_1 \Big(\sin\left(\omega_1\ln(\vec{z})\right)+\sin\left(\omega_2\ln\left(\vec{z}\right)\right)  \Big)\bigg) \\
    & = \vec{z} \exp{\bigg(\mu \underbrace{\sin\left(\omega_1\ln(\vec{z})\right)+\sin\left(\omega_2\ln\left(\vec{z}\right)\right)}_{\eta(\vec{z})} \bigg)}
\end{align}
Letting $\eta =\sin\left(\omega_1\ln(\vec{z})\right)+\sin\left(\omega_2\ln\left(\vec{z}\right)\right)$, we can rewrite $\mathbb{T} = \vec{z} e^{\mu\eta}$. Thus,
\begin{align}
\frac{\partial\mathbb{T}\left(\vec{z}\right)}{\partial\vec{z}} 
    = e^{\mu\eta(\vec{z})}+ \vec{z}\frac{\partial e^{\mu\eta(\vec{z})}}{\partial\vec{z}} 
    = e^{\mu\eta(\vec{z})}+ \vec{z}e^{\mu\eta(\vec{z})} \frac{\partial {\mu\eta(\vec{z})}}{\partial\vec{z}}
\end{align}
Now, to calculate $\frac{\partial {\mu\eta(\vec{z})}}{\partial\vec{z}}$ we have:
\begin{align}
    \frac{\partial {\mu\eta(\vec{z})}}{\partial\vec{z}}
    &= \mu\left( \frac{\partial}{\partial\vec{z}}\sin\left(\omega_1 \ln(\vec{z})\right)+\frac{\partial}{\partial\vec{z}}\sin\left(\omega_2 \ln(\vec{z})\right)  \right) \\
    &= \mu\left( \frac{\omega_1}{\vec{z}}\cos\left(\omega_1 \ln(\vec{z})\right)+\frac{\omega_2}{\vec{z}}\cos\left(\omega_2 \ln(\vec{z})\right)  \right) := \gamma
\end{align}
Therefore, 
\begin{align}
\frac{\partial f(\vec{z})}{\partial \vec{z}} 
    &= 2\xi \mathbb{T}(\vec{z})^\top \mathbf{H}\left(e^{\mu\eta(\vec{z})} + \vec{z}e^{\mu\eta(\vec{z})}  \cdot \gamma\right), \text{ where } \label{eq:fod} \\
    \gamma &=\mu\left( \frac{\omega_1}{\vec{z}}\cos\left(\omega_1 \ln(\vec{z})\right)+\frac{\omega_2}{\vec{z}}\cos\left(\omega_2 \ln(\vec{z})\right)  \right), \label{eq:gamma} \\
    \xi &= \lambda\left(\mathbb{T}\left( \mathbf{R}\vec{z}\right)^\top\mathbf{H}\mathbb{T}\left(\mathbf{R}\vec{z}\right)\right)^{(\lambda-1)} \label{eq:xi}, \\
    & \:\:\:\:\;\mathbf{R} = \mathbf{I}, \text{ and } \mathbf{H} \text{ is diagonal.} \nonumber
\end{align}

Equations~\eqref{eq:fod}-\eqref{eq:xi} clearly show that the second-order partial derivative is non-zero when $\vec{z} \ne \vec{0}$. Now by setting the first-order partial derivatives to zero, we show that the optimum for each dimension can be found independently of the values taken by other dimensions. Note that $\mathbf{H}$ is symmetric and positive-definite, therefore $\xi > 0$ when $\vec{z} \ne \vec{0}$. Thus, the extremums can be found as follows:

\begin{align}
\resizebox{\columnwidth}{!}{$
\frac{\partial f(\vec{z})}{\partial \vec{z}}= 0 \implies
    \begin{cases}
        \mathbb{T}(\vec{z}) = 0, \text{ or} \\
        \\
        e^{\mu\eta(\vec{z})} +\vec{z}e^{\mu\eta(\vec{z})}\cdot\gamma = 0 \\
    \end{cases}
    \implies
    \begin{cases}
        \vec{z}=0, \text{ or} \\
        \\
        -\vec{z}\gamma=1\\
    \end{cases} $}
\end{align}

\begin{align}
    -\vec{z}\gamma &= 1 \notag \\
    &\implies \vec{z}\mu \bigg( 
        \frac{\omega_1}{\vec{z}} \cos\big(\omega_1 \ln(\vec{z})\big) 
        + \frac{\omega_2}{\vec{z}} \cos\big(\omega_2 \ln(\vec{z})\big)
    \bigg) = 1 \notag \\
    &\implies -\mu \omega_1 \cos\big(\omega_1 \ln(\vec{z})\big) - \mu \omega_2 \cos\big(\omega_2 \ln(\vec{z})\big) = 1 
    \label{eq:extermum}
\end{align}

This clearly shows that when the transformation function $\mathbb{T}(\cdot)$ is applied and $\omega_1$ and $\omega_2$ are nonzero, the optimum of each dimension can be found by solving $\eqref{eq:extermum}$ separately for each dimension since all operations in \eqref{eq:extermum} are element-wise.


\section{Summary of Parameters in GNBG}
\label{sec:sup:GNBGparameters}

This section provides a summary of the parameters, functions, and notations used in GNBG:

\begin{itemize}
    \item $f(\cdot)$: GNBG's baseline function.
    \item $d$: Number of dimensions.
    \item $\mathbb{X}$: $d$-dimensional search space.
    \item $\vec{x}$: A $d$-dimensional solution $(x_1,x_2,\ldots,x_d)$ in the search space $\mathbb{X}$.
    \item $l_i$: Lower bound of search range in the $i$-th dimension.
    \item $u_i$: Upper bound of search range in the $i$-th dimension.
    \item $o$: Number of components.
    \item $\mathrm{min}(\cdot)$: Defines the basin of attraction of components.
    \item $\vec{m}_{k}$: Minimum position of the $k$-th component.
    \item $\sigma_{k}$: Minimum value of the $k$-th component, i.e., $f(\vec{m}_{k})=\sigma_{k}$.
    \item $\mathbf{H}_{k}$: A $d \times d$ diagonal matrix, i.e., $\mathbf{H}_{k} = \mathrm{diag}(h_1,h_2, \ldots,h_d)\in \mathbb{R}^{d \times d}$, where $h_i = \mathbf{H}_{k}(i,i)$. The principal diagonal elements of $\mathbf{H}_{k}$ are scaling factors that influence the heights of the basin associated with the $k$-th component across different dimensions. Moreover, this matrix directly affects the condition number of the $k$-th component.
    \item $\mathbf{R}_{k}$: A $d \times d$ orthogonal matrix used for rotating the $k$-th component. It is obtained by Algorithm~\ref{alg:RotationControlled}.
    \item $\bm\Theta_{k}$: A $d \times d$ matrix, whose elements are utilized to compute the rotation matrix $\mathbf{R}_{k}$ by Algorithm~\ref{alg:RotationControlled}. The elements on and below the principal diagonal of $\bm\Theta_{k}$ are zero. An element located at the $p$-th row and $q$-th column of $\bm\Theta_{k}$, denoted as $\bm\Theta_{k}(p,q)$, specifies the rotation angle for the plane $x_{p}–x_{q}$, where $p < q$. In essence, $\bm\Theta_{k}(p,q)$ governs the interaction between variables $x_p,x_q\in \vec{x}$.
    \item $\mathfrak{p}_k$: In cases where $\bm\Theta_{k}$ is randomly generated, $0 \leq \mathfrak{p}_k \leq 1$ controls the random generation of elements above the principal diagonal in the $\bm\Theta_{k}$ matrix. For each $\bm\Theta_{k}(p,q)$ where $p<q$, a random number is drawn from a uniform distribution. If this number is less than or equal to $\mathfrak{p}_k$, $\bm\Theta_{k}(p,q)$ is set to zero. Otherwise, $\bm\Theta_{k}(p,q)$ is assigned a predefined or a random angle.
    \item $\lambda_{k}$: A positive constant that affects the rate at which the basin of the $k$-th component increases. The specific pattern can range from super-linear ($\lambda_{k}>0.5$) to linear ($\lambda_{k}=0.5$) to sub-linear ($0<\lambda_{k}<0.5$).
    \item $\mathbb{T}(\cdot)$: An element-wise non-linear transformation that plays a role in controlling the modality, irregularity, roughness, and symmetry of each component. The specific characteristics of the $k$-th component are determined by the values of $\bm\mu_{k}$ and $\bm\omega_{k}$ within the transformation function.
    \item $\bm\mu_{k}$: The vector $\bm\mu_{k}$ consists of two elements $(\mu_{k,1},\mu_{k,2})$.
     These elements control the depth of the local optima within the basin of the $k$-th component. 
     By assigning different values to $\mu_{k,1}$ and $\mu_{k,2}$, an asymmetry is introduced into the basin of the $k$-th component.
    \item $\bm\omega_{k}$: The vector $\bm\omega_{k}$ consists of four elements $(\omega_{k,1},\omega_{k,2},\omega_{k,3},\omega_{k,4})$. 
    Together with the $\bm\mu_{k}$ values, these elements play a significant role in shaping the characteristics of the local optima within the basin of the $k$-th component. 
    The values of $\omega_{k,1},\omega_{k,2},\omega_{k,3}$, and $\omega_{k,4}$ contribute to determining the number and width of local optima within the basin of the $k$-th component.
     Furthermore, differences between these elements impact the symmetry of the basin of the $k$-th component.
\end{itemize}

\section{Impact of various degrees of GNBG's controllable characteristics on the performance of algorithms}
\label{sec:sup:ImpactONalgorithms}
In this section, we present a preliminary investigation into the impact of various degrees of problem characteristics generated by GNBG, discussed in Sections~\ref{sub:sub:linearity} through \ref{sub:sub:MultipleComponents}, on the performance of three well-known optimization algorithms: 
\begin{itemize}
    \item {Pattern Search (PS)}~\cite{torczon1997convergence}: A direct search method and an example of single-solution-based techniques.
    \item {Particle Swarm Optimization (PSO)}~\cite{kennedy1995particle,bonyadi2017particle}: A well-known swarm intelligence algorithm and a population-based method.
    \item {Differential Evolution (DE)}~\cite{storn1997differential,das2010differential}: A representative evolutionary algorithm, also population-based.
\end{itemize}
Note that the experiments conducted in this section aim to gain preliminary insights into the impact of different degrees of problem characteristics on the performance of the aforementioned optimization algorithms.
While we aim to provide initial findings in this section, conducting a comprehensive empirical study, involving a wider range of algorithms and problem instances, is beyond the scope of this paper and is a potential avenue for future work. 
However, this preliminary investigation lays the foundation for future research in exploring the performance of other algorithms with GNBG-generated problem characteristics.

We employed the generalized PS with adaptive mesh and orthogonal polling directions~\cite{abramson2009orthomads,audet2002analysis} in the experiments. 
We configured PSO with a constriction factor and global star neighborhood topology, setting $c_1 = c_2 = 2.05$ and $\chi = 0.729843788$~\cite{eberhart2001comparing}. 
For DE, the strategy adopted is $DE/rand/1/bin$ with $F$ and $Cr$ values of 0.9 and 0.5, respectively~\cite{das2010differential}. 
The population size is set to 100 for both PSO and DE.

Our experimental methodology involves repeating each test 31 times, employing distinct random seeds for the algorithms in each run. 
Each experiment is designed to examine the effects of varying a specific GNBG parameterin Sections~\ref{sub:sub:linearity} to \ref{sub:sub:MultipleComponents}. 
This variation allows us to observe algorithmic performance as they address varying intensities of a particular problem characteristic. 
Results are reported in Tables~\ref{tab:LambdaResults} to~\ref{tab:MultiComponentResults}, where, we provide the mean absolute error (and standard deviation in the parenthesis) of the best-found solution at function evaluation milestones: 100,000, 250,000, and 500,000. 
Moreover, with an acceptance threshold set at $10^{-8}$ on the objective function value\footnote{Here, the objective function value is identical to the absolute error value because we set the optimum function value to zero.}, we provide the mean number of required function evaluations to find a solution within this error bound~\cite{hansen2021coco,bartz2020benchmarking}. 
Any run reaching this threshold is labeled successful, and we subsequently report the success rate over the 31 iterations.

In Table~\ref{tab:LambdaResults}, we investigate the influence of basin linearity variations on the performance of algorithms. 
Adjusting the parameter $\lambda$ changes the basin's linearity for each component, ranging from highly sub-linear to linear, and to highly super-linear.
Our focus for the experiments outlined in this table is on unimodal problem instances, offering a targeted investigation into the effects of basin linearity (see Section~\ref{sub:sub:linearity}).
The results show the increased difficulty in exploiting problem instances with smaller $\lambda$ values. 
This observation is supported by the evident increase in function evaluations needed to meet the acceptance threshold as $\lambda$ decreases. 
Moreover, assessing the results at different function evaluation checkpoints reveals a clear decline in performance for PSO and DE when dealing with instances with lower $\lambda$ values. 
This suggests that components with sub-linear basins of attraction pose a greater challenge compared to linear and super-linear ones.

One important observation from Table~\ref{tab:LambdaResults} is the rapid convergence speeds of PS and PSO. 
Although their convergence speeds are affected by increased degrees of sub-linearity (specifically, smaller $\lambda$ values), they always reach the acceptance threshold.
In contrast, DE's performance is sensitive to increasing sub-linearity. 
When $\lambda$ is set at 0.1, DE struggles to find an acceptable solution even after 500,000 function evaluations.

\begin{table*}[!t]
    \caption{Impact of various values of $\lambda$ on the performance of algorithms.
For the remaining GNBG parameters, we set $o=1$, $d=30$, $\vec{m}= \{ m_i = 0 \,|\, i = 1,2,\ldots, d \}$, $\sigma=0$, $\bm\mu=(0,0)$, $\bm\omega=(0,0,0,0)$, $\mathbf{R}_{d \times d} = \mathbf{H} = \mathbf{I}_{d \times d}$, and the search space is bounded to [-100,100].}
    \center
    \scriptsize
    \begin{adjustbox}{max width=\textwidth}
    \begin{tabular}{llccccc}
        \toprule
        \multirow{2}{*}{Alg.}  & \multirow{2}{*}{$\lambda$} & \multicolumn{3}{c}{Average of absolute error at}         & \multirow{2}{*}{Average FE to success}       &        \multirow{2}{*}{Success rate}            \\ \cmidrule{3-5}
                                                  &                                            &100,000 FE & 250,000 FE & 500,000 FE   &                                                                       &            \\ 
                                                  \midrule 
        \multirow{9}{*}{PS}            & 0.10                                     &  \makecell{0\\(0)} & \makecell{0\\(0)} & \makecell{0\\(0)} & \makecell{72382.12\\(1903.93)} & 100 \\
                                                  & 0.25                                     &  \makecell{0\\(0)} & \makecell{0\\(0)} & \makecell{0\\(0)} & \makecell{72533.03\\(1780.78)} & 100 \\
                                                  & 0.50                                     &  \makecell{0\\(0)} & \makecell{0\\(0)} & \makecell{0\\(0)} & \makecell{48158.00\\(1246.49)} & 100 \\
                                                  & 0.75                                     &  \makecell{0\\(0)} & \makecell{0\\(0)} & \makecell{0\\(0)} & \makecell{35463.09\\(1255.63)} & 100 \\
                                                  & 1.00                                     &  \makecell{0\\(0)} & \makecell{0\\(0)} & \makecell{0\\(0)} & \makecell{29062.74\\(1003.90)} & 100 \\

                                                  \midrule
        \multirow{9}{*}{PSO}         & 0.10                                     & \makecell{0.0069\\(0.0011)} & \makecell{8.43e-07\\(1.90e-07)} & \makecell{2.59e-13\\(9.45e-14)} & \makecell{323428.09\\(5423.11)} & 100 \\
                                                  & 0.25                                     &  \makecell{4.25e-06\\(2.12e-06)} & \makecell{7.08e-16\\(5.64e-16)} & \makecell{4.34e-32\\(4.1896e-32)} & \makecell{140119.51\\(3766.76)} & 100 \\
                                                  & 0.50                                     &  \makecell{2.17e-11\\(1.60e-11)} & \makecell{7.47e-31\\(1.07e-30)} & \makecell{3.70e-63\\(1.1108e-62)} & \makecell{78339.70\\(2503.35)} & 100 \\
                                                  & 0.75                                     &  \makecell{1.07e-16\\(2.03e-16)} & \makecell{7.59e-46\\(2.49e-45)} & \makecell{5.26e-94\\(1.5645e-93)} & \makecell{57584.54\\(2701.37)} & 100 \\
                                                  & 1.00                                     &  \makecell{2.34e-22\\(4.52e-22)} & \makecell{2.62e-61\\(5.55e-61)} & \makecell{3.25e-126\\(8.80e-126)} & \makecell{46538.09\\(1617.94)} & 100 \\

                                                  \midrule
        \multirow{9}{*}{DE}            & 0.10                                     & \makecell{0.1788\\(0.0107)} & \makecell{0.0028\\(0.0003)} & \makecell{2.96e-06\\(3.63e-07)} & -- & 0 \\
                                                  & 0.25                                     &  \makecell{0.0134\\(0.0021)} & \makecell{4.40e-07\\(8.22e-08)} & \makecell{1.51e-14\\(5.13e-15)} & \makecell{304393.06\\(2973.64)} & 100 \\
                                                  & 0.50                                     &  \makecell{0.0002\\(5.29e-05)} & \makecell{2.06e-13\\(1.02e-13)} & \makecell{2.66e-28\\(1.60e-28)} & \makecell{171144.48\\(2426.48)} & 100 \\
                                                  & 0.75                                     &  \makecell{3.24e-06\\(1.96e-06)} & \makecell{1.24e-19\\(1.04e-19)} & \makecell{4.67e-42\\(5.19e-42)} & \makecell{127166.41\\(3162.47)} & 100 \\
                                                  & 1.00                                     &  \makecell{4.59e-08\\(2.81e-08)} & \makecell{6.89e-26\\(8.38e-26)} & \makecell{1.23e-55\\(2.39e-55)} & \makecell{104753.80\\(2409.21)} & 100 \\
                                                  \bottomrule
    \end{tabular}
    \end{adjustbox}
    \label{tab:LambdaResults}
\end{table*}

The next controllable attribute for each component is its conditioning. 
This can be adjusted by specifying different values for the principal diagonal elements of the matrix $\mathbf{H}$. 
Table~\ref{tab:ConditioningResults} shows the effects of various condition numbers, leading to different levels of ill-conditioning, on the performance of algorithms. 
The experiments in this table focus on unimodal problem instances, allowing a detailed exploration of the effects of conditioning (see Section~\ref{sub:sub:conditioning}). 
A clear trend from the results is the increasing challenge posed to the algorithms as the condition number of problem instances increases. 
This is shown by the rise in function evaluations needed to reach the acceptance threshold with higher condition numbers. 
Additionally, a careful assessment of outcomes at specific function evaluation points indicates a clear decline in PSO and DE performance when faced with higher condition numbers. 
Another observation is that PSO sometimes shows stagnation issues when the condition number reaches $10^5$ or higher.

\begin{table*}[!t]
    \caption{Impact of various conditioning (represented by the condition number of $\mathbf{H}$) on the performance of optimization algorithms. 
The fixed GNBG parameters are: $o=1$, $d=30$, $\vec{m}= \{ m_i = 0 \,|\, i = 1,2,\ldots, d \}$, $\sigma=0$, $\lambda=1$, $\bm\mu=(0,0)$, $\bm\omega=(0,0,0,0)$, and $\mathbf{R}_{d \times d} = \mathbf{I}_{d \times d}$ with search space constrained to [-100,100]. 
For the Beta distribution, which randomizes the principal diagonal of $\mathbf{H}$, both $\alpha$ and $\beta$ are set to 0.4. 
Note that a consistent random seed is applied to the Beta distribution across all runs.}
    \center
    \scriptsize
    \begin{adjustbox}{max width=0.95\textwidth}
    \begin{tabular}{llccccc}
        \toprule
        \multirow{2}{*}{Alg.}  & \multirow{2}{*}{Condition$\#$} & \multicolumn{3}{c}{Average of absolute error at}         & \multirow{2}{*}{Average FE to success}       &        \multirow{2}{*}{Success rate}            \\ \cmidrule{3-5}
                                                  &                                            &100,000 FE & 250,000 FE & 500,000 FE   &                                                                       &            \\ 
                                                  \midrule 
        \multirow{15}{*}{PS}            & 1                                     &  \makecell{0\\(0)} & \makecell{0\\(0)} & \makecell{0\\(0)} & \makecell{29062.74\\(1003.90)} & 100 \\
                                                  & 10                                   &  \makecell{0\\(0)} & \makecell{0\\(0)} & \makecell{0\\(0)} & \makecell{31262.38\\(1226.66)} & 100 \\
                                                  & $10^2$                             &  \makecell{0\\(0)} & \makecell{0\\(0)} & \makecell{0\\(0)} & \makecell{33226.45\\(1248.04)} & 100 \\
                                                  & $10^3$                             &  \makecell{0\\(0)} & \makecell{0\\(0)} & \makecell{0\\(0)} & \makecell{35694.80\\(1198.01)} & 100 \\
                                                  & $10^4$                             &  \makecell{0\\(0)} & \makecell{0\\(0)} & \makecell{0\\(0)} & \makecell{37590.77\\(1392.36)} & 100 \\
                                                  & $10^5$                             &  \makecell{0\\(0)} & \makecell{0\\(0)} & \makecell{0\\(0)} & \makecell{40329.41\\(1221.96)} & 100 \\
                                                  & $10^6$                             &  \makecell{0\\(0)} & \makecell{0\\(0)} & \makecell{0\\(0)} & \makecell{42726.35\\(1244.90)} & 100 \\
                                                  & $10^7$                             &  \makecell{0\\(0)} & \makecell{0\\(0)} & \makecell{0\\(0)} & \makecell{45317.61\\(1382.11)} & 100 \\
                                                  \midrule
        \multirow{15}{*}{PSO}         & 1                                     &   \makecell{2.34e-22\\(4.52e-22)} & \makecell{2.62e-61\\(5.55e-61)} & \makecell{3.25e-126\\(8.80e-126)} & \makecell{46538.09\\(1617.94)} & 100 \\
                                                  & 10                                   &  \makecell{1.42e-21\\(2.25e-21)} & \makecell{1.28e-59\\(3.20e-59)} & \makecell{7.36e-124\\(3.21e-123)} & \makecell{49473.61\\(1548.64)} & 100 \\
                                                  & $10^2$                             &  \makecell{2.91e-18\\(1.38e-17)} & \makecell{1.06e-56\\(5.57e-56)} & \makecell{1.20e-122\\(5.18e-122)} & \makecell{55086.16\\(3907.14)} & 100 \\
                                                  & $10^3$                             &  \makecell{7.25e-16\\(4.02e-15)} & \makecell{3.42e-57\\(1.03e-56)} & \makecell{2.83e-122\\(7.39e-122)} & \makecell{59079.80\\(4484.42)} & 100 \\
                                                  & $10^4$                             &  \makecell{6.87e-17\\(2.52e-16)} & \makecell{1.16e-55\\(4.22e-55)} & \makecell{7.91e-120\\(3.80e-11)} & \makecell{63156.87\\(3645.66)} & 100 \\
                                                  & $10^5$                             &  \makecell{322.5806\\(1796.0530)} & \makecell{322.5806\\(1796.0530)} & \makecell{322.5806\\(1796.0530)} & \makecell{67873.96\\(3505.80)} & 96.77 \\
                                                  & $10^6$                             &  \makecell{645.1613\\(2497.3104)} & \makecell{645.1613\\(2497.3104)} & \makecell{645.1613\\(2497.3104)} & \makecell{72797.48\\(5149.41)} & 93.54 \\
                                                  & $10^7$                             &  \makecell{322.5806\\(1796.0530)} & \makecell{322.5806\\(1796.0530)} & \makecell{322.5806\\(1796.0530)} & \makecell{77139.33\\(6396.48)} & 96.77 \\
                                                  \midrule
        \multirow{15}{*}{DE}            & 1                                     &  \makecell{4.59e-08\\(2.81e-08)} & \makecell{6.89e-26\\(8.38e-26)} & \makecell{1.23e-55\\(2.39e-55)} & \makecell{104753.80\\(2409.21)} & 100 \\
                                                  & 10                                   &  \makecell{1.56e-07\\(1.37e-07)} & \makecell{2.50e-25\\(3.87e-25)} & \makecell{3.34e-55\\(4.84e-55)} & \makecell{109257.19\\(2219.66)} & 100 \\
                                                  & $10^2$                             &  \makecell{1.29e-06\\(8.49e-07)} & \makecell{2.26e-24\\(2.48e-24)} & \makecell{2.60e-54\\(4.48e-54)} & \makecell{117002.19\\(2287.84)} & 100 \\
                                                  & $10^3$                             &  \makecell{9.78e-06\\(5.60e-06)} & \makecell{1.20e-23\\(1.37e-23)} & \makecell{2.06e-53\\(3.06e-53)} & \makecell{124048.58\\(2728.30)} & 100 \\
                                                  & $10^4$                             &  \makecell{0.0001\\(7.21e-05)} & \makecell{1.30e-22\\(1.21e-22)} & \makecell{3.16e-52\\(3.70e-52)} & \makecell{133027.70\\(2543.55)} & 100 \\
                                                  & $10^5$                             &  \makecell{0.0009\\(0.0005)} & \makecell{1.1e-21\\(1.08e-21)} & \makecell{1.49e-51\\(1.63e-51)} & \makecell{140942.16\\(2467.07)} & 100 \\
                                                  & $10^6$                             &  \makecell{0.0070\\(0.0046)} & \makecell{1.13e-20\\(1.76e-20)} & \makecell{1.24e-50\\(1.51e-50)} & \makecell{148441.58\\(2736.83)} & 100 \\
                                                  & $10^7$                             &  \makecell{0.0601\\(0.0286)} & \makecell{7.67e-20\\(9.44e-20)} & \makecell{1.28e-49\\(2.06e-49)} & \makecell{156012.54\\(2733.30)} & 100 \\
                                                  \bottomrule
    \end{tabular}
    \end{adjustbox}
    \label{tab:ConditioningResults}
\end{table*}

The structure of variable interactions stands as an important aspect of optimization problems, significantly influencing the performance of algorithms. 
Table~\ref{tab:SeparabilityResults} shows the results obtained by the algorithms on problem instances with various variable interaction structures: from fully separable (where $\mathfrak{p}=0$) to fully connected non-separable configurations (where $\mathfrak{p}=1$). 
Using the GNBG parameter settings outlined in Section~\ref{sub:sub:interactions}, we generated instances specifically designed to evaluate the impact of various degrees of connectivity in variable interaction structures on algorithm performance.
As the value of $\mathfrak{p}$ increases, the complexity of the variable interaction structure grows, significantly impacting the performance of algorithms. 
While the convergence rates of PS and DE decline when solving instances with more complex variable interaction structures, they still find acceptable solutions, though with more function evaluations. 
In contrast, PSO struggles with increased complexity in variable interaction structures.
This is shown by its occasional stagnation or very slow convergence rate, preventing it from finding an acceptable solution within 500,000 function evaluations. 
Additionally, the success rate of PSO decreases with the increasing complexity of the variable interaction structure due to higher $\mathfrak{p}$ values.

\begin{table*}[!t]
    \caption{Performance impact of varying $\mathfrak{p}$ values, dictating degrees of variable interaction structure connectivity, on optimization algorithms. 
Fixed GNBG parameters include: $o=1$, $d=30$, $\vec{m}= \{ m_i = 0 \,|\, i = 1,2,\ldots, d \}$, $\sigma=0$, $\lambda=1$, $\bm\mu=(0,0)$, and $\bm\omega=(0,0,0,0)$, with the search space bound to [-100,100]. 
The principal diagonal of $\mathbf{H}$ is randomized using a uniform distribution in [1,100] to induce moderate ill-conditioning and make component rotation dependent. 
A consistent random seed is applied for both $\mathbf{H}$ and $\mathfrak{p}$ randomizations across runs.}
    \center
    \scriptsize
    \begin{adjustbox}{max width=\textwidth}
    \begin{tabular}{lcccccc}
        \toprule
        \multirow{2}{*}{Alg.}  & \multirow{2}{*}{$\mathfrak{p}$} & \multicolumn{3}{c}{Absolute error at}         & \multirow{2}{*}{Average FE to success}       &        \multirow{2}{*}{Success rate}            \\ \cmidrule{3-5}
                                                  &                                            &100,000 FE & 250,000 FE & 500,000 FE   &                                                                       &            \\ 
                                                  \midrule 
        \multirow{9}{*}{PS}            & 0.00                                    &   \makecell{0\\(0)} & \makecell{0\\(0)} & \makecell{0\\(0)} & \makecell{33405.09\\(1061.08)} & 100 \\
                                                  & 0.25                                    &   \makecell{0.057616\\(0.0829)} & \makecell{1.63e-10\\(3.71e-10)} & \makecell{1.81e-24\\(7.72e-24)} & \makecell{203970.61\\(22421.50)} & 100 \\
                                                  & 0.50                                    &   \makecell{0.011233\\(0.0157)} & \makecell{1.50e-12\\(1.92e-12)} & \makecell{6.81e-29\\(1.64e-28)} & \makecell{186118.77\\(10970.198)} & 100 \\
                                                  & 0.75                                    &   \makecell{0.037089\\(0.0354)} & \makecell{4.52e-11\\(6.40e-11)} & \makecell{2.84e-26\\(4.44e-26)} & \makecell{203580.54\\(15404.26)} & 100 \\
                                                  & 1.00                                    &   \makecell{0.090223\\(0.1733)} & \makecell{5.85e-10\\(1.01e-09)} & \makecell{1.08e-23\\(2.86e-23)} & \makecell{213361.41\\(21342.67)} & 100 \\
                                                  \midrule
        \multirow{9}{*}{PSO}         & 0.00                                    &   \makecell{2.3459e-18\\(1.2246e-17)} & \makecell{7.1694e-58\\(2.5105e-57)} & \makecell{1.9225e-123\\(8.4677e-123)} & \makecell{55584.16\\(3610.69)} & 100 \\
                                                  & 0.25                                    &  \makecell{9.0549\\(11.9187)} & \makecell{0.0050\\(0.0062)} & \makecell{7.8924e-07\\(1.5512e-06)} & \makecell{454501.70\\(35548.89)} & 32.25 \\
                                                  & 0.50                                    &  \makecell{25.4761\\(28.5542)} & \makecell{0.0611\\(0.1141)} & \makecell{1.2766e-05\\(2.1924e-05)} & \makecell{458912.37\\(38169.85)} & 22.58 \\
                                                  & 0.75                                    &  \makecell{22.5156\\(20.0315)} & \makecell{0.0571\\(0.1871)} & \makecell{5.1357e-06\\(3.7560e-05)} & \makecell{458343.00\\(34410.20)} & 12.90 \\
                                                  & 1.00                                    &  \makecell{17.1599\\(18.5575)} & \makecell{0.0174\\(0.0237)} & \makecell{9.2786e-06\\(2.2028e-05)} & \makecell{490021.50\\(11434.62)} & 6.45 \\

                                                  \midrule
        \multirow{9}{*}{DE}            & 0.00                                    &   \makecell{1.38e-06\\(8.69e-07)} & \makecell{1.91e-24\\(1.74e-24)} & \makecell{3.99e-54\\(7.02e-54)} & \makecell{117446.22\\(2673.65)} & 100 \\
                                                  & 0.25                                    &   \makecell{0.8026\\(0.9358)} & \makecell{1.13e-07\\(3.78e-07)} & \makecell{3.84e-19\\(1.99e-18)} & \makecell{247990.00\\(23662.44)} & 100 \\
                                                  & 0.50                                    &   \makecell{1.5307\\(1.9609)} & \makecell{6.36e-08\\(8.41e-08)} & \makecell{3.07e-19\\(1.42e-18)} & \makecell{258567.16\\(17237.18)} & 100 \\
                                                  & 0.75                                    &   \makecell{0.6017\\(0.9387)} & \makecell{2.75e-09\\(4.07e-09)} & \makecell{1.19e-22\\(2.1e-22)} & \makecell{234734.48\\(10813.28)} & 100 \\
                                                  & 1.00                                    &   \makecell{0.8036\\(0.7247)} & \makecell{3.92e-08\\(7.48e-08)} & \makecell{1.35e-20\\(4.28e-20)} & \makecell{250733.80\\(15214.93)} & 100 \\
                                                  \bottomrule
    \end{tabular}
    \end{adjustbox}
    \label{tab:SeparabilityResults}
\end{table*}

In our experiments evaluating the impact of variable interaction structures, we use randomized angles for each connection in $\bm\Theta$. 
These angles determine the strength of the connections between variables.
Table~\ref{tab:AngleResults} shows the influence of three specific angles---0, $\frac{5\pi}{180}$, and $\frac{\pi}{4}$---on the performance of algorithms. 
Following the GNBG parameter settings outlined in Section~\ref{sub:sub:interactions}, we set $\mathfrak{p}=1$, meaning all elements above the principal diagonal of $\bm\Theta$ take the chosen angle. 
Unsurprisingly, the best results occur when the angle is zero, maintaining the component's original variable interaction and creating a fully separable structure.
Assigning angles of $\frac{5\pi}{180}$ and $\frac{\pi}{4}$ changes the component's interaction structure into a non-separable, fully-connected state, increasing the problem's complexity. 
However, the degree of this increase depends on the strength of variable interactions.
The results indicate that the problem is easier to solve when the angle is set to $\frac{5\pi}{180}$ compared to $\frac{\pi}{4}$. 
This suggests that components become easier to exploit as angles approach axes-aligned values (i.e., angles close to $k\frac{\pi}{2}$ where $k \in \mathbb{Z}$).

\begin{table*}[!t]
    \caption{Impact of various angles on the performance of algorithms.
The angles determine interaction strength within a fully-connected variable interaction structure. 
The remaining GNBG settings are: $o=1$, $d=30$, $\vec{m}= \{ m_i = 0 \,|\, i = 1,2,\ldots, d \}$, $\sigma=0$, $\lambda=1$, $\bm\mu=(0,0)$, $\bm\omega=(0,0,0,0)$ with search boundaries [-100,100]. 
The uniform distribution randomizes $\mathbf{H}$'s diagonal within [1,100] to introduce moderate ill-conditioning and dependency in component rotation. 
A consistent random seed is applied for  $\mathbf{H}$ randomizations across runs.}
    \center
    \scriptsize
    \begin{adjustbox}{max width=\textwidth}
    \begin{tabular}{lcccccc}
        \toprule
        \multirow{2}{*}{Alg.}  & \multirow{2}{*}{Angle} & \multicolumn{3}{c}{Absolute error at}         & \multirow{2}{*}{Average FE to success}       &        \multirow{2}{*}{Success rate}            \\ \cmidrule{3-5}
                                                  &                                            &100,000 FE & 250,000 FE & 500,000 FE   &                                                                       &            \\ 
                                                  \midrule 
        \multirow{5}{*}{PS}            & 0                                        & \makecell{0\\(0)} & \makecell{0\\(0)} & \makecell{0\\(0)} & \makecell{33405.09\\(1061.08)} & 100 \\
                                                  & $5\frac{\pi}{180}$               & \makecell{0.0001\\(0.0001)} & \makecell{2.87e-17\\(1.21e-16)} & \makecell{3.51e-39\\(1.70e-38)} & \makecell{140831.58\\(10479.61)} & 100 \\
                                                  & $\frac{\pi}{4}$                     & \makecell{0.0867\\(0.0944)} & \makecell{1.46e-10\\(1.73e-10)} & \makecell{1.09e-24\\(1.80e-24)} & \makecell{211963.00\\(12301.48)} & 100 \\
                                                  \midrule
        \multirow{5}{*}{PSO}         & 0                                        & \makecell{2.34e-18\\(1.22e-17)} & \makecell{7.16e-58\\(2.51e-57)} & \makecell{1.92e-123\\(8.46e-123)} & \makecell{55584.16\\(3610.69)} & 100 \\
                                                  & $5\frac{\pi}{180}$               & \makecell{1777.687\\(9896.2039)} & \makecell{0.0054\\(0.0304)} & \makecell{2.20e-11\\(1.189e-10)} & \makecell{309116.29\\(44254.11)} & 100 \\
                                                  & $\frac{\pi}{4}$                     & \makecell{44.8274\\(49.1275)} & \makecell{0.1231\\(0.1691)} & \makecell{5.63e-05\\(8.07e-05)} & -- & 0 \\
                                                  \midrule
        \multirow{5}{*}{DE}            & 0                                        & \makecell{1.38e-06\\(8.69e-07)} & \makecell{1.91e-24\\(1.74e-24)} & \makecell{3.99e-54\\(7.02e-54)} & \makecell{117446.22\\(2673.65)} & 100 \\ 
                                                  & $5\frac{\pi}{180}$               & \makecell{0.0463\\(0.0783)} & \makecell{9.25e-12\\(3.06e-11)} & \makecell{3.26e-28\\(1.28e-27)} & \makecell{192322.38\\(9867.75)} & 100 \\
                                                  & $\frac{\pi}{4}$                     & \makecell{2.7704\\(2.9605)} & \makecell{1.61e-06\\(3.72e-06)} & \makecell{4.30e-17\\(1.31e-16)} & \makecell{281307.45\\(18729.14)} & 100 \\
                                                  \bottomrule
    \end{tabular}
    \end{adjustbox}
    \label{tab:AngleResults}
\end{table*}

In prior experiments, we employed unimodal components by setting all elements of $\bm\mu$ and $\bm\omega$ to zero. 
This configuration neutralized the $\mathbb{T}$ transformation. 
Table~\ref{tab:MultimodalityResults} examines the effects of varying $\bm\mu$ and $\bm\omega$ on algorithmic performance, with GNBG parameter adjustments described in Section~\ref{sub:sub:multimodality}.
The analysis indicates that better results are obtained when both $\bm\mu$ and $\bm\omega$ are set to smaller values, which correspond to fewer and shallower local optima. 
In addition, higher $\bm\mu$ values create more difficult instances, marked by deeper local optima. 
These depths increase the risk of premature convergence: deeper local optima require longer convergence periods, during which algorithms might lose their diversity, leading to potential trapping in local optima. 
An important observation is that with larger $\bm\mu$ values, components with larger $\bm\omega$ values are more challenging than those with lower $\bm\omega$ values.
For example, when $\bm\mu$ is [0.25,0.25], an increase in $\bm\omega$ from 5 to 50 reduces the exploration capabilities of DE and PS, causing them to get stuck in local optima. 
This behavior occurs because higher $\bm\omega$ values create more local optima, increasing the problem's ruggedness. 
Additionally, the results show that all tested algorithms are likely to get trapped in local optima with $\bm\mu$ values of 0.5 or above. 
It is evident that, as $\bm\mu$ values increase, while algorithms may produce better average absolute error values, these improvements are due to the depth of local optima and the resulting lower function values at their bottoms.

\begin{table*}[!t]
    \caption{Impact of various values of $\bm\mu$ and $\bm\omega$ on the performance of algorithms.
    For the remaining GNBG parameters, we set $o=1$, $d=30$, $\vec{m}= \{ m_i = 0 \,|\, i = 1,2,\ldots, d \}$, $\sigma=0$, $\bm\mu=(0,0)$, $\lambda=1$, $\mathbf{R}_{d \times d} = \mathbf{H} = \mathbf{I}_{d \times d}$, and the search space is bounded to [-100,100].}   
    \center
    \scriptsize
    \begin{adjustbox}{max width=0.91\textwidth}
    \begin{tabular}{lccccccc}
        \toprule
        \multirow{2}{*}{Alg.}  & \multirow{2}{*}{$\bm\mu$} & \multirow{2}{*}{$\bm\omega$} & \multicolumn{3}{c}{Absolute error at}         & \multirow{2}{*}{Average FE to success}       &        \multirow{2}{*}{Success rate}            \\ \cmidrule{4-6}
                                                  &                   &                         &100,000 FE & 250,000 FE & 500,000 FE   &                                                                       &            \\ 
                                                  \midrule 
        \multirow{20}{*}{PS}            & [0.00,0.00]          &     [0,0,0,0]    & \makecell{0\\(0)} & \makecell{0\\(0)} & \makecell{0\\(0)} & \makecell{29062.74\\(1003.90)} & 100 \\
                                                  & [0.10,0.10]          &     [5,5,5,5]    & \makecell{0\\(0)} & \makecell{0\\(0)} & \makecell{0\\(0)} & \makecell{29305\\(1123.002)} & 100 \\
                                                  & [0.25,0.25]          &     [5,5,5,5]    & \makecell{0\\(0)} & \makecell{0\\(0)} & \makecell{0\\(0)} & \makecell{29609.5161\\(980.6566)} & 100 \\
                                                  & [0.50,0.50]          &     [5,5,5,5]    & \makecell{18.0895\\(12.3199)} & \makecell{18.0895\\(12.3199)} & \makecell{18.0895\\(12.3199)} & -- & 0 \\
                                                  & [0.75,0.75]          &     [5,5,5,5]    & \makecell{95.264\\(60.0162)} & \makecell{95.264\\(60.0162)} & \makecell{95.264\\(60.0162)} & -- & 0 \\
                                                  & [1.00,1.00]          &     [5,5,5,5]    & \makecell{558.9562\\(406.1115)} & \makecell{558.9562\\(406.1115)} & \makecell{558.9562\\(406.1115)} & -- & 0 \\
                                                  & [0.10,0.10]   &    [50,50,50,50]    & \makecell{0\\(0)} & \makecell{0\\(0)} & \makecell{0\\(0)} & \makecell{29208.6774\\(1051.6833)} & 100 \\
                                                  & [0.25,0.25]   &    [50,50,50,50]    & \makecell{0\\(0)} & \makecell{0\\(0)} & \makecell{0\\(0)} & \makecell{29404.5806\\(940.7578)} & 100 \\
                                                  & [0.50,0.50]   &    [50,50,50,50]    & \makecell{647.9814\\(632.4114)} & \makecell{647.9814\\(632.4114)} & \makecell{647.9814\\(632.4114)} & -- & 0 \\
                                                  & [0.75,0.75]   &    [50,50,50,50]    & \makecell{638.8819\\(395.0779)} & \makecell{638.8819\\(395.0779)} & \makecell{638.8819\\(395.0779)} & -- & 0 \\
                                                  & [1.00,1.00]   &    [50,50,50,50]    & \makecell{444.1268\\(220.0597)} & \makecell{444.1268\\(220.0597)} & \makecell{444.1268\\(220.0597)} & -- & 0 \\
                                                  \midrule
        \multirow{20}{*}{PSO}         & [0.00,0.00]          &     [0,0,0,0]    & \makecell{2.34e-22\\(4.52e-22)} & \makecell{2.62e-61\\(5.55e-61)} & \makecell{3.25e-126\\(8.80e-126)} & \makecell{46538.09\\(1617.94)} & 100 \\
                                                  & [0.10,0.10]          &     [5,5,5,5]    & \makecell{1.77e-18\\(5.10e-18)} & \makecell{8.96e-51\\(4.95e-50)} & \makecell{1.27e-108\\(6.00e-108)} & \makecell{53903.74\\(3078.32)} & 100 \\
                                                  & [0.25, 0.25]          &     [5, 5, 5, 5]    & \makecell{4035.4152\\(3934.659)} & \makecell{4035.4152\\(3934.659)} & \makecell{4035.4152\\(3934.659)} & -- & 0 \\
                                                  & [0.50, 0.50]          &     [5, 5, 5, 5]    & \makecell{11898.8015\\(5539.7218)} & \makecell{11898.8015\\(5539.7218)} & \makecell{11898.8015\\(5539.7218)} & -- & 0 \\
                                                  & [0.75, 0.75]          &     [5, 5, 5, 5]    & \makecell{7893.1048\\(2127.2207)} & \makecell{7889.9318\\(2130.3158)} & \makecell{7889.9318\\(2130.3158)} & -- & 0 \\
                                                  & [1.00, 1.00]          &     [5, 5, 5, 5]    & \makecell{3779.0472\\(1189.2953)} & \makecell{3754.5283\\(1169.8409)} & \makecell{3744.1726\\(1172.5886)} & -- & 0 \\
                                                  & [0.10, 0.10]   &    [50, 50, 50, 50]    & \makecell{15.7274\\(47.7612)} & \makecell{15.7274\\(47.7612)} & \makecell{15.7274\\(47.7612)} & \makecell{57410.625\\(3422.444)} & 51.6129 \\
                                                  & [0.25, 0.25]   &    [50, 50, 50, 50]    & \makecell{415.5687\\(980.777)} & \makecell{415.5687\\(980.777)} & \makecell{415.5687\\(980.777)} & -- & 0 \\
                                                  & [0.50, 0.50]   &    [50, 50, 50, 50]    & \makecell{4895.7732\\(3780.7165)} & \makecell{4701.8173\\(3589.9746)} & \makecell{4701.8094\\(3589.9644)} & -- & 0 \\
                                                  & [0.75, 0.75]   &    [50, 50, 50, 50]    & \makecell{5820.2932\\(2246.9847)} & \makecell{4876.6711\\(2224.7195)} & \makecell{4815.8523\\(2259.8326)} & -- & 0 \\
                                                  & [1.00, 1.00]   &    [50, 50, 50, 50]    & \makecell{2749.6023\\(1014.685)} & \makecell{2435.603\\(956.475)} & \makecell{2291.4915\\(1003.7711)} & -- & 0 \\
                                                  \midrule
        \multirow{20}{*}{DE}            & [0.00, 0.00]          &     [0, 0, 0, 0]    & \makecell{4.59e-08\\(2.81e-08)} & \makecell{6.89e-26\\(8.38e-26)} & \makecell{1.23e-55\\(2.39e-55)} & \makecell{104753.80\\(2409.21)} & 100 \\ 
                                                  & [0.10, 0.10]          &     [5, 5, 5, 5]    & \makecell{2.2493e-07\\(1.4234e-07)} & \makecell{3.5348e-24\\(3.2999e-24)} & \makecell{4.4086e-52\\(6.6203e-52)} & \makecell{110902.1613\\(2229.8682)} & 100 \\
                                                  & [0.25, 0.25]          &     [5, 5, 5, 5]    & \makecell{0.0012916\\(0.00067359)} & \makecell{1.1893e-14\\(1.4515e-14)} & \makecell{6.4408e-33\\(1.3301e-32)} & \makecell{166925.7742\\(5081.816)} & 100 \\
                                                  & [0.50, 0.50]          &     [5, 5, 5, 5]    & \makecell{880.0988\\(453.4637)} & \makecell{14.5252\\(12.9632)} & \makecell{0.032015\\(0.035991)} & -- & 0 \\
                                                  & [0.75, 0.75]          &     [5, 5, 5, 5]    & \makecell{2887.0515\\(581.4986)} & \makecell{1406.0535\\(433.1411)} & \makecell{322.9837\\(148.5842)} & -- & 0 \\
                                                  & [1.00, 1.00]          &     [5, 5, 5, 5]    & \makecell{4716.9576\\(1387.7579)} & \makecell{2134.4184\\(627.5284)} & \makecell{619.71\\(207.0421)} & -- & 0 \\
                                                  & [0.10, 0.10]   &    [50, 50, 50, 50]    & \makecell{2.8002e-07\\(1.6446e-07)} & \makecell{9.3976e-24\\(7.7403e-24)} & \makecell{3.5197e-51\\(5.6973e-51)} & \makecell{112765.9355\\(2191.2871)} & 100 \\
                                                  & [0.25, 0.25]   &    [50, 50, 50, 50]    & \makecell{0.0015645\\(0.001421)} & \makecell{9.6231e-15\\(7.3691e-15)} & \makecell{2.7353e-33\\(3.2895e-33)} & \makecell{167646.6129\\(3837.7284)} & 100 \\
                                                  & [0.50, 0.50]   &    [50, 50, 50, 50]    & \makecell{177.2765\\(86.7903)} & \makecell{0.19943\\(0.22682)} & \makecell{2.7076e-06\\(7.8304e-06)} & -- & 0 \\
                                                  & [0.75, 0.75]   &    [50, 50, 50, 50]    & \makecell{5587.9928\\(1846.9089)} & \makecell{2000.9528\\(880.6734)} & \makecell{472.9545\\(359.5097)} & -- & 0 \\
                                                  & [1.00, 1.00]   &    [50, 50, 50, 50]    & \makecell{11919.8398\\(3413.1434)} & \makecell{6632.5713\\(2143.617)} & \makecell{3237.6569\\(1430.2877)} & -- & 0 \\
                                                  \bottomrule
    \end{tabular}
    \label{tab:MultimodalityResults}
\end{adjustbox}
\end{table*}

Finally, Table~\ref{tab:MultiComponentResults} shows the influence of the number of components in the landscape on algorithm performance. 
The GNBG parameter configurations for these tests are detailed in Section~\ref{sub:sub:MultipleComponents}. 
Due to the presence of multiple promising regions, each with a substantial basin of attraction, these problems can prove exceptionally deceptive.
A deceptive search space arises when extensive, low-function-value regions\footnote{In minimization problems, low-function-values signify values of superior quality.} mislead optimization algorithms, making them appear more favorable than they truly are. 
In such situations, algorithms might become trapped in these deceptive basins, preventing them from discovering the global optimum. 
Results suggest that the presence of multiple components significantly impacts algorithmic efficiency. 
Interestingly, when $o$ is set to two, the success rates drop considerably. 
Our analyses reveal that, based on the constant random seed used for generating this instance, the component containing the global optimum occupies a smaller portion of the landscape than the other component. 
Under this deceptive scenario, algorithms mainly gravitate toward the larger but misleading promising region. 
Data from instances with more promising regions indicate that algorithms quickly converge to a deceptive region and become stuck there. 
It is important to note that the average absolute error values in these scenarios are influenced not only by the component count but also by the level of deception and the quality of the dominant deceptive components. 
Nonetheless, it is clear that algorithms struggle to find the global optimum when the number of components exceeds two.

\begin{table*}[!t]
    \caption{Assessing the performance impact of varying component numbers on optimization algorithms. 
For each component $k$, parameters are set as $d=30$, $\bm\mu_k=(0,0)$, $\lambda_k=1$, and $\mathbf{R}_k = \mathbf{I}_{d \times d}$ with search boundaries [-100,100]. 
The minimum position ($\vec{m}_k$) and value ($\sigma_k$) for each component $k$ are randomly designated within [-100,100] and [0,10], respectively. 
Furthermore, all elements on the principal diagonal of $\mathbf{H}_k$ are uniformly randomized within [0.001,0.1] (all have the same value), signifying varied sizes among components while ensuring they remain well-conditioned. 
Consistent random seeds are used across runs for GNBG instance generation.}       
    \center
    \scriptsize
    \begin{adjustbox}{max width=\textwidth}
    \begin{tabular}{lcccccc}
        \toprule
        \multirow{2}{*}{Alg.}  & \multirow{2}{*}{$o$} & \multicolumn{3}{c}{Absolute error at}         & \multirow{2}{*}{Average FE to success}       &        \multirow{2}{*}{Success rate}            \\ \cmidrule{3-5}
                                                  &                                            &100,000 FE & 250,000 FE & 500,000 FE   &                                                                       &            \\ 
                                                  \midrule 
        \multirow{11}{*}{PS}            & 1               &   \makecell{0\\(0)} & \makecell{0\\(0)} & \makecell{0\\(0)} & \makecell{29062.74\\(1003.90)} & 100 \\
                                                  & 2               &   \makecell{8.2578\\(2.2045)} & \makecell{8.2578\\(2.2045)} & \makecell{8.2578\\(2.2045)} & \makecell{30194\\(2049.1955)} & 6.4516 \\
                                                  & 5               & \makecell{44.7651\\(1.7597e-14)} & \makecell{44.7651\\(1.7597e-14)} & \makecell{44.7651\\(1.7597e-14)} & -- & 0 \\
                                                  & 10             &  \makecell{12.8672\\(7.2125)} & \makecell{12.8672\\(7.2125)} & \makecell{12.8672\\(7.2125)} & -- & 0 \\
                                                  & 25             &  \makecell{19.03\\(0.48748)} & \makecell{19.03\\(0.48748)} & \makecell{19.03\\(0.48748)} & -- & 0 \\
                                                  & 50             &  \makecell{17.358\\(14.8416)} & \makecell{17.358\\(14.8416)} & \makecell{17.358\\(14.8416)} & -- & 0 \\
                                                  \midrule
        \multirow{11}{*}{PSO}         & 1               &   \makecell{2.34e-22\\(4.52e-22)} & \makecell{2.62e-61\\(5.55e-61)} & \makecell{3.25e-126\\(8.80e-126)} & \makecell{46538.09\\(1617.94)} & 100 \\
                                                  & 2               &   \makecell{7.4035\\(3.3003)} & \makecell{7.4035\\(3.3003)} & \makecell{7.4035\\(3.3003)} & \makecell{52109\\(5809.1577)} & 16.129 \\
                                                  & 5               & \makecell{99.7869\\(306.3483)} & \makecell{99.7869\\(306.3483)} & \makecell{99.7869\\(306.3483)} & -- & 0 \\
                                                  & 10             &  \makecell{13.7992\\(6.8719)} & \makecell{13.7992\\(6.8719)} & \makecell{13.7992\\(6.8719)} & -- & 0 \\
                                                  & 25             &  \makecell{43.5034\\(136.2469)} & \makecell{43.5034\\(136.2469)} & \makecell{43.5034\\(136.2469)} & -- & 0 \\
                                                  & 50             &   \makecell{20.5219\\(14.7587)} & \makecell{20.5219\\(14.7587)} & \makecell{20.5219\\(14.7587)} & -- & 0 \\
                                                  \midrule
        \multirow{11}{*}{DE}            & 1               &   \makecell{4.59e-08\\(2.81e-08)} & \makecell{6.89e-26\\(8.38e-26)} & \makecell{1.23e-55\\(2.39e-55)} & \makecell{104753.80\\(2409.21)} & 100 \\ 
                                                  & 2               &    \makecell{8.5426\\(1.5854)} & \makecell{8.5426\\(1.5854)} & \makecell{8.5426\\(1.5854)} & \makecell{114346\\(0)} & 3.2258 \\
                                                  & 5               & \makecell{44.7651\\(4.47e-08)} & \makecell{44.7651\\(1.44e-14)} & \makecell{44.7651\\(1.44e-14)} & -- & 0 \\
                                                  & 10             &  \makecell{17.111\\(2.03e-08)} & \makecell{17.111\\(1.08e-14)} & \makecell{17.111\\(1.08e-14)} & -- & 0 \\
                                                  & 25             &  \makecell{18.9424\\(2.01e-08)} & \makecell{18.9424\\(3.61e-15)} & \makecell{18.9424\\(3.61e-15)} & -- & 0 \\
                                                  & 50             &   \makecell{12.997\\(15.3686)} & \makecell{12.997\\(15.3686)} & \makecell{12.997\\(15.3686)} & -- & 0 \\
                                                  \bottomrule
    \end{tabular}
    \end{adjustbox}
    \label{tab:MultiComponentResults}
\end{table*}

\section{A GNBG-Generated Test Suite}
\label{sec:sup:Suite}

This section introduces a set of 24 problem instances, denoted as $f_{1}$ through $f_{24}$, all generated using GNBG.
These instances cover a wide range of problem features, including varying degrees of modality, ruggedness, symmetry, conditioning, variable interaction structures, basin linearity, and deceptiveness.
By presenting these problems, we provide researchers with a platform to assess the strengths and weaknesses of their algorithms against challenges with known, controlled characteristics.
These instances can be categorized as:
\begin{itemize}
    \item $f_{1}$ to $f_{6}$: Unimodal instances,
    \item $f_{7}$ to $f_{15}$: Multimodal instances with a single component, and
    \item $f_{16}$ to $f_{24}$: Multimodal instances with multiple components.
\end{itemize}
All instances are defined in a 30-dimensional solution space with search boundaries set within $[-100, 100]^d$.
An overview of the 24 problem instances is presented in Table~\ref{tab:SuiteOverview}. 
The MATLAB  and Python source codes for these problem instances are available for access~\cite{yazdani2024GNBsuiteMatlab,yazdani2024GNBsuitePython}. 
The following subsections give a detailed overview of each problem instance.

\begin{table}[t!] 
\footnotesize
\centering
  \caption{An overview of characteristics of the presented test suite containing 24 problem instances generated by GNBG. }  \label{tab:SuiteOverview}
\Rotatebox{90}{
\resizebox{1.22\textwidth}{!}{%
 \begin{threeparttable}
  \begin{tabular}{lcccccccccccccccccccccccccc}
    \toprule
   \multirow{2}{*}{Characteristic} & \multicolumn{24}{c}{Problem instances}\\ 
    \cmidrule{2-25} 
   & $f_{1}$ & $f_{2}$ & $f_{3}$ & $f_{4}$ & $f_{5}$ & $f_{6}$ & $f_{7}$ & $f_{8}$ & $f_{9}$ & $f_{10}$ & $f_{11}$ & $f_{12}$ & $f_{13}$ & $f_{14}$ & $f_{15}$ & $f_{16}$ & $f_{17}$ & $f_{18}$ & $f_{19}$ & $f_{20}$ & $f_{21}$ & $f_{22}$ & $f_{23}$& $f_{24}$  \\
\midrule
  Modality& \multicolumn{6}{c}{Unimodal}&   \multicolumn{9}{c}{\cellcolor{gray!15}Multimodal with single component}& \multicolumn{9}{c}{Multimodal with multiple competing components}\\
\midrule
   Basin local optima\tnote{$\bullet$}& \XM & \XM & \XM & \XM & \XM & \XM & \CM & \CM & \CM & \CM & \CM & \CM & \CM & \CM & \CM & \XM & \XM  & \CM & \CM & \CM & \CM  & \CM & \CM & \CM \\
\midrule
   Separability & $\mathcal{F}$\tnote{$\bm \dagger$} & $\mathcal{F}$ & $\mathcal{F}$ & $\mathcal{N}$\tnote{$\bm \dagger$} & $\mathcal{N}$ & $\mathcal{N}$ & $\mathcal{F}$ & $\mathcal{F}$ & $\mathcal{F}$ & $\mathcal{F}$ & $\mathcal{N}$ & $\mathcal{P}$\tnote{$\bm \dagger$} & $\mathcal{N}$ & $\mathcal{N}$ & $\mathcal{N}$ & $\mathcal{N}$ & $\mathcal{N}$ & $\mathcal{N}$ &  $\mathcal{N}$ & $\mathcal{N}$ & $\mathcal{N}$ & $\mathcal{N}$ & $\mathcal{N}$& $\mathcal{N}$ \\
   \midrule
   Varying variable interactions &  \XM & \XM & \XM & \XM & \XM & \XM & \XM & \XM & \XM & \XM & \XM & \XM & \XM & \XM & \XM & \XM & \CM & \CM & \CM & \CM & \CM & \CM & \CM & \CM  \\
     \midrule
   Symmetry &  $\mathcal{S}$\tnote{$\bm\ddagger$} &  $\mathcal{S}$ &  $\mathcal{S}$ &  $\mathcal{S}$ &  $\mathcal{S}$ &  $\mathcal{S}$ &  $\mathcal{S}$ &  $\mathcal{S}$ &  $\mathcal{S}$ &  $\mathcal{A}$\tnote{$\bm\ddagger$} &  $\mathcal{A}$ &  $\mathcal{A}$ &   $\mathcal{S}$  & $\mathcal{A}$ &  $\mathcal{S}$ &  $\mathcal{A}$ &  $\mathcal{A}$ &   $\mathcal{A}$ &  $\mathcal{A}$ &  $\mathcal{A}$ &  $\mathcal{A}$ &  $\mathcal{A}$ &  $\mathcal{A}$ &  $\mathcal{A}$  \\
     \midrule
     Ill-conditioning &  \XM & \XM & \CM & \XM & \CM & \CM & \XM & \XM & \XM & \XM & \XM & \XM & \XM & \CM & \CM & \XM & \CM & \XM & \XM & \XM & \XM & \XM & \XM & \CM \\
      \midrule
    Basin linearity &  $\mathcal{E}$\tnote{$\bm\ast$} &  $\mathcal{L}$\tnote{$\bm\ast$} &  $\mathcal{E}$ &  $\mathcal{E}$ &  $\mathcal{L}$ &  $\mathcal{L}$ &  $\mathcal{E}$ &  $\mathcal{E}$ &  $\mathcal{E}$ &  $\mathcal{E}$ &  $\mathcal{E}$ &  $\mathcal{E}$ &   $\mathcal{E}$  & $\mathcal{E}$ &  $\mathcal{L}$ &  $\mathcal{E}$ &  $\mathcal{E}$ &  $\mathcal{E}$ &  $\mathcal{E}$ &  $\mathcal{L}$ & $\mathrm{L}$\tnote{$\bm\ast$} & $\mathcal{E}$ &  $\mathcal{L}$   &  $\mathcal{L}$  \\
      \midrule
     Deceptive &  \XM & \XM & \XM & \XM & \XM & \XM & \XM & \XM & \XM & \XM & \XM & \XM & \XM & \XM & \XM & \CM & \CM & \CM & \CM & \CM & \CM & \CM & \XM  & \CM \\
    \bottomrule
  \end{tabular}
 \begin{tablenotes}
\item[$\bullet$] {Existence of local optima within the basin of each component.}
\item[$\bm\dagger$] {$\mathcal{F}$, $\mathcal{N}$ and  $\mathcal{P}$ stand for fully separable, non-separable, and partially separable, respectively.}
\item[$\bm\ddagger$] {$\mathcal{S}$ and $\mathcal{A}$ stand for symmetric and asymmetric, respectively.}
\item[$\bm\ast$] {$\mathcal{E}$, $\mathcal{L}$, and $\mathrm{L}$ stand for super-linear, sub-linear, and linear, respectively.}
\end{tablenotes}
\end{threeparttable}}}
 \end{table}

\subsection{Unimodal problem instances}
\label{sec:sec:unimodalScenarios}
This subsection introduces a set of unimodal problem instances, each containing a single component in the landscape (i.e., $o=1$). 
These functions are defined in a 30-dimensional space ($d=30$) with parameters $\bm\mu=(0,0)$ and $\bm\omega=(0,0,0,0)$. 
The minimum position $\vec{m}$ (global optimum position) is randomly selected from the range $[-80,80]^d$, and $\sigma$ (global optimum value) is randomly selected from the range $[-1200,0]$.
Note that the random seed used for generating the minimum position $\vec{m}$ remains constant for each problem instance with a single component. This uniform seed ensures consistent and stable performance measurements across all runs.

\subsubsection{$f_1$}

The first instance, labeled as $f_1$, is constructed by setting $\lambda=1$ with $\mathbf{H}=\mathbf{R}=\mathbf{I}_{d \times d}$. 
It mirrors characteristics found in the widely recognized Sphere function. 
This problem is symmetric, regular, smooth, well-conditioned, and fully separable, which facilitates dimension-wise exploitation.
Researchers can employ this instance to examine the convergence speed of optimization algorithms. 
Figure~\ref{fig:f1} visualizes a 2-dimensional representation of $f_1$.

\subsubsection{$f_2$}

The configuration for $f_2$ is very similar to that of $f_1$.
The distinguishing factor is the setting of $\lambda=0.05$ for this instance. 
Much like $f_1$, $f_2$ is symmetric, regular, well-conditioned, and separable. 
However, unlike $f_1$, $f_2$ has a sub-linear basin, resulting in a narrow shape around its optimal point.
This attribute tests the optimization algorithms' ability to find the global optimum within a specific region, making it useful for evaluating algorithms' exploitation performance.
Since $f_2$ lacks other significant challenges, it serves as a focused benchmark to assess the precision of algorithmic exploitation. 
Figure~\ref{fig:f2} shows a 2-dimensional projection of $f_2$.

\subsubsection{$f_3$}

For the configuration of $f_3$, we set $\lambda=1$. 
To initialize the principal diagonal of the matrix $\mathbf{H}$, a linearly spaced set of values ranging from 0.1 to $10^6$ is first generated, inclusive of both endpoints. 
These values are then randomly permuted to assign to the elements of $\mathbf{H}$'s principal diagonal, ensuring a uniform but non-sequential distribution across this range.
Moreover, we use $\mathbf{R}=\mathbf{I}_{d \times d}$ for this problem instance. 

Similar to $f_1$, this instance maintains smoothness, regularity, and full separability.
However, $f_3$ is a highly ill-conditioned unimodal problem with a condition number of $10^7$. 
It can be seen as a stretched version of $f_1$, allowing for a comparative analysis of algorithmic performance under ill-conditioned challenges. 
The complexity in $f_3$ comes from the significant difference in the magnitudes of the elements on the principal diagonal of $\mathbf{H}$, resulting in the observed ill-conditioning. 
Comparing the results obtained on $f_1$ and $f_3$ provides insights into an algorithm's ability to handle ill-conditioned scenarios and helps identify any performance differences between the two instances. 
Figure~\ref{fig:f3} shows a visual representation of $f_3$ in a 2-dimensional space.

\subsubsection{$f_4$}

The first three problem instances are fully separable, which facilitates dimension-wise optimization. 
Both $f_1$ and $f_2$ have condition numbers of unity, making them rotation-invariant.
This property means that the component's orientation remains unchanged even when rotated around its centroid. 
To introduce rotation dependency in these instances, their condition numbers must be greater than one.

For the configuration of $f_4$, we set $\lambda=1$, with every element on the principal diagonal of $\mathbf{H}$ being uniformly randomized within the range [1, 10]. 
Such a setup creates a unimodal problem instance with a condition number between one and ten.
Consequently, $f_4$ exhibits rotation-dependency while still being moderately well-conditioned. 
The associated rotation matrix $\mathbf{R}$ gets formulated via Algorithm~\ref{alg:RotationControlled}, applying $\mathfrak{p}=1$, and each angle in $\bm\Theta$ is uniformly drawn from the interval $(-\pi,\pi)$.

$f_4$ is a unimodal, regular, smooth, and non-separable instance. 
While its condition number is greater than one, it is not intended to be a severely ill-conditioned problem. 
The aim here is to assess the impact of the non-separable variable interaction framework on the behavior of optimization algorithms without the overwhelming effect of extreme ill-conditioning. 
Comparing $f_1$ and $f_4$ will allow for a focused assessment, isolating the effect of non-separability on algorithmic performance. 
Figure~\ref{fig:f4} shows a 2-dimensional projection of $f_4$.

\subsubsection{$f_5$}

For $f_5$, we set $\lambda=0.05$. 
The elements of the principal diagonal of the matrix $\mathbf{H}$ are determined using a method similar to that employed in $f_3$: we first generate a linearly spaced set of values ranging from 0.1 to $10^6$, inclusive of both endpoints. 
These values are then randomly permuted to assign to the principal diagonal of $\mathbf{H}$.
In $f_5$, each variable $i$, interacts with the succeeding variable, $(i+1)$, resulting in a chain-like variable interaction structure that defines a minimally connected, non-separable problem instance. 
The rotation matrix $\mathbf{R}$ is computed using Algorithm~\ref{alg:RotationControlled}, with the rotation angles for each plane being randomly determined from the interval $(-\pi,\pi)$.

$f_5$ is defined by its non-separability and a high condition number of $10^7$.
Moreover, it has a very sub-linear basin.
The synergy of ill-conditioning, sub-linear basin, and chain-like variable interaction crafts a distinctive landscape: a sharply defined, rotated valley. 
Such a landscape typically presents a difficult challenge for many optimization algorithms, especially in terms of accurate exploitation and quick convergence.
With $f_5$, our objective is to critically assess how optimization algorithms handle multiple complexities. 
This problem instance is a challenging unimodal test, designed to highlight the algorithms' ability (or lack thereof) to navigate complex, narrow search landscapes. 
Figure~\ref{fig:f5} offers a visual representation of $f_5$ in a 2-dimensional domain.

\subsubsection{$f_6$}

\begin{figure*}[!t]
\centering
\begin{tabular}{ccc}
    \subfigure[{\scriptsize $f_1$.}]{\includegraphics[width=0.30\linewidth]{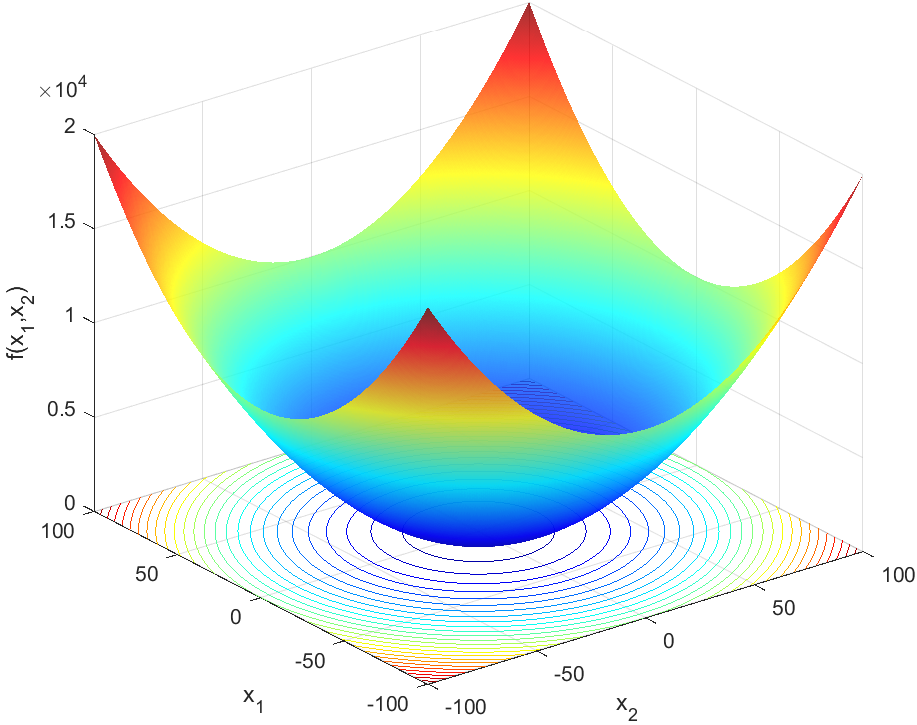}\label{fig:f1}}
&
     \subfigure[{\scriptsize $f_2$.}]{\includegraphics[width=0.30\linewidth]{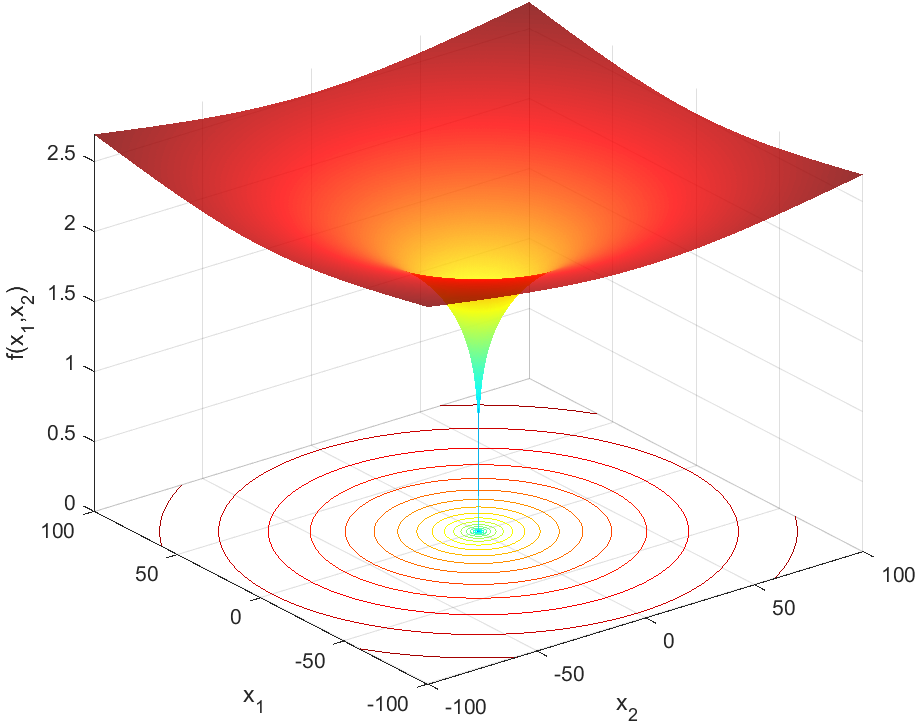}\label{fig:f2}}
&
\subfigure[{\scriptsize $f_3$.}]{\includegraphics[width=0.30\linewidth]{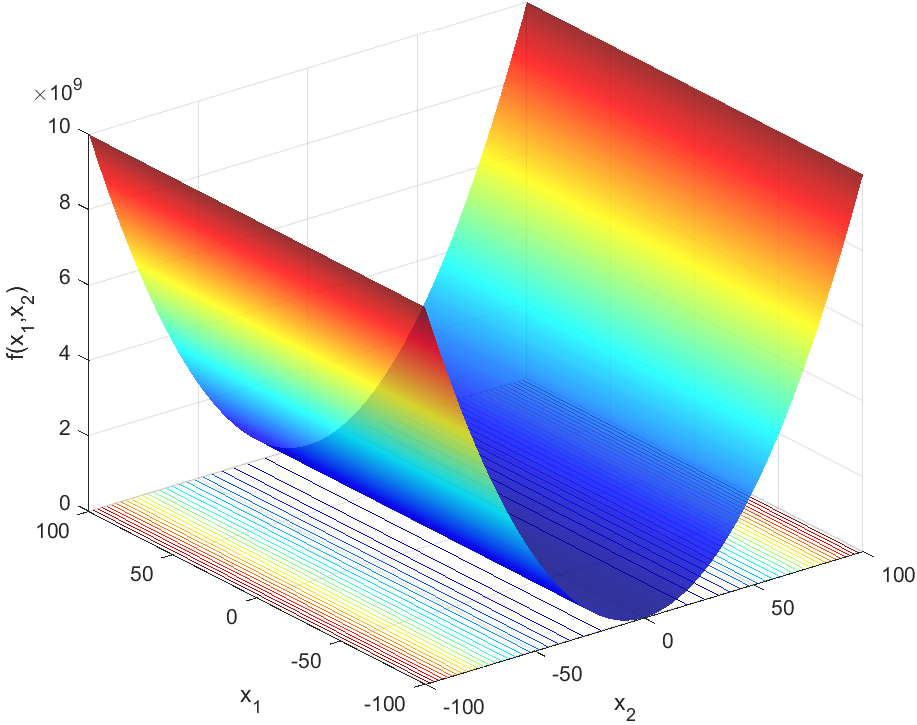}\label{fig:f3}}
\end{tabular}
\begin{tabular}{cc}
 \subfigure[{\scriptsize $f_4$.}]{\includegraphics[width=0.30\linewidth]{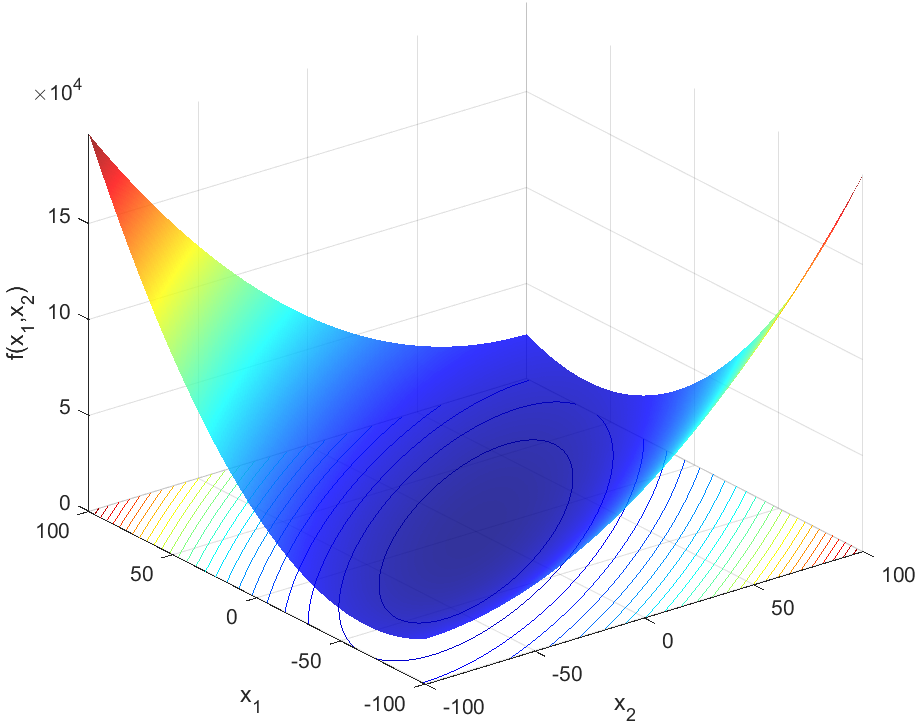}\label{fig:f4}}
&
\subfigure[{\scriptsize $f_5$ and $f_6$.}]{\includegraphics[width=0.30\linewidth]{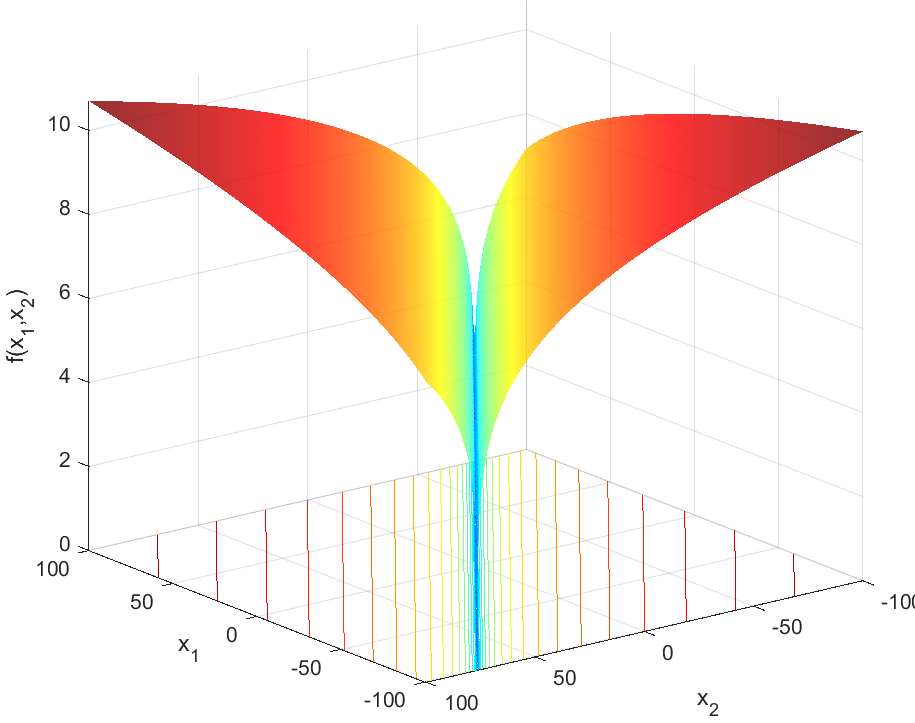}\label{fig:f5}}
\end{tabular}
\caption{Visualization of the 2-dimensional search space for unimodal functions $f_1$ to $f_6$. 
To enhance the clarity of the morphology depiction, we set $\sigma$ to zero and $\vec{m}$ to [0,0].}
\label{fig:Unimodal}
\end{figure*}

For $f_6$, we use a configuration similar to $f_5$. 
However, a key difference is in the variable interaction structure: $f_6$ has the maximum connectivity, making it a fully-connected, non-separable function. 
By comparing the outcomes on $f_6$ with those on $f_5$, we can see how algorithmic performance is affected by the details of connectivity structures in non-separable functions. 
Since in a 2-dimensional domain, $f_6$ and $f_5$ are similar, a separate illustration for $f_6$ has not been included in Figure~\ref{fig:Unimodal}.

\subsection{Multimodal Problem Instances with a Single Component}
\label{sec:sec:MultimodalScenarios}

This section introduces a series of multimodal problem instances, labeled as $f_7$ to $f_{15}$, each containing a single component in their landscape (i.e., $o=1$). 
Each instance is designed in a 30-dimensional space ($d=30$), with the minimum position, $\vec{m}$, chosen randomly within the range $[-80, 80]^d$. 
The global optimum value, $\sigma$, is randomly selected from the range $[-1200,0]$. 
To ensure consistent and reproducible performance evaluations, the random seed employed for generating both the minimum position $\vec{m}$ and value $\sigma$ is kept constant across all instances. 
This consistency aids in offering reliable comparisons of optimization algorithms across these multimodal scenarios. 
In the subsequent sections, we describe these instances, providing details on their unique attributes, variable interactions, and other relevant characteristics to enable a thorough analysis of optimization algorithms on these landscapes.

\subsubsection{$f_7$}

For $f_7$, we set the parameters as follows: $\lambda=1$, $\bm\mu=(0.2,0.2)$, $\bm\omega=(20,20,20,20)$, and $\mathbf{H}=\mathbf{R}=\mathbf{I}_{d \times d}$. 
This specific multimodal problem instance is distinguished by its symmetry, well-conditioning, and fully separability. 
A comparison of results from $f_7$ and $f_1$ can examine the challenges introduced by the presence of local optima in the basin of a component.
Note that the ruggedness and modality of $f_7$ are low-moderate.
Figure~\ref{fig:f7} visualizes a 2-dimensional representation of $f_7$.

\subsubsection{$f_8$}

The configuration of $f_8$ is similar to that of $f_7$, except $\bm\omega=(50,50,50,50)$.
Similar to $f_7$, $f_8$ is multimodal, symmetric, well-conditioned, and fully separable.
However, in contrast to $f_7$, $f_8$ contains a significantly larger number of basin local optima, and these optima are narrower than those present in $f_7$.
By comparing the results obtained by the algorithms on $f_8$ with those on $f_7$, we can observe the impact of having a larger number of narrower local optima.
This characteristic poses additional challenges for some optimization algorithms, as they must navigate through a more rugged landscape with increased roughness, irregularities, and local optima. 
Figure~\ref{fig:f8} depicts a visual representation of $f_8$ in a 2-dimensional domain.

\subsubsection{$f_9$}
The configuration of $f_9$ is similar to that of $f_7$, but $\bm\mu=(1,1)$.
Similar to $f_7$, $f_9$ is also multimodal, symmetric, well-conditioned, and fully separable.
One difference in $f_9$ compared to $f_7$ is the significantly greater depth of its basin local optima.
By comparing the results obtained by the algorithms on $f_9$ with those on $f_7$, we can analyze the impact of the depth of the basin local optima on the algorithms' performance. 
The deeper optima in $f_9$ pose a greater challenge for some optimization algorithms, as they need to exploit deeper in the local optima, which may result in loss of diversity and trapping in them. 
Figure~\ref{fig:f9} visualizes a 2-dimensional representation of $f_9$.

\subsubsection{$f_{10}$}

For $f_{10}$, we aim to examine the influence of an asymmetric basin of attraction. 
To this end, we configure this instance with parameters $\bm\mu=(0.2,0.5)$ and $\bm\omega=(20,50,10,25)$, while keeping the other parameters consistent with $f_7$ to $f_9$.  
The specific values chosen for $\bm\mu$ and $\bm\omega$ give $f_{10}$ a unique feature: when observing the basin of $f_{10}$ from various directions, the complexities introduced by local optima fall between those seen in $f_{7}$ to $f_{9}$. 
The asymmetric nature of the basin sets $f_{10}$ apart. 
By comparing algorithmic results on $f_{10}$ with those from $f_{7}$ to $f_{9}$, we can identify the challenges posed by an asymmetric basin of attraction.
Figure~\ref{fig:f10} shows a visual representation of $f_{10}$ in a 2-dimensional domain.

\subsubsection{$f_{11}$}

$f_{11}$ is configured as a non-separable problem instance. 
While it shares similarities with $f_{10}$, a key distinction is its fully connected variable interaction structure. 
For $f_{11}$, the rotation matrix $\mathbf{R}$ is derived using Algorithm~\ref{alg:RotationControlled}, where $\mathfrak{p}=1$ and every angle in $\bm\Theta$ is uniformly and randomly selected from the range $(- \pi, \pi)$. 
A comparison between the algorithmic outcomes on $f_{10}$ and $f_{11}$ highlights the impact of transitioning the variable interaction structure from separable to fully non-separable. 
Specifically, it illustrates how an algorithm navigates an asymmetric multimodal space where each variable interacts with every other variable.
Figure~\ref{fig:f11} illustrates a 2-dimensional projection of $f_{11}$.

\subsubsection{$f_{12}$}

This problem instance is a partially separable version of $f_{10}$, and exhibits a unique variable interaction structure, consisting of three groups of variables with fully connected interactions within each group. 
Each group contains 10 randomly selected variables from the 30 variables of the problem instance.
No variable from one group interacts with any variable from the other two groups. 
For each group, we use a distinct angle from $\{\frac{\pi}{4}, \frac{3\pi}{4}, \frac{\pi}{8}\}$ in $\bm\Theta$ to establish variable interactions.
The purpose of introducing $f_{12}$ is to evaluate whether the algorithms can effectively identify and exploit the partial separable variable interaction structure in the problem instances.
By comparing the results obtained on $f_{12}$ and $f_{11}$ we gain insights into how well the algorithms can adapt to and benefit from the partially separable variable interaction of $f_{12}$ in comparison to the fully non-separable $f_{11}$.
Since in a 2-dimensional domain, $f_{12}$ and $f_{11}$ are similar, we do not provide a separate figure for $f_{12}$.

\subsubsection{$f_{13}$}

This problem instance is designed to examine the exploration capability of optimization algorithms within a non-separable setting. Its configuration parameters include: $\lambda=1$, $\bm\mu=(1,1)$, $\bm\omega=(50,50,50,50)$, and $\mathbf{H}=\mathbf{I}_{d \times d}$.
The rotation matrix, $\mathbf{R}$, is computed using Algorithm~\ref{alg:RotationControlled}, with $\mathfrak{p}=1$. 
Furthermore, each angle within $\bm\Theta$ is uniformly and randomly determined from the interval $(- \pi, \pi)$. 
Characteristically, $f_{13}$ is well-conditioned, symmetric, and highly rugged. 
Its fully connected non-separable nature is marked by numerous deep local optima, intensifying the optimization challenge.
Figure~\ref{fig:f13} shows a projection of $f_{13}$ in a 2-dimensional space.

\subsubsection{$f_{14}$}

This problem instance is designed to include several challenging problem characteristics simultaneously.
The configuration of this problem instance is as follows: $\lambda$ is set to 0.6, $\bm\mu=(0.7,0.2)$, and $\bm\omega=(25,10,20,50)$.
The rotation matrix $\mathbf{R}$ is obtained by Algorithm~\ref{alg:RotationControlled}, where $\mathfrak{p}=1$ and every angle in $\bm\Theta$ is uniformly and randomly selected from the range $(- \pi, \pi)$.
Additionally, we set two randomly chosen elements of the principal diagonal of $\mathbf{H}$ to 0.01 and $[10^3]$, while the remaining elements are randomly chosen (with uniform distribution) from the range $[1,10^3]$.
The characteristics of $f_{14}$ include being ill-conditioned (condition number of $\mathbf{H}$ is $[10^5]$, but the condition number of the function is lower as it is damped by the value of $\lambda$), super-linear, asymmetric, rough, irregular, and non-separable with a fully connected variable interaction structure.
Figure~\ref{fig:f14} illustrates a 2-dimensional representation of $f_{14}$.

\subsubsection{$f_{15}$}

$f_{15}$ is crafted to represent a challenging optimization landscape characterized by multiple valleys, with the optimal solution located within a narrow valley that contains local minima. 
This configuration intensifies the optimization challenge, as algorithms must adeptly navigate the rugged and complex terrain to find the global optimum.
Specifically, the configuration parameters for this instance are set as: $\lambda=0.1$, $\bm\mu=(1,1)$, and $\bm\omega=(10,10,10,10)$. The rotation matrix $\mathbf{R}$ is computed using Algorithm~\ref{alg:RotationControlled}, with $\mathfrak{p}=1$. 
Each angle in $\bm\Theta$ is uniformly and randomly determined from the interval $(- \pi, \pi)$. 

\begin{figure*}[!t]
\centering
\begin{tabular}{ccc}
    \subfigure[{\scriptsize $f_7$}]{\includegraphics[width=0.30\linewidth]{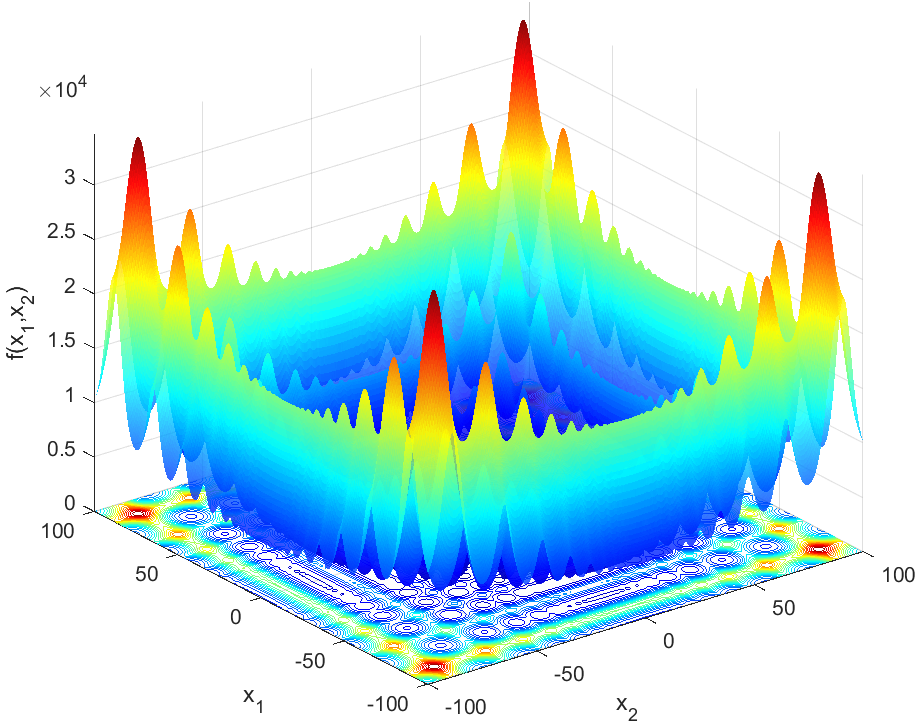}\label{fig:f7}}
&
     \subfigure[{\scriptsize $f_8$}]{\includegraphics[width=0.30\linewidth]{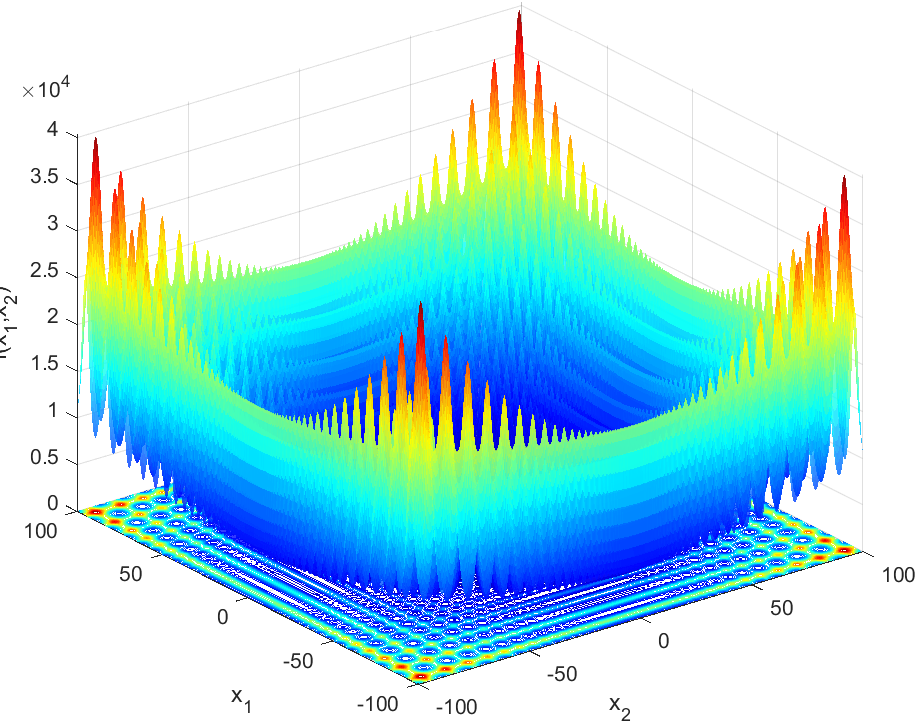}\label{fig:f8}}
&
\subfigure[{\scriptsize $f_9$}]{\includegraphics[width=0.30\linewidth]{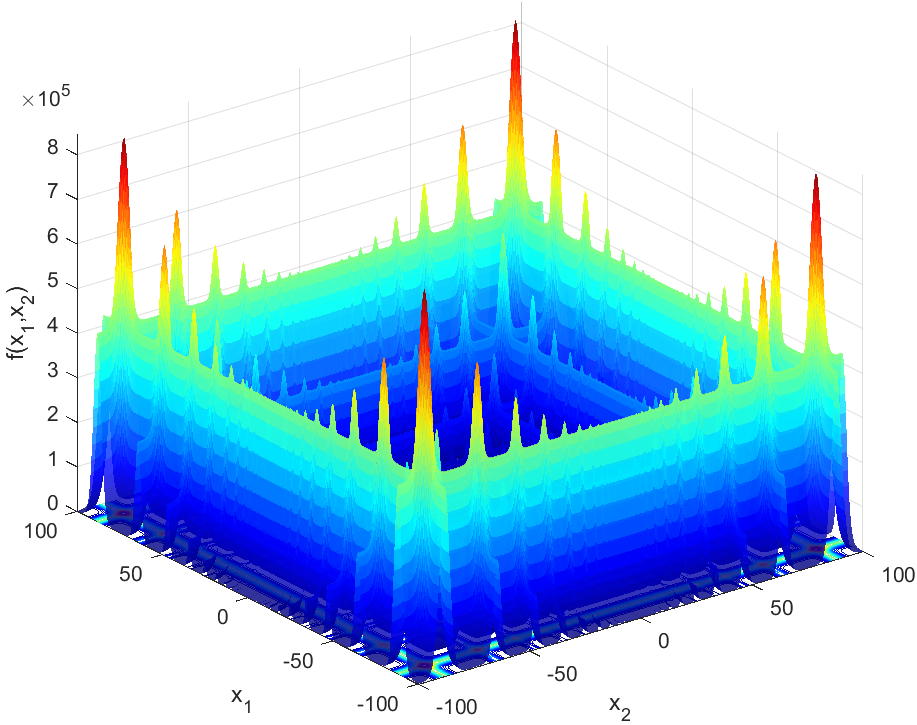}\label{fig:f9}}
\end{tabular}
\begin{tabular}{ccc}
    \subfigure[{\scriptsize $f_{10}$}]{\includegraphics[width=0.30\linewidth]{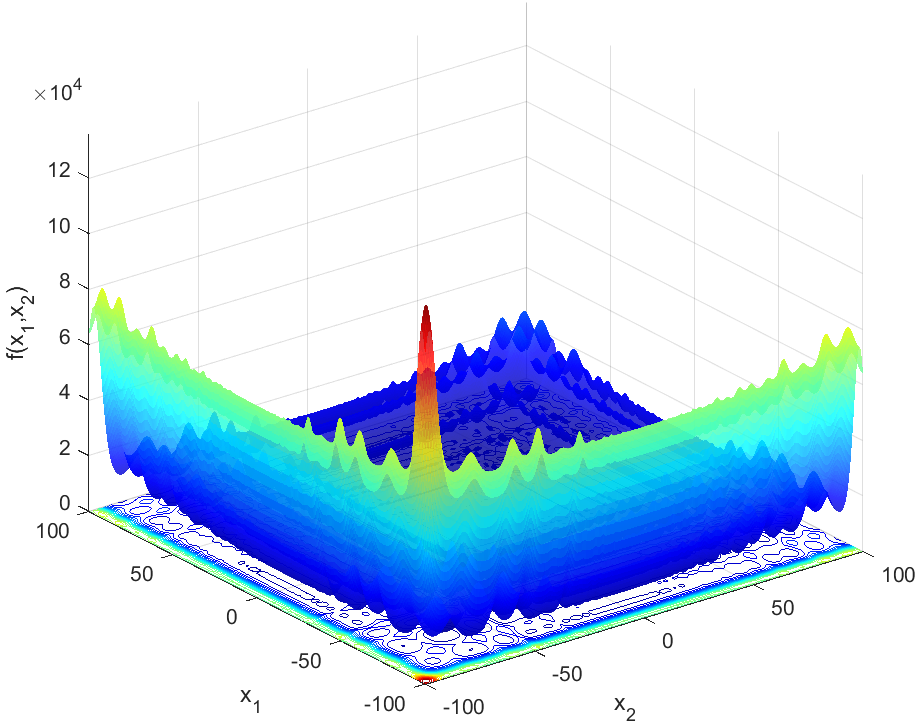}\label{fig:f10}}
&
     \subfigure[{\scriptsize $f_{11}$ and $f_{12}$}]{\includegraphics[width=0.30\linewidth]{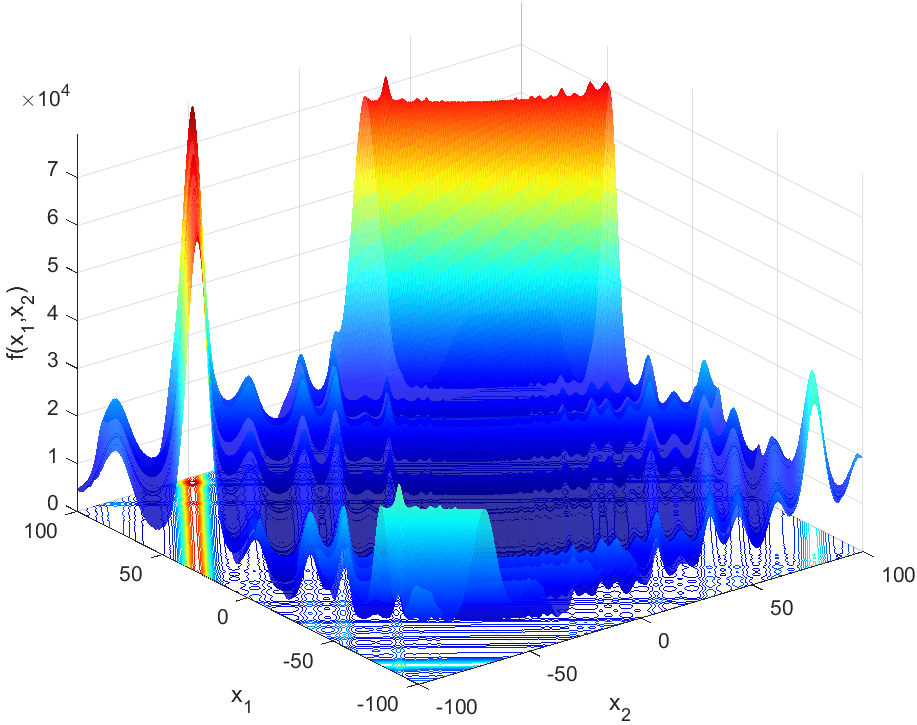}\label{fig:f11}}
&
\subfigure[{\scriptsize $f_{13}$}]{\includegraphics[width=0.30\linewidth]{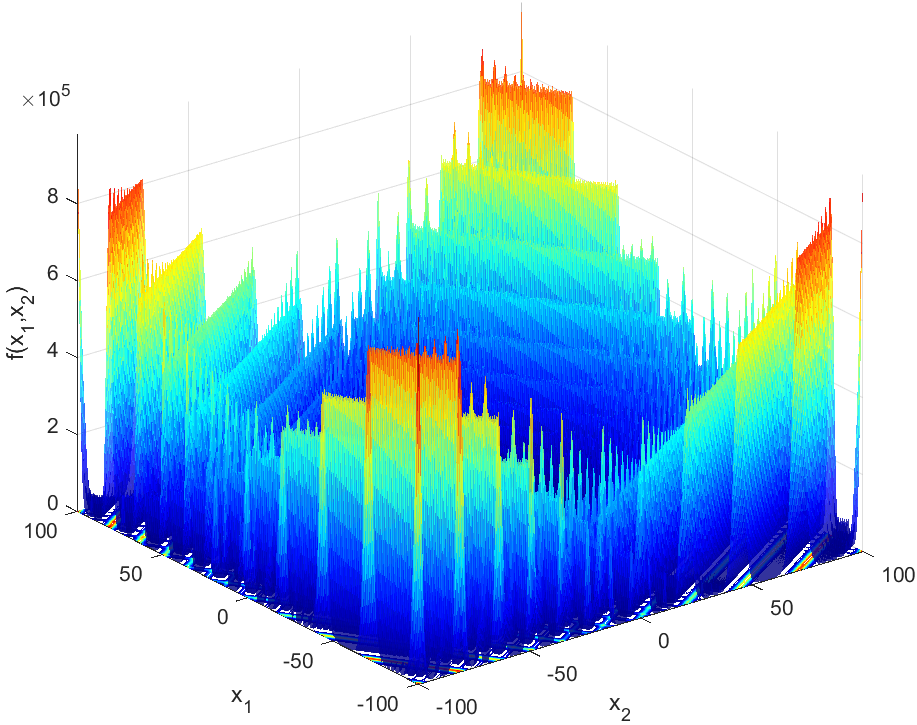}\label{fig:f13}}
\end{tabular}
\begin{tabular}{cc}
 \subfigure[{\scriptsize $f_{14}$}]{\includegraphics[width=0.30\linewidth]{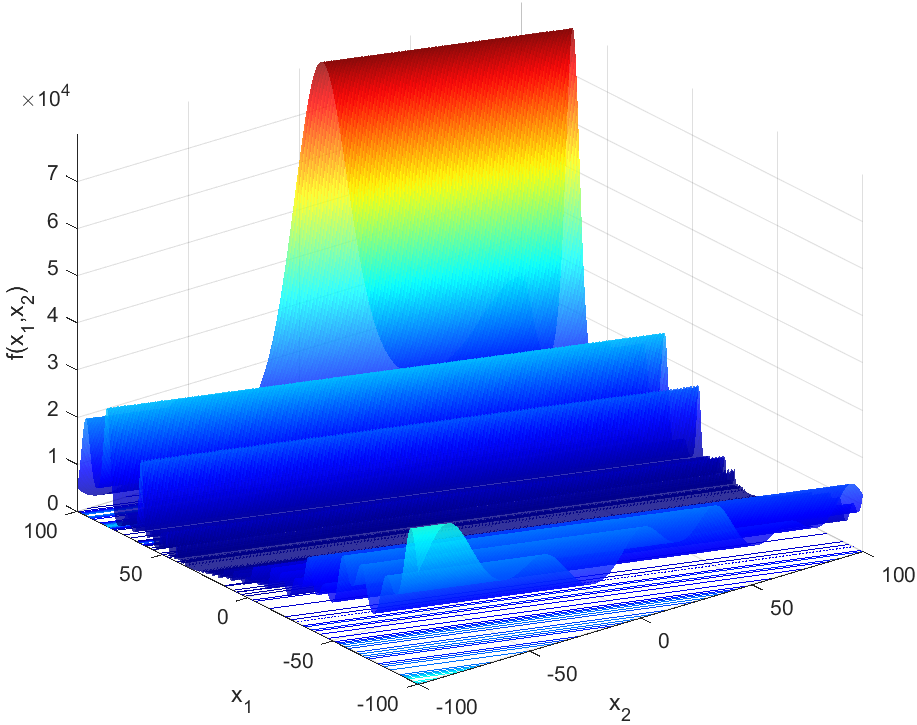}\label{fig:f14}}
&
\subfigure[{\scriptsize $f_{15}$}]{\includegraphics[width=0.30\linewidth]{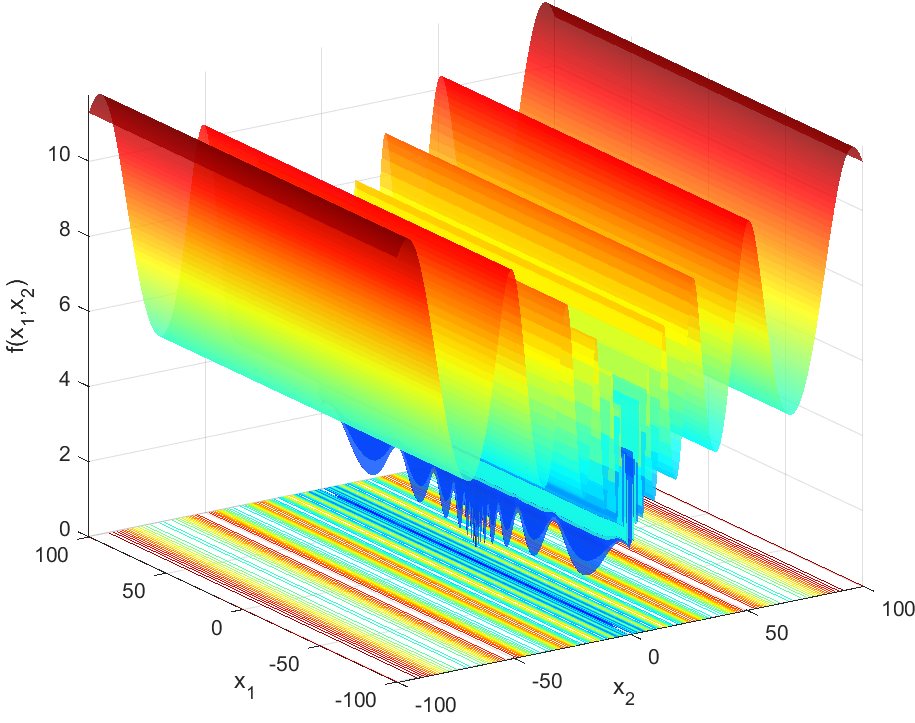}\label{fig:f15}}
\end{tabular}
\caption{Visualization of the 2-dimensional search space for multimodal functions $f_7$ to $f_{15}$. 
To enhance the clarity of the morphology depiction, we set $\sigma$ to zero and $\vec{m}$ to [0,0].}
\label{fig:Multimodal}
\end{figure*}

In addition, two elements from the principal diagonal of $\mathbf{H}$ are set to values of 1 and $10^5$ respectively. 
The remaining elements are then selected from a heavy-tail distribution by using a beta distribution with both $\alpha$ and $\beta$ parameters set to 0.2, from the range $[1,10^5]$.
$f_{15}$ exhibits properties of ill-conditioning, high sub-linearity, asymmetry, and full-connectivity in its non-separable variable interaction structure.
Figure~\ref{fig:f15} depicts a 2-dimensional projection of $f_{15}$.

\subsection{Multimodal Problem Instances with Multiple Components}
\label{sec:sec:MultiOptimaScenarios}

This section introduces a set of problem instances with multiple components (i.e., $o \geq 2$).
The landscapes of these instances present complex terrains populated by multiple competing optima, each distinguished by its unique morphological attributes and basin of attraction. 
This setup poses a substantial exploration challenge for optimization algorithms. 
Each component has an associated $\sigma_i \leq 0$, with the global optimum value being $\min_{i\in\{1,\ldots,o\}}(\sigma_i)$, and the global optimum position corresponding to the minimum position ($\vec{m}$) of the component with the smallest $\sigma$. 
Note that all problem instances in this category are inherently non-separable due to the coexistence of multiple components.
This non-separability persists irrespective of the particular variable interaction configurations within individual components~\cite{yazdani2018thesis,yazdani2019scaling}.

\subsubsection{$f_{16}$}

For $f_{16}$, we design a landscape containing  five homogeneous components. 
Each component is characterized by unimodality, smoothness, regularity, symmetry, full separability (despite the overall landscape's non-separability), and well-conditioning.
For each component $i \in \{1, \ldots, 5\}$, the configuration is delineated as follows: $\bm\mu_i=(0,0)$, $\bm\omega_i=(0,0,0,0)$, and both $\mathbf{R}_i$ and $\mathbf{H}$ are set to $\mathbf{I}_{d \times d}$, with $\lambda_i=1$. 
One component has its $\sigma$ value set to -5000, while the $\sigma$ values for the remaining components are randomly determined via $\mathcal{U}(-4500,-4000)$, where $\mathcal{U}(a,b)$ uniformly generates a random number within the interval $(a,b)$. 
The minimum position, $\vec{m}_i$, for each component is randomly drawn from the range $[-80,80]^{d}$.

In this problem instance, all parameter settings for each component are intentionally chosen to represent the most straightforward configurations. 
This design aims to test the algorithms' effectiveness in navigating the landscape and their resistance to being drawn to expansive and high-quality competing components.
Figure~\ref{fig:f16} provides an illustrative example of $f_{16}$ in a 2-dimensional domain.

\subsubsection{$f_{17}$}

$f_{17}$ is built by adding ill-conditioning and non-separability to the components of $f_{16}$.
For each component $i$, elements of the principal diagonal of $\mathbf{H}_i$ are randomly chosen using $\mathcal{U}(0.01,100)$. 
By setting $\mathfrak{p}_i=0.5$, we ensure that each component has different variable interaction structure and connectivity. 
Additionally, each angle within $\bm\Theta_i$ is determined using $\mathcal{U}(-\pi,\pi)$.
As a result, the basin of attraction for each component in $f_{17}$ exhibits varied condition numbers and distinct variable interactions, creating a landscape with heterogeneous components. 
Such heterogeneity necessitates that, when transitioning between basins in $f_{17}$, optimization algorithms adjust both their convergence direction and step size across different dimensions.

While the configurations for $\mathbf{H}_i$ and $\mathbf{R}_i$ distinguish $f_{17}$ from $f_{16}$, the remaining parameters align with those of $f_{16}$. 
The primary objective of introducing $f_{17}$ is to assess the resilience and adaptability of optimization algorithms when facing varied condition numbers and  variable interaction structures.
An example landscape of $f_{17}$ is visually depicted in Figure~\ref{fig:f17}.

\subsubsection{$f_{18}$}

This problem instance is configured similarly to $f_{16}$, but with distinct variations. 
For each component, the values for $\mu_{1,i}$ and $\mu_{2,i}$ are chosen using $\mathcal{U}(0.2,0.5)$, and values for $\omega_{1,i}$ through $\omega_{4,i}$ are randomly chosen by $\mathcal{U}(5,50)$. 
We set $\mathfrak{p}_i=0.5$, therefore, each component has different variable interaction structure and connectivity. 
In addition, each angle within $\bm\Theta_i$ is determined using $\mathcal{U}(-\pi,\pi)$.

$f_{18}$ has a highly multimodal nature with rough, irregular landscapes, along with high degrees of local and global asymmetry. 
While each component is fully separable, they have significant differences in the morphologies of their local optima basins and patterns. 
These characteristics further challenge optimization algorithms to navigate through complex landscapes filled with numerous local optima with various shapes and basin attributes.
Figure~\ref{fig:f18} shows a representative 2-dimensional visualization of $f_{18}$.

\subsubsection{$f_{19}$}

In $f_{19}$, we maintain a configuration similar to $f_{18}$, but with specific alterations. 
For each component, both $\mu_{1,i}$ and $\mu_{2,i}$ are set to 0.5, and each $\omega_{1,i}$ to $\omega_{4,i}$ value is chosen by $\mathcal{U}(50,100)$. 
This particular setting of $\bm\mu$ and $\bm\omega$ leads to the creation of an extremely rough and rugged landscape with enormous number of local optima. 
A sample 2-dimensional representation of $f_{19}$ is presented in Figure~\ref{fig:f19}.

\subsubsection{$f_{20}$}

In $f_{20}$, we use a configuration similar to $f_{18}$, with the following modifications: $\lambda_i$ for each component is set to 0.25, the minimum position $\vec{m}_i$ for each component is randomly chosen in a small sub region limited in $[-75, -25]^{d}$, and for one component $\sigma$ value is set to $-100$, while the $\sigma$ values for the remaining components are randomly determined via $\mathcal{U}(-99,-98)$. 

Due to these parameter settings, $f_{20}$ features components with a highly sub-linear basin.
Moreover, all the components in $f_{20}$ are closely packed together within a small sub-region, which occupies $\frac{1}{4^d}$ of the entire search space.
The presence of multiple sharp components in a limited space challenges certain algorithms' ability to transition between components, especially after they have converged in this small sub-region and lost their diversity. 
Furthermore, $f_{20}$ has heterogeneous components due to the varying basin local optima patterns and morphologies associated with each. 
This problem instance serves as a distinctive test case for evaluating the algorithms' efficiency in situations where several competing components are in close quarters.
In Figure~\ref{fig:f20}, a 2-dimensional example of $f_{20}$ is depicted.

\subsubsection{$f_{21}$}

For each component $i\in\{1,\ldots,5\}$ in $f_{21}$, the configuration is set as: $\lambda_i=0.5$, the values for $\mu_{1,i}$ and $\mu_{2,i}$ are chosen using $\mathcal{U}(0.1,0.2)$, and values for $\omega_{1,i}$ through $\omega_{4,i}$ are randomly chosen by $\mathcal{U}(5,10)$. 
We set $\mathfrak{p}_i=0.5$, therefore, each component has different variable interaction structure and connectivity. 
In addition, each angle within $\bm\Theta_i$ is determined using $\mathcal{U}(-\pi,\pi)$.
The $\sigma$ values for the components are set to $[-50,-45,-40,-40,-40]$ respectively. 
The second component (with $\sigma=-45$) is centrally positioned in the search space, i.e., $\vec{m}_2 = (0,0,\ldots,0)$, and has $\mathbf{H}_2=\mathrm{diag}(1,1,\ldots,1)$. 
For the other components, $\mathbf{H}_i=\mathrm{diag}(5,5,\ldots,5)$ and their minimum positions are randomly selected to lie outside the domain $[-30,30]^{d}$ but within $[-90,90]^{d}$.

The landscape of $f_{21}$ showcases a heterogeneity in the sizes of the basins of attraction among its components. A broad promising region located at the center of the search space spans approximately half of it. This dominant basin is around five times larger than the others, making $f_{21}$ a deceptive problem. Such a landscape challenges optimization algorithms to bypass the extensive central basin and discover the other narrower components. Given its deceptive nature, $f_{21}$ serves as a robust test case for evaluating algorithmic performance in navigating complex landscapes with heterogeneous basins of attraction.
An illustrative 2-dimensional example of $f_{21}$ is given in Figure~\ref{fig:f21}.

\subsubsection{$f_{22}$}

In $f_{22}$, the landscape features two distant multimodal components located at opposite ends of the search space. 
For the first component, $\vec{m}_1$ is chosen using $\mathcal{U}(80,90)^d$, while for the second component, $\vec{m}_2$ is chosen using $\mathcal{U}(-80,-90)^d$. 
$\sigma_1$ and $\sigma_2$ are  set to $-1,000$ and $-950$, respectively. 
Moreover, $\lambda_1$ and $\lambda_2$ values are set to 1 and 0.9, respectively. 
Each principal diagonal element of $\mathbf{H}_1$ and $\mathbf{H}_2$ is chosen using $\mathcal{U}(1,10)$. 
$\mathfrak{p}_1 = \mathfrak{p}_2 = 0.7$ and the angle values are chosen using $\mathcal{U}(-\pi, \pi)$. 
Furthermore, the $\bm\mu_1$ and $\bm\mu_2$ values are set to [0.5,0.5], and the $\bm\omega$ values are determined by $\mathcal{U}(20,50)$. 
$f_{22}$ represents a deceptive problem instance highlighted by the presence of two high-quality, distant components. 
Using a smaller $\lambda$ value for the inferior component expands its basin of attraction, which in turn increases the deceptiveness of this instance.
In Figure~\ref{fig:f22}, a 2-dimensional landscape generated by $f_{22}$ is depicted.

\subsubsection{$f_{23}$}

In $f_{23}$, the landscape is generated by composing five overlapped components. 
Each component has $\lambda_i$ set to 0.4, and $\mathbf{H}_i = \mathbf{I}_{d \times d}$. 
The minimum position $\vec{m}_i$ for all components is identical and is randomly chosen using $\mathcal{U}(-80,80)^d$. 
Additionally, the $\sigma_i$ values for all components are set to -100. 
For all components, $\mathfrak{p}_{i} = 0.75$ and the angle values for each $\bm\Theta_i$ are chosen randomly using $\mathcal{U}(-\pi, \pi)$, ensuring that each component is uniquely rotated in different planes. 
Moreover, for each component, both $\mu_{1,i}$ and $\mu_{2,i}$ are set to 0.5, and each of $\omega_{1,i}$ to $\omega_{4,i}$ is randomly chosen using $\mathcal{U}(20,50)$.
These five overlapped and randomly rotated components yield a single, highly irregular and rough visible component in the landscape. 
This results in a complex local optima pattern, wherein the variable interaction structure across these local optima is diverse. 
A 2-dimensional landscape generated by $f_{23}$ is illustrated in Figure~\ref{fig:f23}.

\subsubsection{$f_{24}$}

\begin{figure*}[!t]
\centering
\begin{tabular}{ccc}
    \subfigure[{\scriptsize $f_{16}$}]{\includegraphics[width=0.30\linewidth]{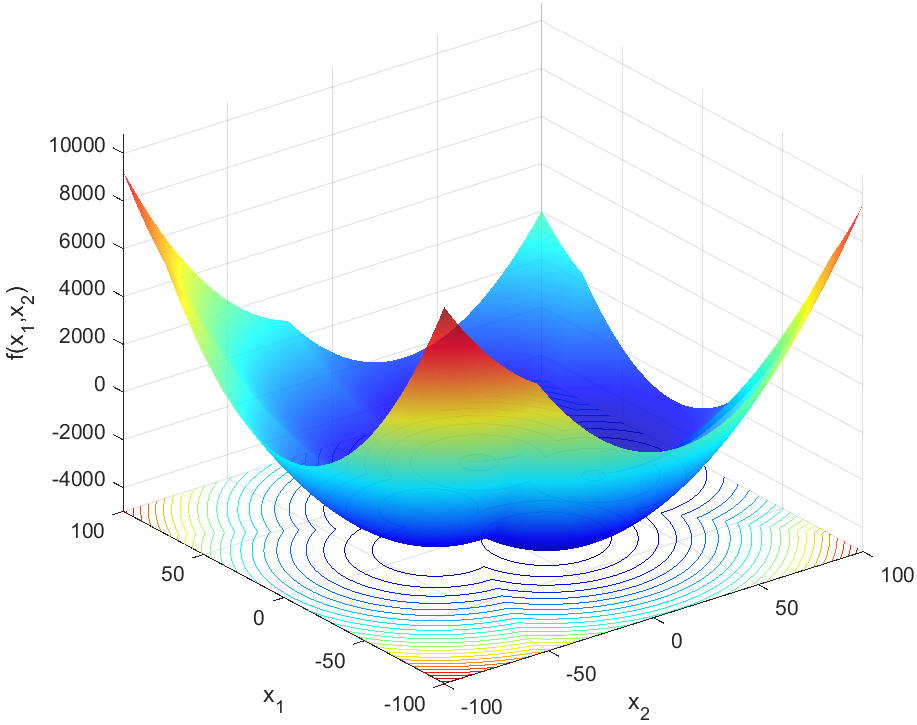}\label{fig:f16}}
&
     \subfigure[{\scriptsize $f_{17}$}]{\includegraphics[width=0.30\linewidth]{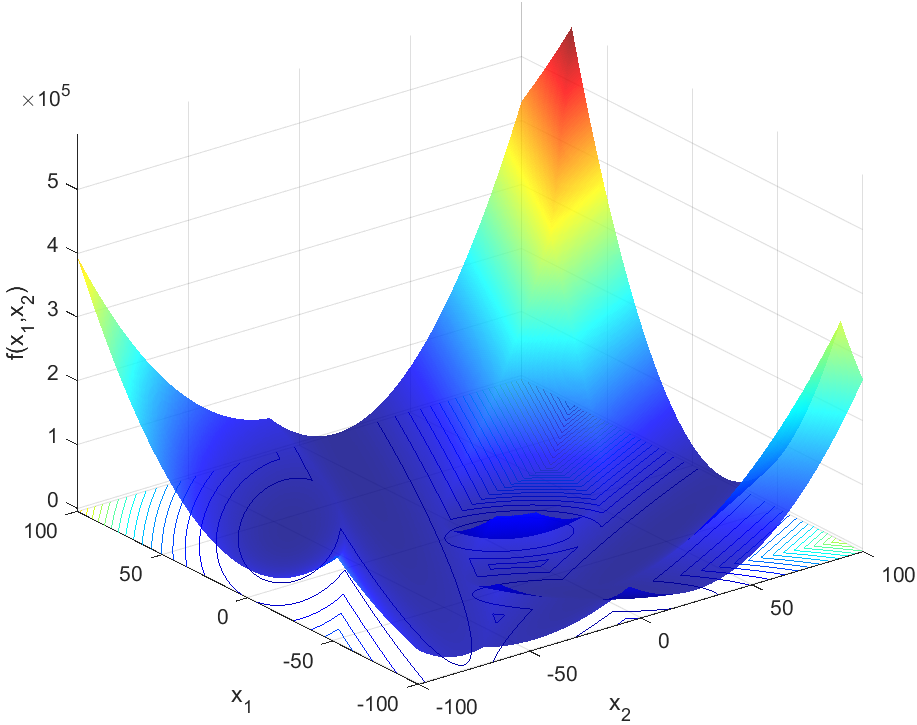}\label{fig:f17}}
&
\subfigure[{\scriptsize $f_{18}$}]{\includegraphics[width=0.30\linewidth]{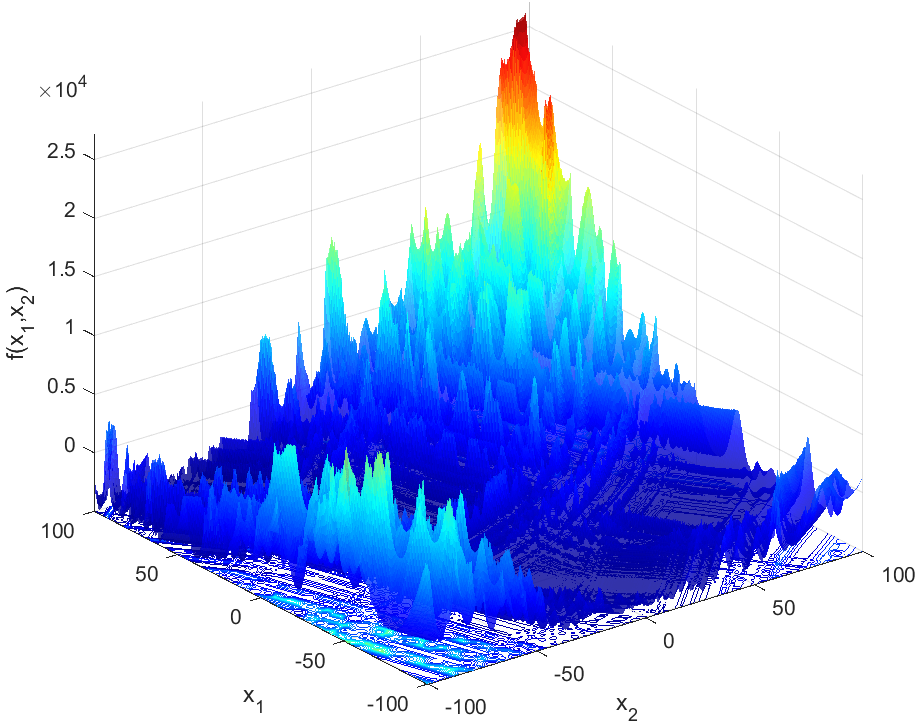}\label{fig:f18}}
\\
    \subfigure[{\scriptsize $f_{19}$}]{\includegraphics[width=0.30\linewidth]{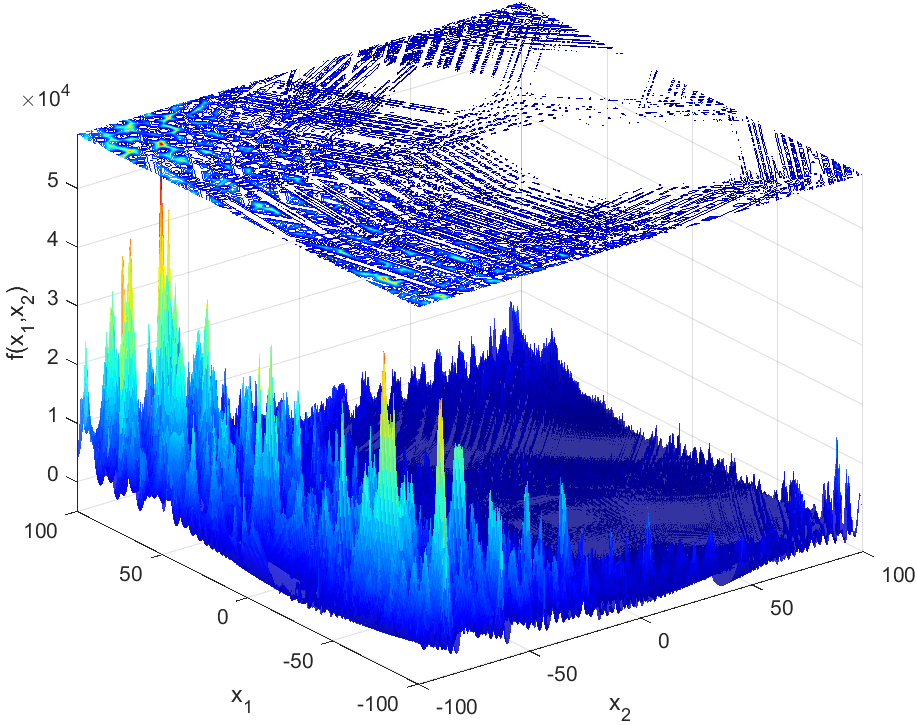}\label{fig:f19}}
&
     \subfigure[{\scriptsize $f_{20}$}]{\includegraphics[width=0.30\linewidth]{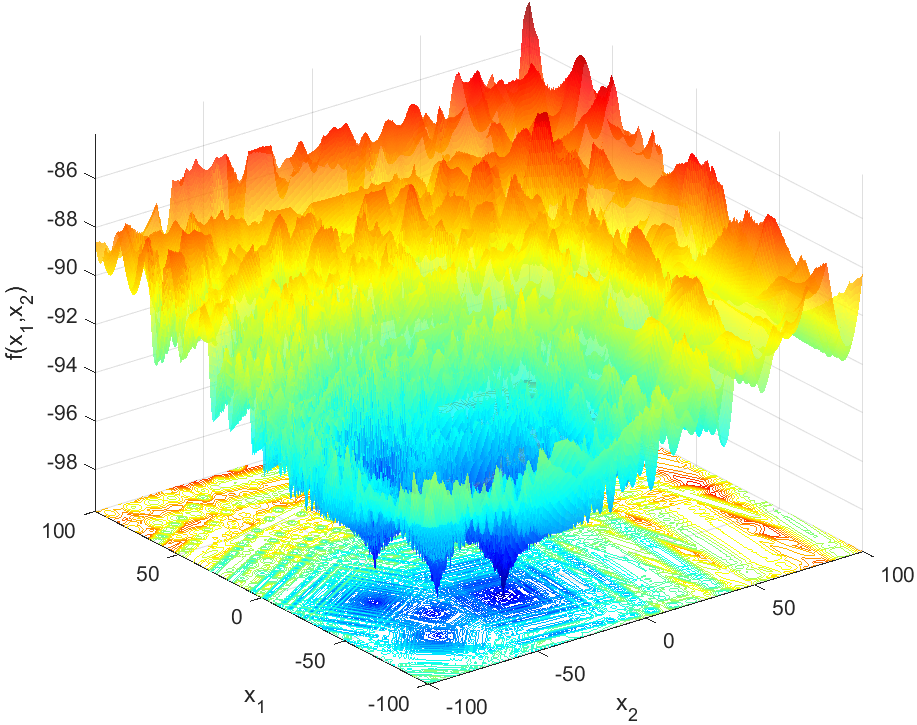}\label{fig:f20}}
&
\subfigure[{\scriptsize $f_{21}$}]{\includegraphics[width=0.30\linewidth]{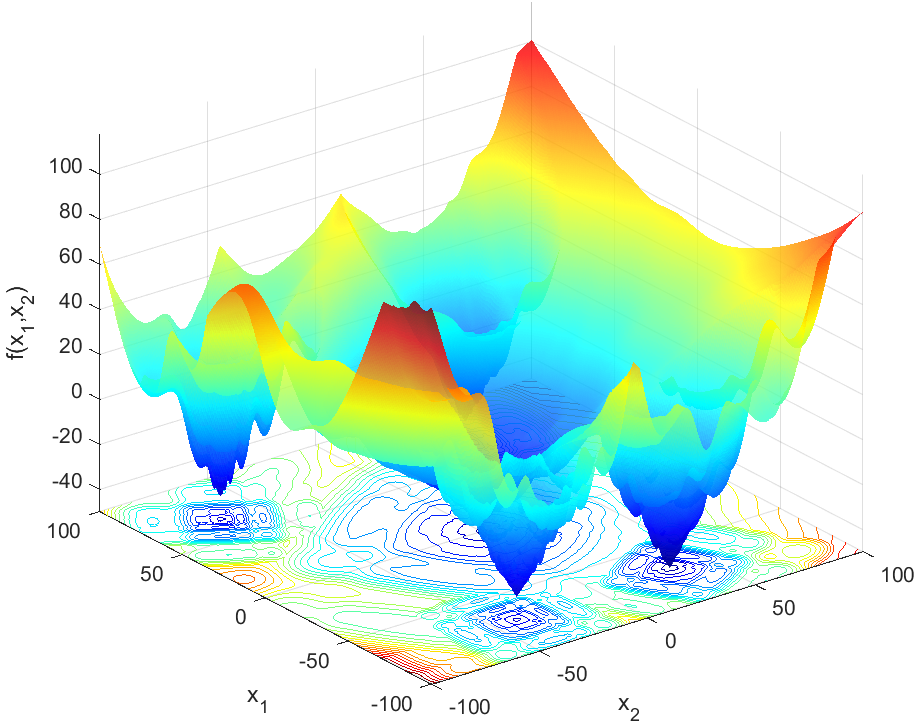}\label{fig:f21}}
\\
    \subfigure[{\scriptsize $f_{22}$}]{\includegraphics[width=0.30\linewidth]{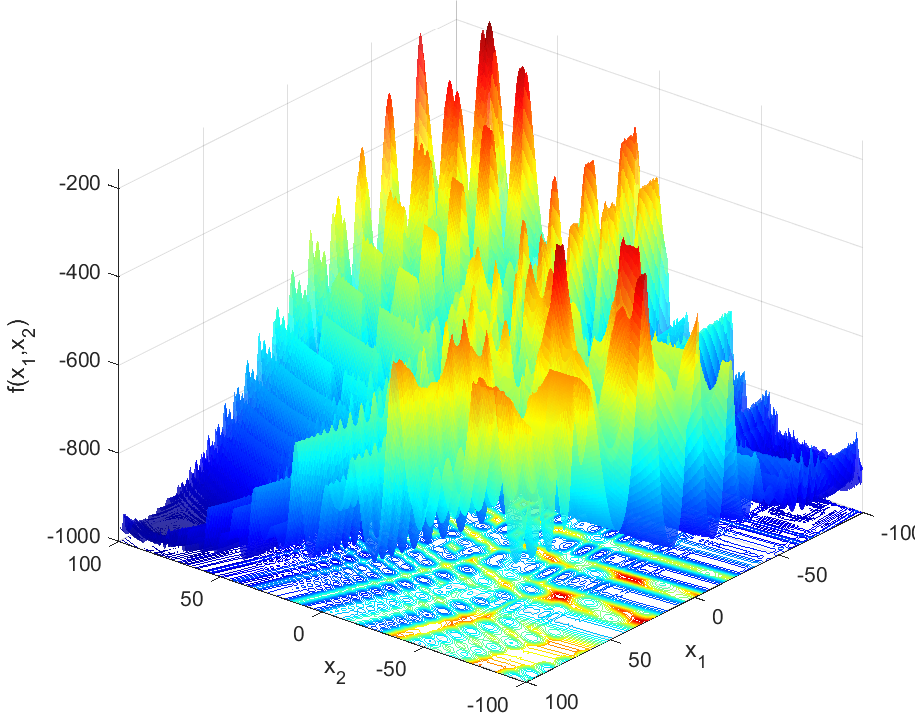}\label{fig:f22}}
&
     \subfigure[{\scriptsize $f_{23}$}]{\includegraphics[width=0.30\linewidth]{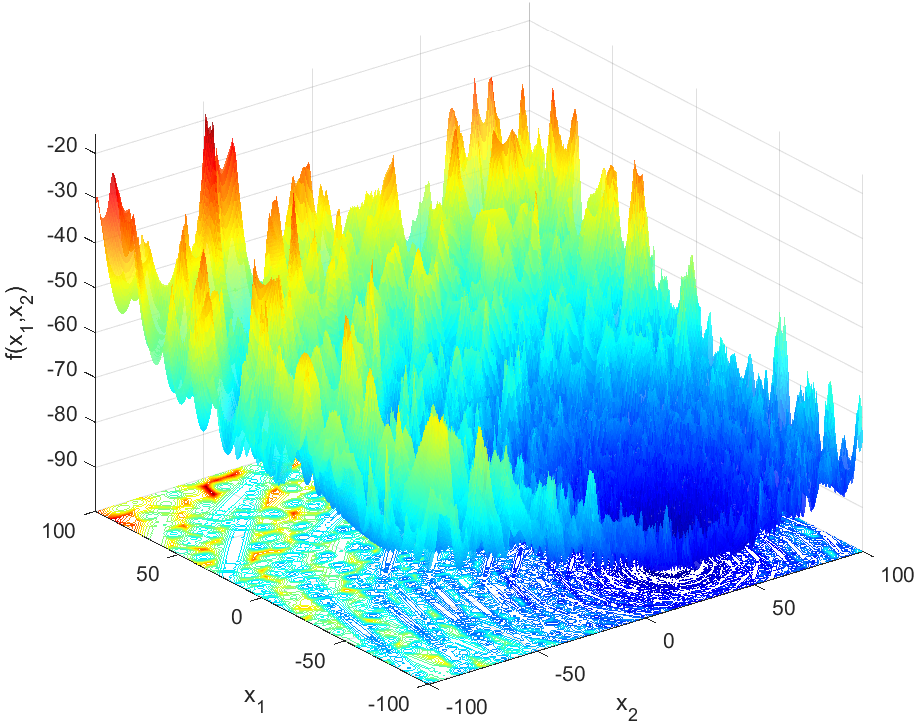}\label{fig:f23}}
&
\subfigure[{\scriptsize $f_{24}$}]{\includegraphics[width=0.30\linewidth]{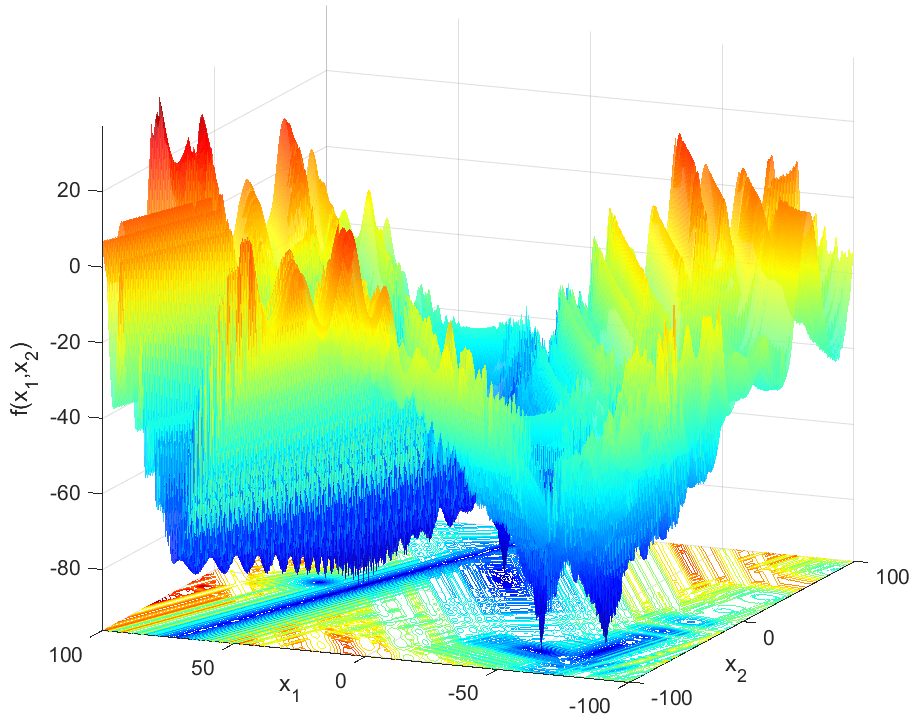}\label{fig:f24}}
\end{tabular}
\caption{Visualization of the 2-dimensional search space for multimodal functions with multiple components $f_{16}$ to $f_{24}$. }
\label{fig:MultiComponent}
\end{figure*}

This problem instance contains five heterogeneous components with various challenging characteristics. 
For each component, $\lambda_i$ is set to 0.25. 
One component is assigned a $\sigma$ of -100, while the $\sigma$ values for the remaining components are randomly determined via $\mathcal{U}(-99,-98)$.  
The minimum position, $\vec{m}_i$, for each component is randomly selected from the range $[-80,80]^d$.
The values for $\mu_{1,i}$ and $\mu_{2,i}$ are chosen using $\mathcal{U}(0.2,0.5)$, and values for $\omega_{1,i}$ through $\omega_{4,i}$ are randomly selected by $\mathcal{U}(5,50)$. 
Therefore, each component exhibits an asymmetric, rugged, and sub-linear basin. 
Moreover, for each component $i$, the elements of the principal diagonal of $\mathbf{H}_i$ are randomly chosen using $\mathcal{U}(1,10^5)$, leading to the generation of ill-conditioned components with different conditioning degrees. 
By setting $\mathfrak{p}_i = 0.75$, we ensure that each component exhibits a different non-separable variable interaction structure with a high degree of connectivity. 
Additionally, each angle within $\bm\Theta_i$ is determined using $\mathcal{U}(-\pi,\pi)$.
Due to these complex characteristics, $f_{24}$ presents a highly challenging landscape where both the exploration and exploitation capabilities of optimization algorithms are thoroughly tested.
Figure~\ref{fig:f24} shows a 2-dimensional landscape generated by $f_{24}$.

\setstretch{0.96}
\small

\bibliography{bib}
\bibliographystyle{IEEEtran}